\documentclass[10pt,twocolumn,letterpaper]{article}

\usepackage[pagenumbers]{cvpr}

\usepackage{iftex}
\ifPDFTeX\pdfsuppressptexinfo=-1\fi
\ifPDFTeX\else
  \usepackage{fontspec}
\fi
\usepackage{bm}
\usepackage{tabularx,array}
\usepackage{multirow}
\usepackage{placeins,float,needspace}
\usepackage{xurl}

\newcommand{\nmodels}{fifteen}

\newcommand{\nvideos}{54{,}600}

\newcommand{\mResp}{Response}            %
\newcommand{\mHold}{Hold}                %
\newcommand{\mComp}{Composition}         %
\newcommand{\mRoute}{Route}              %
\newcommand{\mPair}{Pair}                %
\newcommand{\mView}{Cross-view variation} %
\newcommand{\mRep}{Repeat variation}     %
\newcommand{\mScene}{Scene stability}    %

\usepackage{colortbl}
\definecolor{fone}{HTML}{DFEAF9}
\definecolor{ftwo}{HTML}{DBF0E5}
\definecolor{fthree}{HTML}{F8EBCF}
\definecolor{tabhi}{HTML}{E6EBF2}  %
\definecolor{alertred}{HTML}{D64541}
\definecolor{oursrow}{HTML}{F1F4F9}
\renewcommand{\mHold}{Pace score}              %
\renewcommand{\mComp}{Composition score}       %

\definecolor{cvprblue}{rgb}{0.21,0.49,0.74}
\usepackage[breaklinks,colorlinks,allcolors=cvprblue]{hyperref}

\newcommand{\papertitle}{StrideWorld: Benchmarking Scene Rigidity and Action Consistency\\in Interactive Video World Models}
\title{\papertitle}

\author{Chuang Wang$^{1,2*}$\quad Rui Nie$^{4*}$\quad Jie Song$^{5*}$\quad Junyuan Gao$^{2*}$\\
Jiahe Song$^{1,2}$\quad Haitao Zhou$^{4}$\quad Boyu Zhu$^{2}$\quad Xing Li$^{5}$\\
Jiang Wu$^{2}$\quad Conghui He$^{2}$\quad Qian Yu$^{4}$\quad Weijia Li$^{3,2\dagger}$\\
\normalsize $^{1}$Shanghai Jiao Tong University\quad $^{2}$Shanghai Artificial Intelligence Laboratory\\
\normalsize $^{3}$Shenzhen International Graduate School, Tsinghua University\quad $^{4}$Beihang University\\
\normalsize $^{5}$Tianjin University\\
\small \texttt{qianyu@buaa.edu.cn}\quad \texttt{liweijia@sz.tsinghua.edu.cn}}

\newif\ifwithsupp
\withsupptrue

\begin{document}
\twocolumn[{%
\renewcommand\twocolumn[1][]{#1}%
\maketitle
\vspace{-2mm}
\begin{center}
    \centering
    \includegraphics[width=\textwidth]{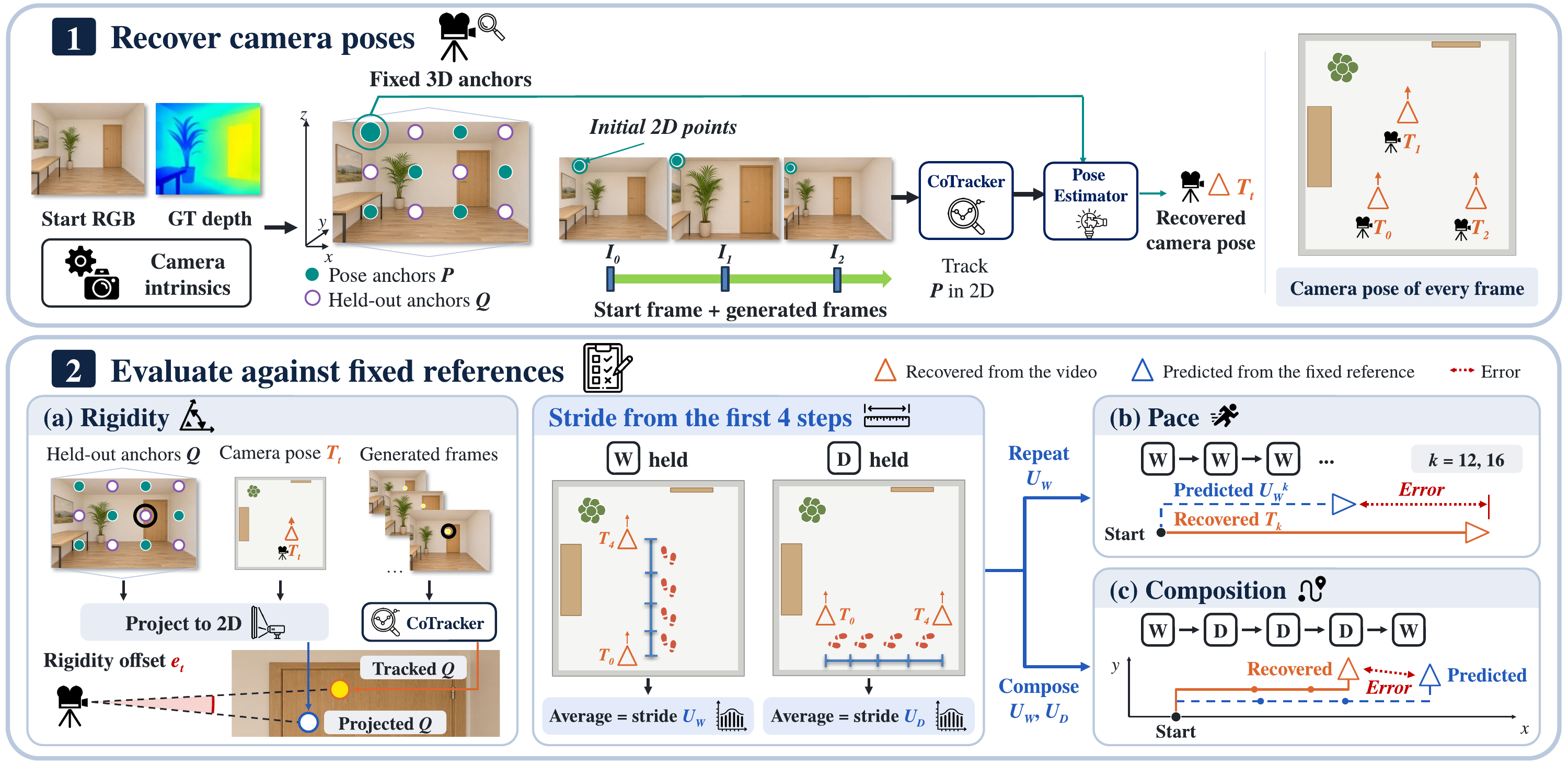}
    \vspace{-2mm}
    \captionof{figure}{\textbf{StrideWorld fixes every reference before the motion it judges.} Step~1 recovers the camera of each generated frame from pose anchors placed at the ground-truth depth of the start view. Step~2 tests (a)~whether held-out anchors stay where this camera projects them, and whether (b)~a held action and (c)~a composed route end where the strides averaged over the first four steps predict (schematic).}
    \label{fig:teaser}
\end{center}
\vspace{0mm}
}]

\begingroup
\renewcommand{\thefootnote}{}
\footnotetext{$^{*}$Equal contribution. $^{\dagger}$Corresponding authors.}
\endgroup
\begin{abstract}
Interactive video world models turn user commands into the exploration of a generated scene, which requires the scene to keep its 3D structure as the camera moves and each action to keep a stable effect. Existing benchmarks mostly evaluate these properties against references fitted to the video under test, either a scene reconstructed from its own frames or a camera trajectory aligned to the command path with a free scale. Such references can absorb gradual scene deformation and changes in pace and route length, so these errors can pass undetected. We therefore introduce StrideWorld, a benchmark whose references are fixed before the motion they judge. It evaluates scene rigidity against the ground-truth geometry of the start view, and action consistency against the stride of each action, the camera motion of one command. The stride is measured on the model's first steps and then used to test the steps that follow, so each model is judged only against its own steps. Its measurements are validated on rendered videos with known camera motion, and its scores agree with human judgments. Across \nvideos{} videos generated by \nmodels{} open models from $140$ start views, all models but one deform their scenes as the camera moves, with offsets up to $7.3$ times those of a rigid rendering along the same camera path. Most models keep a steady pace when a single action is held. Along composed routes, several models often stall, and most completed routes end away from the positions predicted by each model's own strides. Only the two models that scale commands by the estimated scene depth bring most routes close to these positions. Fine-tuning two models on a few rendered clips removes most of their mid-route stalls. Code, data, and a leaderboard are publicly available at \url{https://github.com/chuangwang123/strideworld}.
\end{abstract}

\section{Introduction}\label{sec:introduction}

Interactive video world models shift video generation from producing clips for passive viewing to synthesizing environments that respond to user commands~\citep{gamengen,matrix,matrix2,gamecraft,Worldmem,worldplay,lyra2,lingbotv2,dreamxworld}. Systems such as Genie 3~\citep{genie3,genie2,genie,gamengen} build controllable generative worlds on demand, aiming to offer a new form of interactive entertainment. For the generated content to constitute a world rather than a sequence of individually plausible frames, it must first preserve its 3D structure, without deformation, as the viewpoint changes. At the same time, usable interaction requires the model to give stable feedback to user commands. Basic operations such as moving and turning the camera should have a consistent and predictable effect throughout the interaction.

Evaluating whether an interactive world model meets these requirements takes more than measuring the perceptual quality targeted by conventional video benchmarks~\citep{vbench,vbench++,evalcrafter}. Recent benchmarks have therefore begun to assess such models along two axes, spatial consistency and action control. To assess 3D structure, metrics such as those of WBench and MEt3R~\citep{wbench,met3r,worldexam,worldmark} reconstruct the scene from the generated frames with 3D foundation models such as VGGT~\citep{vggt} and Depth Anything 3~\citep{lin2025depth}. They then check whether the frames are consistent with this reconstruction. This geometric reference, however, is fitted to the video under test. Since such foundation models can fit seemingly coherent geometry even to inconsistent images~\citep{paul2026can}, gradual scene deformation can be absorbed into the reference and go undetected by these benchmarks.

For action control, the reference raises the same issue. Benchmarks such as WorldScore and WBench~\citep{worldscore,wbench,worldexam} compare the recovered camera trajectory with the commanded one after aligning their scales, which accommodates models with different control interfaces. This scale is likewise estimated from the trajectory under test, so it averages out changes in the distance per command. A model whose steps lengthen or shorten midway can therefore still match the commanded path. Loop-closure tests~\citep{loopnav,kwon2025stable,worldmark,worldodysseybench,mind,wang2026world,worldcycle} avoid rescaling but typically assume that opposite commands cancel, an assumption that many current models violate, as shown in App.~\ref{app:H5}. They also test whether a loop returns to its start rather than whether each command keeps a stable effect along an open route. How to test scene geometry and action control against references independent of the video under test therefore remains open.

To this end, we introduce \textbf{StrideWorld}, a benchmark for scene rigidity and action consistency. Each of its references is fixed before the motion it judges, as shown in Fig.~\ref{fig:teaser}. The \textbf{Rigidity} test places 3D anchors at the ground-truth depth of each start view and checks whether the visible surfaces stay in place as the camera moves. For actions, we measure the \emph{stride} of each action from the model's own first four steps, so models with different interfaces need no shared unit of motion. The stride is the camera motion of one command, and it predicts the camera pose at every later step. The \textbf{Pace} test checks whether the camera keeps this stride when a single action is held, and the \textbf{Composition} test checks whether the strides add up along multi-action routes. These routes come in pairs designed to reach the same end, for example the same translations in a different order. Each route is compared with its own prediction and with its partner. The camera is recovered at every step, which locates where a route stalls or drifts from its prediction.

We evaluate \nmodels{} open models on \nvideos{} videos generated from $140$ start views across $30$ scenes, as shown in Fig.~\ref{fig:bench}. All models but one deform their scenes as the camera moves. How well a model preserves the scene does not predict how well it composes actions. When a single action is held, most models move in the commanded direction and keep a steady pace. When actions are combined into routes, however, several models often stall partway. Even completed routes mostly end away from the positions predicted by the model's own strides. Part of this error arises at action switches, where models carry the previous action over or start the new one slowly. Only the two models that scale commands by the estimated scene depth bring most routes close to their predictions. Existing metrics largely miss these failures, as a reconstruction-based reprojection error detects scene deformation in only three of the fifteen models. When the pace of a rendered video grows by half, a scale-fitted trajectory error flags $43\%$ of such videos, whereas the Pace test flags $94\%$ at the same false-alarm rate. Finally, fine-tuning two models on $112$ rendered clips with exact camera paths removes most of their mid-route stalls and more than doubles their \mComp s.

In summary, our contributions are as follows:
\begin{enumerate}[leftmargin=*,itemsep=1pt,topsep=2pt]
\item We propose \textbf{StrideWorld}, which measures scene rigidity against the ground-truth geometry of the start view and action consistency against each model's own action strides. We validate its measurements on rendered videos with known camera motion and its scores against the judgments of ten human raters.
\item We benchmark \nmodels{} open interactive video world models and find that most deform their scenes as the camera moves. They keep a steady pace under a single action but often fail when actions are combined into routes.
\item We show that fine-tuning two models on $112$ rendered clips with exact camera paths removes most of their mid-route stalls and brings route endpoints closer to their predictions without lowering image-quality scores.
\end{enumerate}

\section{Related Work}\label{sec:related}

\paragraph{Interactive video world models.}
World models were first studied in reinforcement learning as learned simulators of an agent's environment~\citep{ha2018recurrent,hafner2023mastering}. Building on video generation models trained at scale~\citep{brooks2024sora,wan2025,yang2024unisim,nvidia2025cosmos}, camera-controlled video models condition an entire clip on a prescribed camera trajectory~\citep{cameractrl,cami2v}, whereas 3D world generators lift generated content into explicit 3D scenes~\citep{hunyuanworld,lyra2,HY-World2}. Interactive video world models instead generate each new segment in response to user commands~\citep{genie,genie3,gamengen,gamecraft,matrix}, given as discrete keys~\citep{matrix2,matrix3,sana,dreamxworld,worldplay}, camera poses~\citep{matrix35,HY-World2,lyra2,lingbotv2,evoke,worldcrafter,solarwm,alayaworld}, or text prompts~\citep{h3world}. To keep the scene stable over long interactions, recent models maintain a memory of past frames, patches, or implicit 3D features~\citep{Worldmem,matrix3,matrix35,worldcrafter,worldplay}, condition on an explicit 3D cache~\citep{lyra2}, encode camera geometry in their positional embeddings~\citep{viewrope}, or use reinforcement learning to reduce long-horizon drift~\citep{worldcompass,worldcycle}. Their control interfaces express motion in different units, from key presses of unspecified length to camera displacements normalized by scene depth. The same command therefore moves the camera by different amounts across models.

\paragraph{Evaluating action control.}
Early video benchmarks assessed perceptual quality, physical plausibility, and instruction following~\citep{vbench,vbench++,evalcrafter,videophy,videophy2,worldmodelbench}. Interactive benchmarks add action control and use different motion references, as listed in Tab.~\ref{tab:related}. Direction metrics check the commanded motion axis~\citep{worldmark,mbench}, and WorldMark also measures held-command speed decay in optical flow. Trajectory metrics compare the recovered camera path with a reference after scale alignment or normalization~\citep{worldscore,iworld,wbench,worldexam,mind,playworld}, so their sensitivity to pace depends on the reference and the temporal alignment. WorldRoamBench~\citep{worldodysseybench} builds an adaptive reference from each predicted segment's displacement and compares paths at matched arc length. Revisit tests avoid rescaling by comparing return views, matched portions of retraced paths, or reconstructed scene segments after a loop~\citep{loopnav,kwon2025stable,worldmark,iworld,wbench,worldexam,worldodysseybench,mind}. Similarly, group-action tests~\citep{wang2026world} check identity, inverse, and composition relations, and WorldCycle~\citep{worldcycle} evaluates endpoint closure, mirrored forward--reverse frames, and phase-aligned repeated cycles. StrideWorld instead uses separately calibrated strides to measure each step on open routes, without fitting the route's scale or requiring opposite actions to cancel, since each action has its own stride.

\paragraph{Evaluating geometric consistency.}
Evaluating 3D consistency hinges on what serves as the geometric reference, which Tab.~\ref{tab:related} also lists. Most benchmarks derive this reference from the generated video itself. Like MEt3R~\citep{met3r} for generated images, they reconstruct the scene and camera from the generated frames with SLAM or 3D foundation models and measure how well the views agree with this reconstruction~\citep{worldscore,wbench,worldexam,wu2026quantitative,worldmark,worldprobe,orbia}. This approach needs no external 3D reference, but the reference is fitted to the generated frames. Because 3D foundation models can fit seemingly consistent geometry even to inconsistent images~\citep{paul2026can}, gradual deformation can be absorbed into the reference. Other benchmarks compare generated views with ground truth rendered by a game engine or simulator~\citep{mind,worldodysseybench,wmeval} or with registered RGB-D captures~\citep{orbia}. Such references constrain the available viewpoints, and control errors can enter the comparison when the generated and reference viewpoints differ. Orbia also evaluates self-consistency at recovered cameras and renders held-out generated views from a reconstructed Gaussian scene~\citep{orbia}. StrideWorld instead takes ground-truth depth only from the start view and evaluates rigidity at the recovered camera, using disjoint anchors for pose fitting and testing.

\begin{table}[t]
\centering
\caption{\textbf{Comparison with interactive world-model benchmarks.} Interf.: keys or tokens (K), camera poses (C), text (T). Reference: target motion of an action. Pace: tests of held-action speed or repeated-action proxies. Seq.: sequence protocols, not necessarily endpoint-only. Geom.: reference for geometric or image comparisons. 3D test: scene-geometry consistency, without implying disjoint fitting and testing. Hum.: reported comparisons of metrics with human judgments. $\checkmark$ yes, 2D in image space only, $\sim$ indirect proxy, -- not reported, $^\dagger$ concurrent work (App.~\ref{app:related}).}
\label{tab:related}
\begingroup
\scriptsize
\setlength{\tabcolsep}{0.4pt}
\renewcommand{\arraystretch}{1.06}
\begin{tabular*}{\linewidth}{@{\extracolsep{\fill}}lccccccc@{}}
\toprule
Benchmark & Interf. & Reference & Pace & Seq. & Geom. & 3D test & Hum. \\
\midrule
WorldScore~\citep{worldscore} & C T & trajectory & -- & multi-scene & recon. & $\checkmark$ & $\checkmark$ \\
WorldMark~\citep{worldmark} & K C T & direction & $\checkmark$\textsuperscript{2D} & open/loops & recon. & $\checkmark$ & $\checkmark$ \\
iWorld-Bench~\citep{iworld} & K C T & traj./dir. & -- & open/loops & poses/2D & $\sim$ & $\checkmark$ \\
WBench~\citep{wbench} & K C T & trajectory & $\sim$ & open/loops & recon. & $\checkmark$ & $\checkmark$ \\
WorldExam~\citep{worldexam} & K C T & traj./dir. & -- & open/loops & recon. & $\checkmark$ & $\checkmark$ \\
WorldRoamBench~\citep{worldodysseybench} & K C T & adapt./dir. & -- & open/loops & recon./GT & $\checkmark$ & $\checkmark$ \\
MIND~\citep{mind} & K & traj./GT & -- & loops & GT frames & $\sim$ & -- \\
Group Actions~\citep{wang2026world} & C & algebraic & -- & algebraic & poses & -- & -- \\
WorldCycle~\citep{worldcycle} & K C & algebraic & $\sim$\textsuperscript{2D} & loops & 2D & -- & -- \\
Orbia$^\dagger$~\citep{orbia} & K C T & direction & -- & loops & recon./GT & $\checkmark$ & $\checkmark$ \\
\midrule
\textbf{StrideWorld} & K C T & \cellcolor{tabhi}\textbf{own strides} & $\checkmark$ & \cellcolor{tabhi}\textbf{open routes} & \cellcolor{tabhi}\textbf{GT depth} & $\checkmark$ & $\checkmark$ \\
\bottomrule
\end{tabular*}
\endgroup
\end{table}

\section{The StrideWorld Benchmark}\label{sec:benchmark}\label{sec:data}

\begin{figure}[t]
    \centering
    \includegraphics[width=\linewidth]{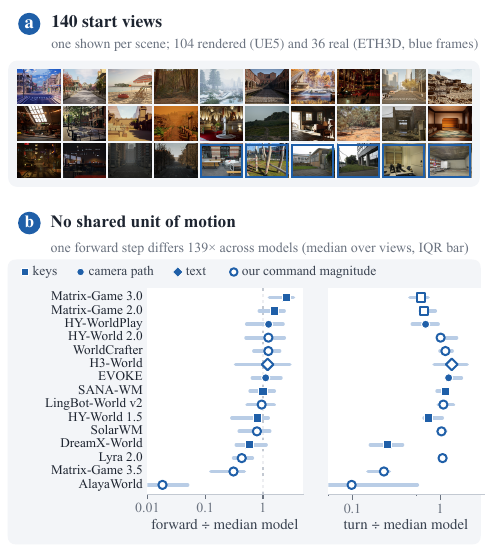}
    \caption{\textbf{StrideWorld at a glance.} (a)~One start view per scene. (b)~Forward motion and turn of a single step, each relative to the median model, across all start views. Filled markers: release default commands. Open markers: commands set to keep $16$ steps within the measurable field of view.}
    \label{fig:bench}
\end{figure}

\paragraph{Start views.}\label{sec:scenes}
To provide a 3D reference that is independent of the generated video, every start view comes with ground-truth depth and camera parameters. The benchmark comprises $140$ start views across $30$ scenes that span indoor and outdoor, natural and built environments, as shown in Fig.~\ref{fig:bench}a. Twenty-four scenes are built in Unreal Engine~5~\citep{unrealengine5} and provide $104$ start views with exact depth, and six are real ETH3D scenes~\citep{eth3d} that provide $36$ start views with depth from laser scans. Preprocessing is described in App.~\ref{app:A} and results by domain in App.~\ref{app:H2}.

\paragraph{Action programs.}\label{sec:tasks}
Every model is driven by six basic actions, forward (W), backward (S), left and right translation (A, D), and left and right yaw turns (YL, YR), with pitch evaluated separately in App.~\ref{app:G2}. One \emph{step} is a single command interval of the model's native interface. From every start view, each action is held for $16$ steps under three random seeds. The first four steps calibrate the action's \emph{stride}, and the later steps test whether the model keeps a steady pace. Each start view is also assigned eight multi-action routes of $8$ to $14$ steps, run with one seed, in which a \emph{switch} occurs whenever two consecutive steps take different actions. The routes form four pairs, and the two routes of each pair are designed to end at the same place under ideal actions. In the \emph{cancel} pair, the two routes insert back-and-forth detours of equal length along different axes into a shared base path, schematically $\mathrm{W{\to}\mathbf{A{\to}D}{\to}W}$ versus $\mathrm{W{\to}\mathbf{W{\to}S}{\to}W}$. The three \emph{swap} pairs execute the same translations in different orders, either directly ($\mathrm{\mathbf{W{\to}W{\to}D}}$ vs.\ $\mathrm{\mathbf{D{\to}W{\to}W}}$), after a turn ($\mathrm{YR{\to}\mathbf{W{\to}D}}$ vs.\ $\mathrm{YR{\to}\mathbf{D{\to}W}}$), or across turns. All sequences are kept short enough for the surfaces of the start view to remain visible, so that the camera can be recovered from ground-truth geometry throughout. Long-horizon exploration and revisits are therefore outside the scope of StrideWorld. In total, each model generates $3{,}640$ videos ($2{,}520$ single-action and $1{,}120$ route videos), for \nvideos{} videos across all models.

\paragraph{Models and interfaces.}\label{sec:adapters}
We evaluate \nmodels{} open models covering three control interfaces. Five are driven by discrete keys or tokens (Matrix-Game 2.0 and 3.0~\citep{matrix2,matrix3}, SANA-WM~\citep{sana}, DreamX-World~\citep{dreamxworld}, and HY-World 1.5~\citep{worldplay}), nine by camera poses (Matrix-Game 3.5~\citep{matrix35}, HY-World 2.0~\citep{HY-World2}, LingBot-World v2~\citep{lingbotv2}, AlayaWorld~\citep{alayaworld}, Lyra 2.0~\citep{lyra2}, EVOKE~\citep{evoke}, WorldCrafter~\citep{worldcrafter}, SolarWM~\citep{solarwm}, and HY-WorldPlay~\citep{worldplay}), and one by text prompts (H3-World~\citep{h3world}). Each model runs its released inference configuration, and a thin adapter maps the six actions to its native interface. As shown in Fig.~\ref{fig:bench}b, a single step moves the camera by widely different amounts across models, and the interfaces share no unit of motion. Because our tests measure motion in each model's own strides, command magnitudes need not match across models. We therefore use a model's default per-step command when its motion is clearly measurable and keeps the start-view surfaces visible over $16$ held steps, and otherwise set a fixed magnitude that does. All command magnitudes are released with the benchmark, as detailed in App.~\ref{app:A} and~\ref{app:duration}.

\section{Evaluation Protocol}\label{sec:method}

StrideWorld tests scene rigidity, single-action pace, and multi-action composition. The headline metrics of the three tests in Tab.~\ref{tab:main} are the Rigidity offset relative to the rigid counterpart, the pace ratio, and the \mComp, and the remaining columns locate failures.

\subsection{Scene rigidity}\label{sec:geometry}\label{sec:rigidity}
If a generated scene is rigid, the 3D points of the start view stay in place in every later frame. Using the ground-truth depth $z$ and intrinsics $\mathbf K$ of the start view, we lift feature pixels $\mathbf p_i$ to 3D anchors $\mathbf X_i=z(\mathbf p_i)\,\mathbf K^{-1}\mathbf p_i^{h}$ and split them alternately into disjoint pose anchors $\mathcal P$ and held-out anchors $\mathcal Q$, as illustrated in Fig.~\ref{fig:teaser}. In generated frame $t$, CoTracker3~\citep{cotracker3} tracks anchor $i$ to the position $\widetilde{\mathbf x}_{i,t}$, and for a camera $\mathbf T\in SE(3)$ we measure the angle between the tracked viewing ray and the ray toward the rigid position of the anchor in the camera frame,
\begin{equation}
\phi_{i,t}(\mathbf T)=\angle\big(\mathbf K^{-1}\widetilde{\mathbf x}_{i,t}^{h},\,\mathbf T^{-1}\mathbf X_i\big).
\label{eq:phi}
\end{equation}
The camera of frame $t$ is fitted to the pose anchors alone, and the \textbf{Rigidity offset ($^\circ$)} $e_t$ of the frame is the mean residual of the held-out anchors under this camera,
\begin{equation}
\begin{aligned}
\widehat{\mathbf T}_t&=\arg\min_{\mathbf T\in SE(3)}\textstyle\sum_{i\in\mathcal P_t}\phi_{i,t}(\mathbf T)^2,\\
e_t&=\frac{1}{|\mathcal Q_t|}\textstyle\sum_{i\in\mathcal Q_t}\phi_{i,t}(\widehat{\mathbf T}_t),
\end{aligned}
\label{eq:rigidity}
\end{equation}
where $\mathcal P_t\subseteq\mathcal P$ holds the RANSAC inliers of an EPnP initialization~\citep{pnp,fischler1981random} and $\mathcal Q_t\subseteq\mathcal Q$ the held-out anchors tracked in frame $t$. Because $\widehat{\mathbf T}_t$ is estimated from the generated frames, errors in following the command do not enter $e_t$. In Tab.~\ref{tab:main}, \textbf{Offset} is the median $e_t$ over route-video frames at matched image motion, since the offset grows with how far the anchors move in the image. \textbf{vs.\ rigid} divides the offset of twenty routes by that of their rigid counterpart, a rendering of the true scene along the same camera paths and in the same video format, so that a ratio of one means no excess over tracking noise (App.~\ref{app:C3} and~\ref{app:ceiling}).

\subsection{Single-action pace}\label{sec:single_action}\label{sec:pace}\label{sec:validity}
Holding a command should move the camera by the same amount at every step. Models share no unit of motion, so we measure each action in units of the motion it produces in its first steps. Let $\mathbf H_{a,k}\in SE(2)$ denote the planar pose of the recovered camera relative to the start view, its horizontal position and heading, after $k$ steps of a single action $a$. The \emph{stride} $\mathbf U_a\in SE(2)$ of the action averages the per-step motion of the first four steps in the Lie algebra $\mathfrak{se}(2)$,
\begin{equation}
\mathbf U_a=\operatorname{Exp}\Big(\frac{1}{4}\sum_{k=1}^{4}\frac{\operatorname{Log}\mathbf H_{a,k}}{k}\Big),
\label{eq:2}
\end{equation}
and the predicted pose after $k$ held steps is $\widehat{\mathbf H}_{a,k}=\mathbf U_a^k$. A stride is calibrated separately for each model, start view, and seed, and strides are never compared across start views, since a model's scale of motion may differ between scenes. Tab.~\ref{tab:main} reports three metrics for single actions. \textbf{Resp.\ (\%)} is the share of videos that move the camera in the commanded direction. The \textbf{pace ratio} divides the observed displacement or turn along the commanded axis at step $16$ by the predicted one, so values above and below $1$ indicate acceleration and slowdown. The \textbf{\mHold} takes the endpoint error at steps $12$ and $16$ relative to the predicted motion, $r$, and maps it to $s(r)=100/(1+r)$, which is $100$ for an exact prediction and $50$ when the error equals the predicted motion. Videos that do not respond or whose camera can no longer be recovered score zero (App.~\ref{app:D}).

\begin{figure*}[t]
    \centering
    \includegraphics[width=\textwidth]{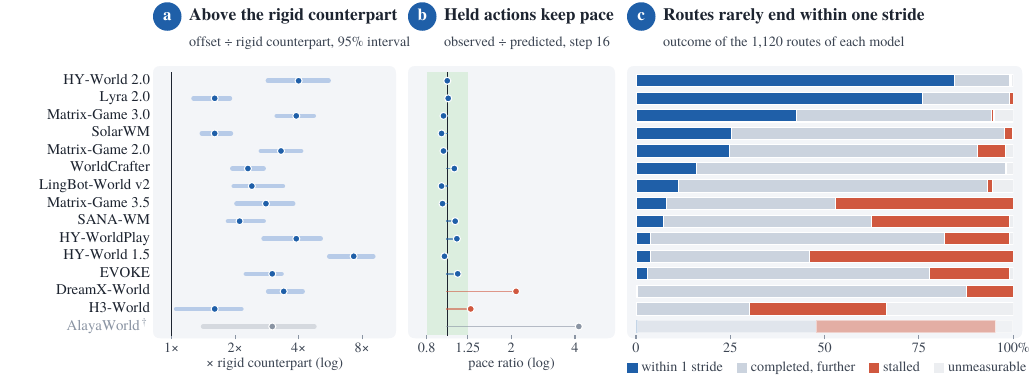}
    \caption{\textbf{The three consistency tests at a glance, one row per model.} (a)~Rigidity offset divided by that of each model's rigid counterpart, with $95\%$ intervals, where $1\times$ means no excess over tracking noise. (b)~Pace ratio after $16$ held steps, with the steady band of $0.8$ to $1.25$ shaded. (c)~Outcome of the $1{,}120$ composed routes of each model. Rows are sorted by Near, the share of routes ending within one stride. $^\dagger$Stride not reliably measurable.}
    \label{fig:findings}\label{fig:results}
\end{figure*}

\begin{table*}[t]
\centering
\caption{\textbf{Main results of \nmodels{} open world models}, grouped by control interface and ordered by \mComp. Resp.: videos that move in the commanded direction. Offset: median Rigidity offset over route frames at matched image motion. vs.\ rigid: the offset of twenty routes divided by that of their rigid counterpart, whose own offset differs across models (App.~\ref{app:ceiling}). Ratio: pace ratio at step $16$, red when outside the $0.8$--$1.25$ steady band under both Eq.~\eqref{eq:2} and equal-weight strides (App.~\ref{app:stridealt}). Stall: routes with a stalled step. End err.: median position error of completed routes. Swap gap: median distance between the ends of a completed swap pair. Near: completed routes ending within one stride of their prediction in position and heading. $^\dagger$First steps too small for a reliable stride, so stride-based metrics are grey and unranked. Bold and underlined: best and second-best among ranked models. MUSIQ~\citep{ke2021musiq} measures no-reference frame quality.}
\label{tab:main}\label{tab:3}
\begingroup
\footnotesize
\setlength{\tabcolsep}{2pt}
\renewcommand{\arraystretch}{1.08}
\begin{tabular*}{\textwidth}{@{\extracolsep{\fill}}lccccccccccc@{}}
\toprule
 & & \multicolumn{2}{c}{\cellcolor{fone}Rigidity} & \multicolumn{2}{c}{\cellcolor{ftwo}Pace} & \multicolumn{5}{c}{\cellcolor{fthree}Composition} & \\
\cmidrule(lr){3-4}\cmidrule(lr){5-6}\cmidrule(lr){7-11}
Model & Resp.\,(\%) & Offset\,($^\circ$)$\downarrow$ & vs.\,rigid$\downarrow$ & Ratio\,(${\to}1$) & Score$\uparrow$ & Stall\,(\%)$\downarrow$ & End err.\,(str.)$\downarrow$ & Swap gap\,(str.)$\downarrow$ & Near\,(\%)$\uparrow$ & Score$\uparrow$ & \textcolor{gray}{MUSIQ} \\
\midrule
\multicolumn{12}{@{}l}{\textit{Discrete keys}}\\[-1pt]
Matrix-Game 3.0 & 99.6 & 0.39 & 3.9 & 0.96 & 77.1 & 0.7 & 0.67 & 0.41 & 42.5 & 59.3 & \textcolor{gray}{72.1} \\
Matrix-Game 2.0 & 84.3 & 0.40 & 3.3 & 0.96 & 64.8 & 7.4 & 1.45 & 1.14 & 24.7 & 49.6 & \textcolor{gray}{68.5} \\
SANA-WM & 99.4 & 0.27 & \underline{2.1} & 1.09 & 76.2 & 36.5 & 2.47 & 1.95 & 7.2 & 28.3 & \textcolor{gray}{69.6} \\
DreamX-World & 99.0 & 0.38 & 3.4 & \textcolor{alertred}{2.11} & 46.4 & 12.3 & 5.22 & 4.92 & 0.5 & 24.2 & \textcolor{gray}{74.3} \\
HY-World 1.5 & 99.6 & 0.33 & 7.3 & 0.97 & 77.7 & 54.0 & 2.42 & 1.56 & 3.8 & 21.8 & \textcolor{gray}{71.6} \\
\midrule
\multicolumn{12}{@{}l}{\textit{Camera path}}\\[-1pt]
HY-World 2.0 & 99.3 & 0.35 & 4.0 & 1.00 & \underline{86.8} & \underline{0.3} & \textbf{0.30} & \textbf{0.10} & \textbf{84.3} & \textbf{83.1} & \textcolor{gray}{72.9} \\
Lyra 2.0 & 99.5 & \textbf{0.11} & \textbf{1.6} & 1.01 & \textbf{88.1} & 1.0 & \underline{0.51} & \underline{0.34} & \underline{75.9} & \underline{78.0} & \textcolor{gray}{75.5} \\
SolarWM & 98.8 & \underline{0.17} & \textbf{1.6} & 0.94 & 85.4 & 2.1 & 1.61 & 0.77 & 25.4 & 66.1 & \textcolor{gray}{73.0} \\
WorldCrafter & 99.4 & 0.33 & 2.3 & 1.08 & 80.2 & \textbf{0.1} & 2.38 & 0.88 & 16.0 & 56.8 & \textcolor{gray}{66.1} \\
LingBot-World v2 & 99.2 & 0.38 & 2.4 & 0.94 & 82.6 & 1.3 & 2.32 & 0.78 & 11.2 & 54.6 & \textcolor{gray}{72.2} \\
HY-WorldPlay & 99.8 & 0.38 & 3.9 & 1.11 & 74.8 & 17.1 & 3.44 & 1.71 & 3.8 & 38.4 & \textcolor{gray}{73.0} \\
EVOKE & 99.2 & 0.34 & 3.0 & 1.12 & 73.9 & 21.3 & 4.57 & 2.59 & 3.1 & 32.9 & \textcolor{gray}{69.6} \\
Matrix-Game 3.5 & 85.6 & 0.34 & 2.8 & 0.95 & 65.6 & 47.1 & 2.13 & 1.10 & 8.0 & 22.2 & \textcolor{gray}{72.3} \\
AlayaWorld$^\dagger$ & 75.5 & 0.26 & 3.0 & \textcolor{gray}{4.17} & \textcolor{gray}{19.8} & \textcolor{gray}{47.5} & \textcolor{gray}{50.57} & \textcolor{gray}{64.76} & \textcolor{gray}{0.4} & \textcolor{gray}{11.0} & \textcolor{gray}{71.5} \\
\midrule
\multicolumn{12}{@{}l}{\textit{Text prompt}}\\[-1pt]
H3-World & 91.9 & 0.32 & \textbf{1.6} & 1.29 & 24.9 & 36.1 & 9.02 & 4.52 & 0.2 & 7.6 & \textcolor{gray}{63.3} \\
\bottomrule
\end{tabular*}

\endgroup
\end{table*}

\subsection{Multi-action composition}\label{sec:composition}
Along a multi-action route $q=(a_1,\dots,a_n)$, each step should move the camera by the stride of its action, expressed in the camera frame at the start of that step. The predicted endpoint therefore chains the strides calibrated under the same start view and seed, $\widehat{\mathbf H}_q=\mathbf U_{a_1}\cdots\mathbf U_{a_n}$, and no parameter is fitted to the route itself. We evaluate endpoints at two levels, each route against its own prediction, $\mathbf E_q=\widehat{\mathbf H}_q^{-1}\mathbf H_q$, and the two routes of a pair $(q,q')$ against each other,
\begin{equation}
\mathbf E_{qq'}=\big(\widehat{\mathbf H}_q^{-1}\widehat{\mathbf H}_{q'}\big)^{-1}\big(\mathbf H_q^{-1}\mathbf H_{q'}\big),
\label{eq:4}
\end{equation}
which compares where the two routes end relative to each other with where their strides place them. Position and heading errors are divided by the median translation stride $b_t$ and the median turn stride $b_r$ of the start view and seed, so all errors are expressed in strides.

A step is \emph{stalled} if it moves or turns less than $10\%$ of its stride along the commanded axis, and a route is \emph{completed} if its first frame matches the start view, its camera is recovered at every step boundary, and none of its steps stalls. Tab.~\ref{tab:main} reports five composition metrics, of which the first three compare each route with its own prediction. \textbf{Stall (\%)} is the share of routes with at least one stalled step. \textbf{End err.\ (str.)} is the median position error of completed routes, and \textbf{Near (\%)} is the share of all routes that are completed and end within one stride of their prediction in both position and heading. The fourth compares the two routes of a swap pair with each other. \textbf{Swap gap (str.)} is the median distance between the ends of the two routes of a completed swap pair, which reflects the order of actions because both routes contain the same actions. The \textbf{\mComp} covers both levels and averages $s(r)$ of the position and heading errors over all routes and route pairs, with incomplete ones scoring zero, so it combines completion and endpoint accuracy in a single number (App.~\ref{app:D}).

\section{Results}\label{sec:results}

We evaluate the \nmodels{} open models on all \nvideos{} videos. Tab.~\ref{tab:main} reports every metric for each model, and Fig.~\ref{fig:findings} places the headline metric of each test side by side. Fourteen models are ranked, since the first steps of AlayaWorld are too small for a reliable stride. The first three subsections take the tests in turn, and Sec.~\ref{sec:validation} checks the measurements against known camera motion and human judgments.

\subsection{Generated scenes deform as the camera moves}\label{sec:res_f1}
Most models deform the generated scene to some degree as the camera moves. The measured offset also contains tracking noise, which depends on a model's resolution, frame rate, and compression. To separate the two, we re-render the true scene along the camera paths of twenty routes of each model in the same video format, which gives a rigid counterpart of these videos, and divide the offset of each model by that of its counterpart on the same frames. Every model deviates $1.6$ to $7.3$ times more than its rigid counterpart, as shown in Fig.~\ref{fig:findings}a, so tracking noise alone does not account for the excess.

This excess has two possible sources, a scene that deforms over time and a scene that stays rigid at a wrong depth. Refitting the depth of each held-out anchor and the camera freely removes the second. The ratio of Lyra 2.0 then falls to one, whereas the other fourteen models stay at $1.2\times$ or above, so the excess of Lyra 2.0 comes mostly from a fixed depth error. It conditions each new view on an explicit 3D cache built from estimated depth~\citep{lyra2}, so the surfaces it has already shown keep the same geometry, which suggests that a 3D representation mitigates deformation.

Preserving the scene does not make a model follow commands. The Rigidity ranking of the models shows no relation to their \mComp\ ranking, with Spearman correlations of $-0.01$ for the offset and $-0.07$ for the counterpart ratio (App.~\ref{app:ceiling} and~\ref{app:f1speed}).

\begin{figure*}[t]
    \centering
    \includegraphics[width=\textwidth]{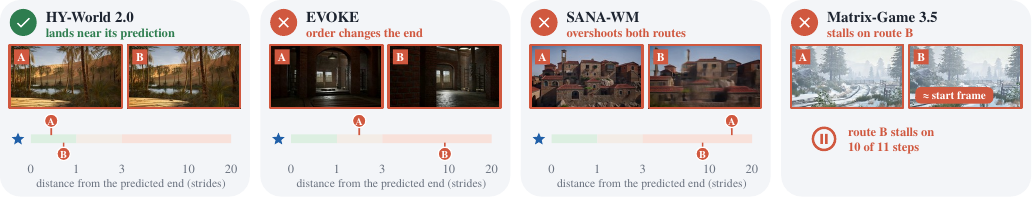}
    \caption{\textbf{One swap pair from each of four models in different scenes.} Routes A and B reorder the same translations, so their predicted ends, marked by $\bigstar$, nearly coincide. Each card shows the last generated frames of both routes and the distance of each end from its predicted end in strides, on an axis logarithmic in one plus the distance, green within one stride. Matrix-Game~3.5 completes neither route.}
    \label{fig:gallery}
\end{figure*}

\subsection{Held actions keep a steady pace}\label{sec:res_f2}
When a single action is held for $16$ steps, most models move in the commanded direction and keep the pace set by their first four steps. Twelve of the fourteen ranked models have a median pace ratio within the steady band of $0.8$ to $1.25$, as shown in Fig.~\ref{fig:findings}b, and the text-driven H3-World is only slightly faster. The same pattern holds with equally weighted strides (App.~\ref{app:stridealt}).

Among the ranked models, only DreamX-World changes pace strongly. It accelerates over the first steps of a hold, so its first four steps no longer predict where a held command leads. AlayaWorld moves slowly during its first steps, which makes its stride hard to measure, and its stride-based metrics are therefore shown in grey, left unranked, and marked by $^\dagger$ in Tab.~\ref{tab:main} (App.~\ref{app:res_f2}).

\subsection{Composed routes stall and drift}\label{sec:res_f3}
Once actions are combined into routes, the strides of most models no longer predict where their routes end, as shown in Fig.~\ref{fig:findings}c. Only HY-World 2.0 and Lyra 2.0 bring more than three quarters of their routes within one stride of the predicted endpoint, whereas most other models do so for fewer than a quarter. Both take camera poses as input and scale translations by the estimated depth of the start view, so the same command moves the camera by a consistent fraction of the scene. Tying commands to scene depth thus appears to help strides add up along a route. The interface alone does not explain this gap, as Near ranges from $3.1\%$ to $84.3\%$ across the ranked camera-pose models and the key-driven Matrix-Game 3.0 outperforms all of them except \mbox{HY-World 2.0} and Lyra 2.0.

The routes of the other models fail in two main ways, illustrated in Fig.~\ref{fig:gallery}. The first is \textbf{mid-route stalls}, where the camera stops for a step partway through a route, most often when a new action starts. Stalls affect more than a third of the routes in four ranked models, as listed in Tab.~\ref{tab:main}, and account for most incomplete routes.

The second is \textbf{endpoint drift}, where even completed routes end well away from their prediction. The median error is above one stride in eleven of the fourteen ranked models. In four ranked models, the same actions in a different order end about two to three times farther apart than repeated generation predicts. Regenerating the routes with two further seeds shows that this drift is systematic, as a route's end error is a median $1.8$ to $3.7$ times the spread of its ends across seeds. Part of the error arises at action switches, where some models carry the previous action over and others start the new action with only about half a stride, whereas the leading models keep the same step before and after the switch (App.~\ref{app:res_f3}).

Existing control metrics miss much of this, as direction checks miss both failure modes and direction accuracy shows no clear relation to a model's \mComp. A trajectory error with a fitted scale also misses many failed routes, even when it is given the stride-predicted path as its reference. After fitting one scale, about a fifth of the stalled routes still have an error below $10\%$ of the path, and so do about a quarter of the routes that end more than two strides from their prediction. The error also does not show where along a route the failure occurs (App.~\ref{app:existing}).

\begin{figure*}[t]
\centering
\includegraphics[width=\textwidth]{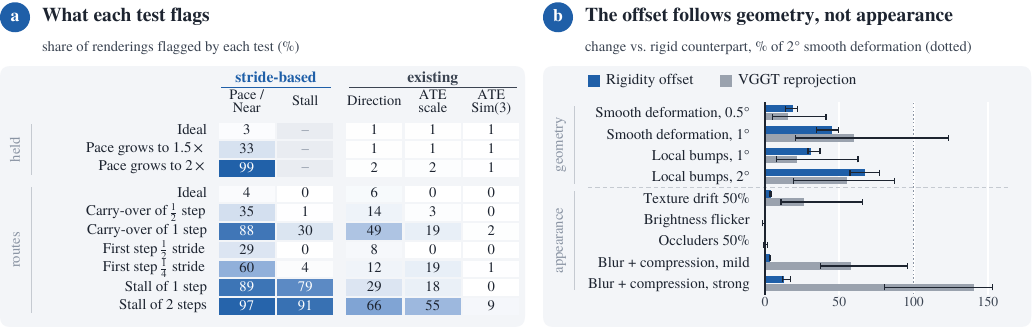}
\caption{\textbf{Which test responds to which known change, pooled over three formats.} (a)~Share of held videos and of routes flagged by each test at its default threshold when one property of an ideal rendering is changed. Pace flags a held video whose pace ratio leaves the steady band, Near a route that does not end within one stride, and Stall a route with a stalled step. Ideal rows give the false alarms. (b)~Median change of the Rigidity offset (blue) and of the VGGT reprojection error (grey) against the rigid counterpart, as a share of the change at $2^\circ$ of smooth deformation, with $95\%$ bootstrap intervals. Flag rules and settings are given in App.~\ref{app:perturb}.}
\label{fig:perturb_summary}\label{fig:perturb_ga}
\end{figure*}

\subsection{Measurements are accurate and robust}\label{sec:validation}
On videos rendered in UE5 along known camera paths, the recovered cameras deviate from the true ones by a median of $1.5\%$ in translation and $0.3\%$ in rotation. Adding blur, compression, flicker, or moderate wobble keeps these median errors within $4.0\%$ and $2.1\%$. On the generated videos, other calibration lengths, stall thresholds, and camera-recovery anchors leave all findings unchanged, and the route findings hold when the commands of three models are halved or doubled (App.~\ref{app:E},~\ref{app:H}, and~\ref{app:magsens}).

We then change one property of a rendered video by a known amount to test whether each metric responds to the failure it targets, as shown in Fig.~\ref{fig:perturb_summary}. Deformation injected into the rigid counterparts raises the Rigidity offset linearly, while texture drift, flicker, and occluders leave it near its level on the rigid counterparts. The reprojection error of a feed-forward reconstruction, by contrast, responds to texture drift and to blur with compression at least as strongly as to deformation. On the generated videos, it separates only three of the fifteen models from their rigid counterparts, whereas the Rigidity offset separates all fifteen.

On the action side, Fig.~\ref{fig:perturb_summary}a applies every test at its default threshold. When each test is instead set to flag $5\%$ of the ideal held videos, the Pace test flags $94\%$ of the held videos whose pace grows to $1.5$ times. A scale-fitted trajectory error flags only $43\%$ of them, because its scale absorbs part of the change. Scene deformation also shifts the recovered camera, which at the level seen in generated videos changes the Near decision of $8\%$ to $20\%$ of routes but leaves the model ranking unchanged (App.~\ref{app:perturb}).

The scores also agree with the judgments of ten non-author raters blind to the models and metrics. They rated $160$ randomly sampled videos for whether the camera follows the commands, with a Fleiss' $\kappa$ of $0.87$, and the score of each video separates those judged good with an AUC of $0.88$. On $140$ completed swap pairs, the pair score separates those judged to end at the same place with an AUC of $0.83$, against at most $0.74$ for scores with a fitted scale. For two videos of the same route whose Rigidity offsets differ at least twofold, the majority of raters chose the one with the smaller offset as more stable in all $39$ untied pairs (App.~\ref{app:human}).

\section{Fine-Tuning on Rendered Camera Paths}\label{sec:sft}

Along composed routes, several models often stall and most completed routes end away from their predictions, as shown in Sec.~\ref{sec:res_f3}. Since every command step in a rendered clip moves the camera by an exact displacement or turn, we test whether fine-tuning on a few such clips reduces these failures. We render $112$ training clips in procedural Blender scenes that share no assets or routes with StrideWorld, and fine-tune LoRA adapters on the non-distilled base models of HY-World 1.5 and SANA-WM.

On the full benchmark of $3{,}640$ videos, both fine-tuned models improve over their base models on every consistency metric without lowering image-quality scores. Stalled routes fall from $72.0\%$ to $11.2\%$ for HY-World 1.5 and from $57.9\%$ to $23.1\%$ for SANA-WM. The \mComp\ of both models more than doubles and exceeds that of the released models in Tab.~\ref{tab:main}, and the Rigidity offset also falls, by about two fifths for SANA-WM. For HY-World 1.5, we further examine what these gains depend on. They hold on the real ETH3D views although all training clips are rendered. We then randomly perturb the step lengths of the training clips, so that every command step still moves the camera but by a varying amount. Fine-tuning on these clips retains about four fifths of the stall reduction, and exact step lengths add the rest. Tab.~\ref{tab:sft} and App.~\ref{app:I} give the full results.

\raggedbottom
\section{Conclusion}\label{sec:conclusion}

StrideWorld tests interactive video world models against references that are fixed before the motion they judge. It compares the scene with the ground-truth geometry of the start view, and each action with the stride set by the model's own first steps. Across \nmodels{} open models, nearly all deform their scenes as the camera moves. Most keep a steady pace under a single held action, but their composed routes often stall partway or end away from the positions predicted by their own strides. Two design choices stand out among the stronger models. The one that conditions on an explicit 3D cache preserves the scene best, and the two that scale commands by the estimated scene depth compose routes best. Fine-tuning two models on a few rendered clips with exact camera paths also removes most of their mid-route stalls. Extending StrideWorld to dynamic scenes and full 6-DoF trajectories is a natural next step (App.~\ref{app:limitations}).

{
    \small
    \bibliographystyle{ieeenat_fullname}
    \bibliography{main}

\begin{thebibliography}{66}
\providecommand{\natexlab}[1]{#1}
\providecommand{\url}[1]{\texttt{#1}}
\expandafter\ifx\csname urlstyle\endcsname\relax
  \providecommand{\doi}[1]{doi: #1}\else
  \providecommand{\doi}{doi: \begingroup \urlstyle{rm}\Url}\fi

\bibitem[Agarwal et~al.(2025)Agarwal, Ali, Bala, Balaji, Barker, Cai, Chattopadhyay, Chen, Cui, Ding, et~al.]{nvidia2025cosmos}
Niket Agarwal, Arslan Ali, Maciej Bala, Yogesh Balaji, Erik Barker, Tiffany Cai, Prithvijit Chattopadhyay, Yongxin Chen, Yin Cui, Yifan Ding, et~al.
\newblock Cosmos world foundation model platform for physical {AI}.
\newblock \emph{arXiv preprint arXiv:2501.03575}, 2025.

\bibitem[Asim et~al.(2025)Asim, Wewer, Wimmer, Schiele, and Lenssen]{met3r}
Mohammad Asim, Christopher Wewer, Thomas Wimmer, Bernt Schiele, and Jan~Eric Lenssen.
\newblock {MEt3R}: Measuring multi-view consistency in generated images.
\newblock In \emph{2025 IEEE/CVF Conference on Computer Vision and Pattern Recognition (CVPR)}, pages 6034--6044. IEEE, 2025.

\bibitem[Bansal et~al.(2025)Bansal, Lin, Xie, Zong, Yarom, Bitton, Jiang, Sun, Chang, and Grover]{videophy}
Hritik Bansal, Zongyu Lin, Tianyi Xie, Zeshun Zong, Michal Yarom, Yonatan Bitton, Chenfanfu Jiang, Yizhou Sun, Kai-Wei Chang, and Aditya Grover.
\newblock {VideoPhy}: Evaluating physical commonsense for video generation.
\newblock In \emph{International Conference on Learning Representations}, pages 102075--102121, 2025.

\bibitem[Bansal et~al.(2026)Bansal, Peng, Bitton, Goldenberg, Grover, and Chang]{videophy2}
Hritik Bansal, Clark Peng, Yonatan Bitton, Roman Goldenberg, Aditya Grover, and Kai-Wei Chang.
\newblock {VideoPhy-2}: A challenging action-centric physical commonsense evaluation in video generation.
\newblock In \emph{International Conference on Learning Representations}, pages 118456--118470, 2026.

\bibitem[Brooks et~al.(2024)Brooks, Peebles, Holmes, DePue, Guo, Jing, Schnurr, Taylor, Luhman, Luhman, et~al.]{brooks2024sora}
Tim Brooks, Bill Peebles, Connor Holmes, Will DePue, Yufei Guo, Li Jing, David Schnurr, Joe Taylor, Troy Luhman, Eric Luhman, et~al.
\newblock Video generation models as world simulators.
\newblock OpenAI Blog, 2024.

\bibitem[Bruce et~al.(2024)Bruce, Dennis, Edwards, Parker-Holder, Shi, Hughes, Lai, Mavalankar, Steigerwald, Apps, et~al.]{genie}
Jake Bruce, Michael~D Dennis, Ashley Edwards, Jack Parker-Holder, Yuge Shi, Edward Hughes, Matthew Lai, Aditi Mavalankar, Richie Steigerwald, Chris Apps, et~al.
\newblock {Genie}: Generative interactive environments.
\newblock In \emph{Forty-first International Conference on Machine Learning}, 2024.

\bibitem[Chen et~al.(2026)Chen, Wang, Lin, Yang, and Jin]{h3world}
Danze Chen, Zeqing Wang, Ziyue Lin, Xingyi Yang, and Yeying Jin.
\newblock {H3-World}: Turning language understanding into world control.
\newblock \emph{arXiv preprint arXiv:2609.01560}, 2026.

\bibitem[Ding et~al.(2026)Ding, Chen, Cai, Xu, Wang, Lu, Li, Chen, Gao, Tao, et~al.]{playworld}
Kaixin Ding, Xi Chen, Minghong Cai, Zhiyuan Xu, Yiyang Wang, Yuxiang Lu, Junyi Li, Shuyang Chen, Yuan Gao, Xin Tao, et~al.
\newblock {PlayWorld}: Benchmarking world models with agent players over long-horizon objectives.
\newblock \emph{arXiv preprint arXiv:2608.13552}, 2026.

\bibitem[Duan et~al.(2025)Duan, Yu, Chen, Fei-Fei, and Wu]{worldscore}
Haoyi Duan, Hong-Xing Yu, Sirui Chen, Li Fei-Fei, and Jiajun Wu.
\newblock {WorldScore}: A unified evaluation benchmark for world generation.
\newblock In \emph{Proceedings of the IEEE/CVF International Conference on Computer Vision}, pages 27713--27724, 2025.

\bibitem[{Epic Games}(2022)]{unrealengine5}
{Epic Games}.
\newblock {Unreal Engine 5}, 2022.
\newblock Computer software.

\bibitem[Fang et~al.(2026)Fang, Lei, Wan, Wang, Huang, Xu, Zhao, Zhang, Gao, Chen, et~al.]{iworld}
Jianjie Fang, Yingshan Lei, Qin Wan, Ziyou Wang, Yuchao Huang, Yongyan Xu, Baining Zhao, Weichen Zhang, Chen Gao, Xinlei Chen, et~al.
\newblock {iWorld-Bench}: A benchmark for interactive world models with a unified action generation framework.
\newblock \emph{arXiv preprint arXiv:2605.03941}, 2026.

\bibitem[Fischler and Bolles(1981)]{fischler1981random}
Martin~A Fischler and Robert~C Bolles.
\newblock Random sample consensus: a paradigm for model fitting with applications to image analysis and automated cartography.
\newblock \emph{Communications of the ACM}, 24\penalty0 (6):\penalty0 381--395, 1981.

\bibitem[Gao et~al.(2026)Gao, Wang, Zhu, Chen, Liu, Bai, Wang, Yuan, Wang, Lu, et~al.]{lingbotv2}
Zelin Gao, Qiuyu Wang, Jiapeng Zhu, Jingye Chen, Zichen Liu, Qingyan Bai, Jiahao Wang, Yufeng Yuan, Hanlin Wang, Yichong Lu, et~al.
\newblock Infinite worlds with versatile interactions.
\newblock \emph{arXiv preprint arXiv:2607.07534}, 2026.

\bibitem[Gu et~al.(2026)Gu, Yuan, Wu, Du, Liu, Pang, Zhang, Lu, Zhong, Zhao, et~al.]{worldcycle}
Bohai Gu, Yueyang Yuan, Taiyi Wu, Dazhao Du, Jian Liu, Xiaoyi Pang, Jie Zhang, Xiaocheng Lu, Haobin Zhong, Xiaotong Zhao, et~al.
\newblock {WorldCycle}: Self-verifiable reinforcement learning for long-horizon video world models.
\newblock \emph{arXiv preprint arXiv:2608.04964}, 2026.

\bibitem[Ha and Schmidhuber(2018)]{ha2018recurrent}
David Ha and J{\"u}rgen Schmidhuber.
\newblock Recurrent world models facilitate policy evolution.
\newblock \emph{Advances in Neural Information Processing Systems}, 31, 2018.

\bibitem[Hafner et~al.(2023)Hafner, Pasukonis, Ba, and Lillicrap]{hafner2023mastering}
Danijar Hafner, Jurgis Pasukonis, Jimmy Ba, and Timothy Lillicrap.
\newblock Mastering diverse domains through world models.
\newblock \emph{arXiv preprint arXiv:2301.04104}, 2023.

\bibitem[He et~al.(2025{\natexlab{a}})He, Xu, Guo, Wetzstein, Dai, Li, and Yang]{cameractrl}
Hao He, Yinghao Xu, Yuwei Guo, Gordon Wetzstein, Bo Dai, Hongsheng Li, and Ceyuan Yang.
\newblock {CameraCtrl}: Enabling camera control for video diffusion models.
\newblock In \emph{International Conference on Learning Representations}, pages 100433--100464, 2025{\natexlab{a}}.

\bibitem[He et~al.(2025{\natexlab{b}})He, Peng, Liu, Wang, Zhang, Cui, Kang, Jiang, An, Ren, et~al.]{matrix2}
Xianglong He, Chunli Peng, Zexiang Liu, Boyang Wang, Yifan Zhang, Qi Cui, Fei Kang, Biao Jiang, Mengyin An, Yangyang Ren, et~al.
\newblock {Matrix-Game 2.0}: An open-source, real-time, and streaming interactive world model.
\newblock \emph{arXiv preprint arXiv:2508.13009}, 2025{\natexlab{b}}.

\bibitem[Hu et~al.(2026)Hu, Huang, Shankarampeta, Dai, Wu, Qian, Liu, Srinivasan, Huo, Lin, et~al.]{wmeval}
Lanxiang Hu, Yixin Huang, Abhilash Shankarampeta, Zilin Dai, Yuyuan Wu, Kun Qian, Junze Liu, Arvind Srinivasan, Mingjia Huo, Will Lin, et~al.
\newblock {WMEval}: Evaluating interactive world models grounded on physical simulations.
\newblock \emph{arXiv preprint arXiv:2610.33894}, 2026.

\bibitem[Huang et~al.(2026)Huang, Fang, Qian, Kong, Zhao, Huang, Du, Zhang, Cui, Gu, et~al.]{solarwm}
Junchao Huang, Guian Fang, Shengju Qian, Xianghao Kong, Zhuoran Zhao, Wei Huang, Yihua Du, Zixin Zhang, Justin Cui, Yuchao Gu, et~al.
\newblock {SolarWM}: Open data and scalable training for long-horizon video world models.
\newblock \emph{arXiv preprint arXiv:2609.02886}, 2026.

\bibitem[Huang et~al.(2024)Huang, He, Yu, Zhang, Si, Jiang, Zhang, Wu, Jin, Chanpaisit, et~al.]{vbench}
Ziqi Huang, Yinan He, Jiashuo Yu, Fan Zhang, Chenyang Si, Yuming Jiang, Yuanhan Zhang, Tianxing Wu, Qingyang Jin, Nattapol Chanpaisit, et~al.
\newblock {VBench}: Comprehensive benchmark suite for video generative models.
\newblock In \emph{2024 IEEE/CVF Conference on Computer Vision and Pattern Recognition (CVPR)}, pages 21807--21818. IEEE, 2024.

\bibitem[Huang et~al.(2025)Huang, Zhang, Xu, He, Yu, Dong, Ma, Chanpaisit, Si, Jiang, et~al.]{vbench++}
Ziqi Huang, Fan Zhang, Xiaojie Xu, Yinan He, Jiashuo Yu, Ziyue Dong, Qianli Ma, Nattapol Chanpaisit, Chenyang Si, Yuming Jiang, et~al.
\newblock {VBench++}: Comprehensive and versatile benchmark suite for video generative models.
\newblock \emph{IEEE Transactions on Pattern Analysis and Machine Intelligence}, 2025.

\bibitem[Karaev et~al.(2025)Karaev, Makarov, Wang, Neverova, Vedaldi, and Rupprecht]{cotracker3}
Nikita Karaev, Yuri Makarov, Jianyuan Wang, Natalia Neverova, Andrea Vedaldi, and Christian Rupprecht.
\newblock {CoTracker3}: Simpler and better point tracking by pseudo-labelling real videos.
\newblock In \emph{Proceedings of the IEEE/CVF International Conference on Computer Vision}, pages 6013--6022, 2025.

\bibitem[Ke et~al.(2021)Ke, Wang, Wang, Milanfar, and Yang]{ke2021musiq}
Junjie Ke, Qifei Wang, Yilin Wang, Peyman Milanfar, and Feng Yang.
\newblock {MUSIQ}: Multi-scale image quality transformer.
\newblock In \emph{2021 IEEE/CVF International Conference on Computer Vision (ICCV)}, pages 5128--5137. IEEE, 2021.

\bibitem[Kwon et~al.(2025)Kwon, Kim, Go, and Baek]{kwon2025stable}
Soonwoo Kwon, Jin-Young Kim, Hyojun Go, and Kyungjune Baek.
\newblock Toward stable world models: Measuring and addressing world instability in generative environments.
\newblock \emph{arXiv preprint arXiv:2503.08122}, 2025.

\bibitem[Lepetit et~al.(2009)Lepetit, Moreno-Noguer, and Fua]{pnp}
Vincent Lepetit, Francesc Moreno-Noguer, and Pascal Fua.
\newblock {EPnP}: An accurate {O(n)} solution to the {PnP} problem.
\newblock \emph{International Journal of Computer Vision}, 81\penalty0 (2):\penalty0 155--166, 2009.

\bibitem[Li et~al.(2025{\natexlab{a}})Li, Fang, Chen, Yang, Cao, Wong, Luo, Wang, Yin, Gonzalez, et~al.]{worldmodelbench}
Dacheng Li, Yunhao Fang, Yukang Chen, Shuo Yang, Shiyi Cao, Justin Wong, Michael Luo, Xiaolong Wang, Hongxu Yin, Joseph~E Gonzalez, et~al.
\newblock {WorldModelBench}: Judging video generation models as world models.
\newblock \emph{arXiv preprint arXiv:2502.20694}, 2025{\natexlab{a}}.

\bibitem[Li et~al.(2025{\natexlab{b}})Li, Tang, Xu, Wu, Zhou, Shao, Yu, Cao, and Lu]{gamecraft}
Jiaqi Li, Junshu Tang, Zhiyong Xu, Longhuang Wu, Yuan Zhou, Shuai Shao, Tianbao Yu, Zhiguo Cao, and Qinglin Lu.
\newblock {Hunyuan-GameCraft}: High-dynamic interactive game video generation with hybrid history condition.
\newblock \emph{arXiv preprint arXiv:2506.17201}, 2025{\natexlab{b}}.

\bibitem[Lian et~al.(2025)Lian, Cai, Liang, and Liu]{loopnav}
Kewei Lian, Shaofei Cai, Yitao Liang, and Anji Liu.
\newblock {LoopNav}: Benchmarking spatial consistency in world models.
\newblock \emph{arXiv preprint arXiv:2505.22976}, 2025.

\bibitem[Lin et~al.(2025)Lin, Chen, Liew, Chen, Li, Shi, Feng, and Kang]{lin2025depth}
Haotong Lin, Sili Chen, Junhao Liew, Donny~Y Chen, Zhenyu Li, Guang Shi, Jiashi Feng, and Bingyi Kang.
\newblock {Depth Anything 3}: Recovering the visual space from any views.
\newblock \emph{arXiv preprint arXiv:2511.10647}, 2025.

\bibitem[Liu et~al.(2024)Liu, Cun, Liu, Wang, Zhang, Chen, Liu, Zeng, Chan, and Shan]{evalcrafter}
Yaofang Liu, Xiaodong Cun, Xuebo Liu, Xintao Wang, Yong Zhang, Haoxin Chen, Yang Liu, Tieyong Zeng, Raymond Chan, and Ying Shan.
\newblock {EvalCrafter}: Benchmarking and evaluating large video generation models.
\newblock In \emph{2024 IEEE/CVF Conference on Computer Vision and Pattern Recognition (CVPR)}, pages 22139--22149. IEEE, 2024.

\bibitem[Lu et~al.(2026)Lu, Fang, Lim, Wu, Wang, Chen, Qian, Liao, Chow, Kong, et~al.]{orbia}
Dongyue Lu, Guian Fang, Hur Lim, Yonggang Wu, Yanrong Wang, Hongshan Chen, Shengju Qian, Yue Liao, Wei Chow, Lingdong Kong, et~al.
\newblock {Orbia}: Do generated videos form a consistent {3D} world?
\newblock \emph{arXiv preprint arXiv:2610.20693}, 2026.

\bibitem[Parker-Holder and Fruchter(2025)]{genie3}
Jack Parker-Holder and Shlomi Fruchter.
\newblock {Genie 3}: A new frontier for world models.
\newblock Google DeepMind Blog, 2025.

\bibitem[Parker-Holder et~al.(2024)Parker-Holder, Ball, Bruce, Dasagi, Holsheimer, Kaplanis, Moufarek, Scully, Shar, Shi, et~al.]{genie2}
Jack Parker-Holder, Philip Ball, Jake Bruce, Vibhavari Dasagi, Kristian Holsheimer, Christos Kaplanis, Alexandre Moufarek, Guy Scully, Jeremy Shar, Jimmy Shi, et~al.
\newblock {Genie 2}: A large-scale foundation world model.
\newblock Google DeepMind Blog, 2024.

\bibitem[Paul et~al.(2026)Paul, Kaushik, and Yuille]{paul2026can}
Soumava Paul, Prakhar Kaushik, and Alan Yuille.
\newblock Can these views be one scene? {E}valuating multiview {3D} consistency when {3D} foundation models hallucinate.
\newblock \emph{arXiv preprint arXiv:2605.18754}, 2026.

\bibitem[Qian et~al.(2026)Qian, Wang, Zhang, Zou, Yu, Li, Liu, Li, Kang, Huang, et~al.]{matrix35}
Runjia Qian, Zile Wang, Jihai Zhang, Kai Zou, Wei Yu, Jiaxing Li, Zexiang Liu, Yaokun Li, Fei Kang, Kaichen Huang, et~al.
\newblock {Matrix-Game 3.5}: Enhancing real-time streaming interactive world models with patch memory.
\newblock \emph{arXiv preprint arXiv:2608.29910}, 2026.

\bibitem[Sch{\"o}ps et~al.(2017)Sch{\"o}ps, Sch{\"o}nberger, Galliani, Sattler, Schindler, Pollefeys, and Geiger]{eth3d}
Thomas Sch{\"o}ps, Johannes~L Sch{\"o}nberger, Silvano Galliani, Torsten Sattler, Konrad Schindler, Marc Pollefeys, and Andreas Geiger.
\newblock A multi-view stereo benchmark with high-resolution images and multi-camera videos.
\newblock In \emph{Proceedings of the IEEE Conference on Computer Vision and Pattern Recognition}, pages 3260--3269, 2017.

\bibitem[Shen et~al.(2026)Shen, Bahmani, He, Srinivasan, Cao, Ren, Li, Wang, Sharp, Gojcic, et~al.]{lyra2}
Tianchang Shen, Sherwin Bahmani, Kai He, Sangeetha~Grama Srinivasan, Tianshi Cao, Jiawei Ren, Ruilong Li, Zian Wang, Nicholas Sharp, Zan Gojcic, et~al.
\newblock {Lyra 2.0}: Explorable generative {3D} worlds.
\newblock \emph{arXiv preprint arXiv:2604.13036}, 2026.

\bibitem[Sun et~al.(2025)Sun, Zhang, Wang, Wu, Wang, Wang, Wang, Zhang, Wang, and Guo]{worldplay}
Wenqiang Sun, Haiyu Zhang, Haoyuan Wang, Junta Wu, Zehan Wang, Zhenwei Wang, Yunhong Wang, Jun Zhang, Tengfei Wang, and Chunchao Guo.
\newblock {WorldPlay}: Towards long-term geometric consistency for real-time interactive world modeling.
\newblock \emph{arXiv preprint arXiv:2512.14614}, 2025.

\bibitem[Team et~al.(2026{\natexlab{a}})Team, Zhang, Li, Zhan, Ge, Yin, Tan, He, Fan, Liu, et~al.]{alayaworld}
AlayaWorld Team, Kaipeng Zhang, Chuanhao Li, Yifan Zhan, Yongtao Ge, Yuanyang Yin, Jiaming Tan, Kang He, Liaoyuan Fan, Ruicong Liu, et~al.
\newblock {AlayaWorld}: Long-horizon and playable video world generation.
\newblock \emph{arXiv preprint arXiv:2607.06291}, 2026{\natexlab{a}}.

\bibitem[Team et~al.(2026{\natexlab{b}})Team, Bai, Chen, Chu, Dang, Dou, Gao, Gu, Hong, Lei, et~al.]{dreamxworld}
DreamX Team, Yancheng Bai, Rui Chen, Xiangxiang Chu, Rujing Dang, Hao Dou, Bingjie Gao, Qiwen Gu, Siyu Hong, Jiachen Lei, et~al.
\newblock {DreamX-World 1.0}: A general-purpose interactive world model.
\newblock \emph{arXiv preprint arXiv:2606.16993}, 2026{\natexlab{b}}.

\bibitem[Team et~al.(2025)Team, Wang, Liu, Wu, Gu, Wang, Zuo, Huang, Li, Zhang, et~al.]{hunyuanworld}
HunyuanWorld Team, Zhenwei Wang, Yuhao Liu, Junta Wu, Zixiao Gu, Haoyuan Wang, Xuhui Zuo, Tianyu Huang, Wenhuan Li, Sheng Zhang, et~al.
\newblock {HunyuanWorld 1.0}: Generating immersive, explorable, and interactive {3D} worlds from words or pixels.
\newblock \emph{arXiv preprint arXiv:2507.21809}, 2025.

\bibitem[Team(2026)]{streameval}
StreamEval Team.
\newblock {StreamEval}: Evaluating interactive streaming video generation models.
\newblock \emph{arXiv preprint arXiv:2610.28402}, 2026.

\bibitem[{Team HY-World} et~al.(2026){Team HY-World}, Cao, Zuo, Wang, Zhang, Wu, Liu, Gong, Liu, Yuan, et~al.]{HY-World2}
{Team HY-World}, Chenjie Cao, Xuhui Zuo, Zhenwei Wang, Yisu Zhang, Junta Wu, Zhenyang Liu, Yuning Gong, Yang Liu, Bo Yuan, et~al.
\newblock {HY-World 2.0}: A multi-modal world model for reconstructing, generating, and simulating {3D} worlds.
\newblock \emph{arXiv preprint arXiv:2604.14268}, 2026.

\bibitem[Valevski et~al.(2025)Valevski, Leviathan, Arar, and Fruchter]{gamengen}
Dani Valevski, Yaniv Leviathan, Moab Arar, and Shlomi Fruchter.
\newblock Diffusion models are real-time game engines.
\newblock In \emph{International Conference on Learning Representations}, pages 73754--73776, 2025.

\bibitem[Wan et~al.(2025)Wan, Wang, Ai, Wen, Mao, Xie, Chen, Yu, Zhao, Yang, et~al.]{wan2025}
Team Wan, Ang Wang, Baole Ai, Bin Wen, Chaojie Mao, Chen-Wei Xie, Di Chen, Feiwu Yu, Haiming Zhao, Jianxiao Yang, et~al.
\newblock {Wan}: Open and advanced large-scale video generative models.
\newblock \emph{arXiv preprint arXiv:2503.20314}, 2025.

\bibitem[Wang et~al.(2025)Wang, Chen, Karaev, Vedaldi, Rupprecht, and Novotny]{vggt}
Jianyuan Wang, Minghao Chen, Nikita Karaev, Andrea Vedaldi, Christian Rupprecht, and David Novotny.
\newblock {VGGT}: Visual geometry grounded transformer.
\newblock In \emph{2025 IEEE/CVF Conference on Computer Vision and Pattern Recognition (CVPR)}, pages 5294--5306. IEEE, 2025.

\bibitem[Wang et~al.(2026{\natexlab{a}})Wang, Peng, Dong, Yao, Wang, Huang, Zhang, Li, Dou, Qu, et~al.]{worldprobe}
Minqi Wang, Jie Peng, Hejun Dong, Ruchang Yao, Puyi Wang, Siyuan Huang, Jihai Zhang, Yafu Li, Qi Dou, Xiaoye Qu, et~al.
\newblock {WorldProbe}: When video world models fall apart under long-horizon spatial exploration.
\newblock \emph{arXiv preprint arXiv:2610.39721}, 2026{\natexlab{a}}.

\bibitem[Wang et~al.(2026{\natexlab{b}})Wang, Liu, Li, Huang, Xu, Kang, An, Wang, Jiang, Wei, et~al.]{matrix3}
Zile Wang, Zexiang Liu, Jiaxing Li, Kaichen Huang, Baixin Xu, Fei Kang, Mengyin An, Peiyu Wang, Biao Jiang, Yichen Wei, et~al.
\newblock {Matrix-Game 3.0}: Real-time and streaming interactive world model with long-horizon memory.
\newblock \emph{arXiv preprint arXiv:2604.08995}, 2026{\natexlab{b}}.

\bibitem[Wang et~al.(2026{\natexlab{c}})Wang, Wang, Zhang, Zuo, Wu, Wang, Sun, Wang, Cao, Zhao, et~al.]{worldcompass}
Zehan Wang, Tengfei Wang, Haiyu Zhang, Xuhui Zuo, Junta Wu, Haoyuan Wang, Wenqiang Sun, Zhenwei Wang, Chenjie Cao, Hengshuang Zhao, et~al.
\newblock {WorldCompass}: Reinforcement learning for long-horizon world models.
\newblock \emph{arXiv preprint arXiv:2602.09022}, 2026{\natexlab{c}}.

\bibitem[Wang et~al.(2026{\natexlab{d}})Wang, Zhang, Zhang, Zhang, Tan, Qin, and Li]{wang2026world}
Zijie Wang, Wei Zhang, Weiming Zhang, Fanqi Zhang, Xiao Tan, Yipeng Qin, and Guanbin Li.
\newblock World models as group actions.
\newblock \emph{arXiv preprint arXiv:2605.24578}, 2026{\natexlab{d}}.

\bibitem[Wu et~al.(2026)Wu, Pi, Zhang, Li, and Zou]{wu2026quantitative}
Jiaxin Wu, Yihao Pi, Yinling Zhang, Yuheng Li, and Xueyan Zou.
\newblock Quantitative video world model evaluation for geometric-consistency.
\newblock \emph{arXiv preprint arXiv:2605.15185}, 2026.

\bibitem[Xiang et~al.(2026)Xiang, Liu, Zhang, Yang, Fang, Wang, Wang, Zou, Su, and Zhu]{viewrope}
Chendong Xiang, Jiajun Liu, Jintao Zhang, Xiao Yang, Zhengwei Fang, Shizun Wang, Zijun Wang, Yingtian Zou, Hang Su, and Jun Zhu.
\newblock Geometry-aware rotary position embedding for consistent video world model.
\newblock \emph{arXiv preprint arXiv:2602.07854}, 2026.

\bibitem[Xiao et~al.(2025)Xiao, Lan, Zhou, Ouyang, Yang, Zeng, and Pan]{Worldmem}
Zeqi Xiao, Yushi Lan, Yifan Zhou, Wenqi Ouyang, Shuai Yang, Yanhong Zeng, and Xingang Pan.
\newblock {WorldMem}: Long-term consistent world simulation with memory.
\newblock \emph{arXiv preprint arXiv:2504.12369}, 2025.

\bibitem[Xu et~al.(2026{\natexlab{a}})Xu, Sui, Gao, Shi, Yang, Liu, Sun, Sun, Pan, Jiang, et~al.]{worldodysseybench}
Ting-Bing Xu, Jiacheng Sui, Zhe Gao, Kewei Shi, Wenjin Yang, Zhicheng Liu, Zhaoxu Sun, Mingchao Sun, Hongyu Pan, Fan Jiang, et~al.
\newblock {WorldRoamBench}: An open-world benchmark for long-horizon stability of interactive world models.
\newblock \emph{arXiv preprint arXiv:2606.31672}, 2026{\natexlab{a}}.

\bibitem[Xu et~al.(2026{\natexlab{b}})Xu, Lin, He, Feng, Mao, Yin, Ge, and Zhang]{worldmark}
Xiaojie Xu, Zhengyuan Lin, Kang He, Yukang Feng, Xiaofeng Mao, Yuanyang Yin, Yongtao Ge, and Kaipeng Zhang.
\newblock {WorldMark}: A unified benchmark suite for interactive video world models.
\newblock \emph{arXiv preprint arXiv:2604.21686}, 2026{\natexlab{b}}.

\bibitem[Yang et~al.(2023)Yang, Du, Ghasemipour, Tompson, Kaelbling, Schuurmans, and Abbeel]{yang2024unisim}
Sherry Yang, Yilun Du, Kamyar Ghasemipour, Jonathan Tompson, Leslie Kaelbling, Dale Schuurmans, and Pieter Abbeel.
\newblock Learning interactive real-world simulators.
\newblock \emph{arXiv preprint arXiv:2310.06114}, 2023.

\bibitem[Yang et~al.(2026)Yang, Shang, Wang, Zhou, Tan, Zeng, Li, Li, Liu, Yang, et~al.]{worldexam}
Yuxue Yang, Shuyao Shang, Jiahe Wang, Zitong Zhou, Liang Tan, Junhan Zeng, Ruizhi Li, Junyan Li, Yu Liu, Xiao Yang, et~al.
\newblock {WorldExam}: Benchmarking world models from apparent appearance to inherent reactivity.
\newblock \emph{arXiv preprint arXiv:2608.02603}, 2026.

\bibitem[Ye et~al.(2026)Ye, Lu, Jiang, Gu, Zhao, Liang, Pan, Zhang, Wu, and Wang]{mind}
Yixuan Ye, Xuanyu Lu, Yuxin Jiang, Yuchao Gu, Rui Zhao, Qiwei Liang, Jiachun Pan, Fengda Zhang, Weijia Wu, and Alex~Jinpeng Wang.
\newblock {MIND}: Benchmarking memory consistency and action control in world models.
\newblock \emph{arXiv preprint arXiv:2602.08025}, 2026.

\bibitem[Yin et~al.(2026)Yin, Wang, Zhan, Li, Zhang, and Zhao]{evoke}
Yuanyang Yin, Gongxuan Wang, Yifan Zhan, Chuanhao Li, Kaipeng Zhang, and Feng Zhao.
\newblock {Alaya-EVOKE}: From linear-scaling supervision to endless world.
\newblock \emph{arXiv preprint arXiv:2608.13546}, 2026.

\bibitem[Ying et~al.(2026)Ying, Hu, Ren, Li, Chen, Wang, Cao, Cai, and Ding]{wbench}
Kaining Ying, Hengrui Hu, Siyu Ren, Jiamu Li, Fengjiao Chen, Ziwen Wang, Xuezhi Cao, Xunliang Cai, and Henghui Ding.
\newblock {WBench}: A comprehensive multi-turn benchmark for interactive video world model evaluation.
\newblock \emph{arXiv preprint arXiv:2605.25874}, 2026.

\bibitem[Yu et~al.(2026)Yu, Liu, Hu, Yuan, Feng, Zhou, Huang, Wang, Zhao, Luo, et~al.]{worldcrafter}
Wangbo Yu, Kunhao Liu, Wenbo Hu, Shenghai Yuan, Chaoran Feng, Haiyang Zhou, Yukun Huang, Yiran Wang, Wang Zhao, Yingmin Luo, et~al.
\newblock {WorldCrafter}: Consistent video world model with implicit {3D}-aware memory.
\newblock \emph{arXiv preprint arXiv:2609.24984}, 2026.

\bibitem[Zhang et~al.(2026)Zhang, Zhang, Huang, Tang, Wang, Dai, Chen, Li, Li, Chen, et~al.]{mbench}
Shengjun Zhang, Zhang Zhang, Simin Huang, Zhenyu Tang, Hanyang Wang, Chensheng Dai, Min Chen, Yifan Li, Yuxin Li, Yingjie Chen, et~al.
\newblock {MBench}: A comprehensive benchmark on memory capability for video world models.
\newblock \emph{arXiv preprint arXiv:2606.00793}, 2026.

\bibitem[Zhang et~al.(2025)Zhang, Peng, Wang, Wang, Zhu, Kang, Jiang, Gao, Li, Liu, et~al.]{matrix}
Yifan Zhang, Chunli Peng, Boyang Wang, Puyi Wang, Qingcheng Zhu, Fei Kang, Biao Jiang, Zedong Gao, Eric Li, Yang Liu, et~al.
\newblock {Matrix-Game}: Interactive world foundation model.
\newblock \emph{arXiv preprint arXiv:2506.18701}, 2025.

\bibitem[Zheng et~al.(2024)Zheng, Li, Jiang, Lu, Wu, and Li]{cami2v}
Guangcong Zheng, Teng Li, Rui Jiang, Yehao Lu, Tao Wu, and Xi Li.
\newblock {CamI2V}: Camera-controlled image-to-video diffusion model.
\newblock \emph{arXiv preprint arXiv:2410.15957}, 2024.

\bibitem[Zhu et~al.(2026)Zhu, Liu, Zhao, Ye, Chen, Yu, He, Han, and Xie]{sana}
Haoyi Zhu, Haozhe Liu, Yuyang Zhao, Tian Ye, Junsong Chen, Jincheng Yu, Tong He, Song Han, and Enze Xie.
\newblock {SANA-WM}: Efficient minute-scale world modeling with hybrid linear diffusion transformer.
\newblock \emph{arXiv preprint arXiv:2605.15178}, 2026.

\end{thebibliography}
}

\ifwithsupp
\clearpage
\onecolumn
\begin{center}
{\Large\textbf{\papertitle}}\\[0.5em]
{\large Supplementary Material}
\end{center}
\vspace{1em}
\appendix
\counterwithin{figure}{section}
\counterwithin{table}{section}
\numberwithin{equation}{section}
\providecolor{alertred}{HTML}{D64541}
\makeatletter
\providecolor{cwblue}{HTML}{1F5FA8}\providecolor{cwink}{HTML}{1D2430}\providecolor{cwcard}{HTML}{F3F5F8}
\providecolor{cwmuted}{HTML}{6B7380}
\def\cvprsection{\@startsection{section}{1}{\z@}{-4pt}{10pt}{\Large\bfseries\color{cwink}}}
\def\cvprsect#1{\cvprsection{#1}}
\def\cvprsubsection{\@startsection{subsection}{2}{\z@}{-10pt plus -2pt minus -2pt}{5pt}{\large\bfseries\color{cwink}}}
\def\cvprsubsect#1{\cvprsubsection{#1}}
\def\cw@fmt@section{\fcolorbox{cwblue}{cwblue}{\color{white}\,\thesection\,}\enskip}
\def\cw@fmt@subsection{{\color{cwblue}\thesubsection}\enskip}
\def\@seccntformat#1{\@ifundefined{cw@fmt@#1}{\csname the#1\endcsname\quad}{\csname cw@fmt@#1\endcsname}}
\let\cw@section\section
\def\section{\clearpage\cw@section}
\let\cw@paragraph\paragraph
\renewcommand\paragraph[1]{\cw@paragraph{\color{cwink}#1}}
\makeatother

\noindent{\Large\bfseries\color{cwink}Contents}\par\smallskip
\noindent This supplement follows the structure of the main paper; each appendix lists the section it supports.\par\medskip
\begingroup\footnotesize\setlength{\tabcolsep}{5pt}\renewcommand{\arraystretch}{1.25}
\noindent\begin{tabularx}{\linewidth}{@{}c>{\raggedright\arraybackslash}Xcr@{}}
\toprule
 & Appendix & Supports & Page \\
\midrule
\textcolor{cwblue}{\textbf{\ref{app:A}}} & \textbf{Benchmark details} \newline {\color{cwmuted}start views, route pairs, adapters, command magnitude, Genie 3, GameCraft, pitch} & Sec.~\ref{sec:benchmark} & \pageref{app:A} \\
\textcolor{cwblue}{\textbf{\ref{app:C}}} & \textbf{Pose recovery and geometric scoring} \newline {\color{cwmuted}anchors, measurable field of view, Rigidity offset, Scene stability} & Sec.~\ref{sec:method} & \pageref{app:C} \\
\textcolor{cwblue}{\textbf{\ref{app:D}}} & \textbf{Strides, Pace, and Composition scoring} \newline {\color{cwmuted}stride calibration, Response, Pace, Composition} & Sec.~\ref{sec:method} & \pageref{app:D} \\
\textcolor{cwblue}{\textbf{\ref{app:res_f1}}} & \textbf{Further results on Rigidity} \newline {\color{cwmuted}matched image speed, alternative statistics} & Sec.~\ref{sec:res_f1} & \pageref{app:res_f1} \\
\textcolor{cwblue}{\textbf{\ref{app:res_f2}}} & \textbf{Further results on Pace} \newline {\color{cwmuted}per-step pace, equal-weight strides, response profiles} & Sec.~\ref{sec:res_f2} & \pageref{app:res_f2} \\
\textcolor{cwblue}{\textbf{\ref{app:res_f3}}} & \textbf{Further results on Composition} \newline {\color{cwmuted}step-by-step routes, regeneration error, repeated generations, scale fitting, existing metrics} & Sec.~\ref{sec:res_f3} & \pageref{app:res_f3} \\
\textcolor{cwblue}{\textbf{\ref{app:E}}} & \textbf{Validation on known motion and rigid renderings} \newline {\color{cwmuted}pose accuracy, degradations, rigid floor, rigid counterparts, controlled changes, VGGT} & Sec.~\ref{sec:validation} & \pageref{app:E} \\
\textcolor{cwblue}{\textbf{\ref{app:H}}} & \textbf{Sensitivity and robustness} \newline {\color{cwmuted}anchor splits, evaluation parameters, opposite actions, stride references, intervals, step duration, matched motion, stride variation} & Sec.~\ref{sec:validation} & \pageref{app:H} \\
\textcolor{cwblue}{\textbf{\ref{app:human}}} & \textbf{Human study} \newline {\color{cwmuted}ten blinded raters, control and rigidity} & Sec.~\ref{sec:validation} & \pageref{app:human} \\
\textcolor{cwblue}{\textbf{\ref{app:I}}} & \textbf{Fine-tuning details} \newline {\color{cwmuted}training routes, checkpoints, command-scaled and perturbed-path controls} & Sec.~\ref{sec:sft} & \pageref{app:I} \\
\textcolor{cwblue}{\textbf{\ref{app:repro}}} & \textbf{Reproducibility and supplementary video} \newline {\color{cwmuted}released artifacts, video contents} & -- & \pageref{app:repro} \\
\textcolor{cwblue}{\textbf{\ref{app:related}}} & \textbf{Evidence for Tab.~\ref{tab:related}} \newline {\color{cwmuted}per-benchmark evidence for the comparison} & Sec.~\ref{sec:related} & \pageref{app:related} \\
\textcolor{cwblue}{\textbf{\ref{app:limitations}}} & \textbf{Limitations} \newline {\color{cwmuted}measurement range, scene and motion scope, fine-tuning} & Sec.~\ref{sec:conclusion} & \pageref{app:limitations} \\
\bottomrule
\end{tabularx}
\endgroup
\bigskip

\makeatletter\let\cwclearsection\section\let\section\cw@section\makeatother
\FloatBarrier
\section{Benchmark Details}\label{app:A}
\subsection{Start views}\label{app:A1}
Camera coordinates follow the right-handed OpenCV convention with $x$ pointing right, $y$ down, and $z$ forward. Ground-truth depth is stored as camera-axis depth $z$, obtained for UE5 views by scaling the EXR depth channel with the conversion factor recorded for each view and for ETH3D views by aligning the laser-scan depth to the same convention, and 3D anchors are placed only at pixels with valid depth. All control metrics divide displacements by the model's own stride or by the median depth of the start view, so the internal length unit cancels out. When an adapter crops or resizes the start view with a 2D affine pixel transform $\mathbf A$, the intrinsics become $\mathbf K'=\mathbf A\mathbf K$ and the anchor pixels are transformed in the same way. As shown in Fig.~\ref{fig:A.cases}, the rendered views have dense depth and the real views have sparse laser-scan depth.

\begin{figure}[!htbp]
\centering
\includegraphics[width=0.9\linewidth]{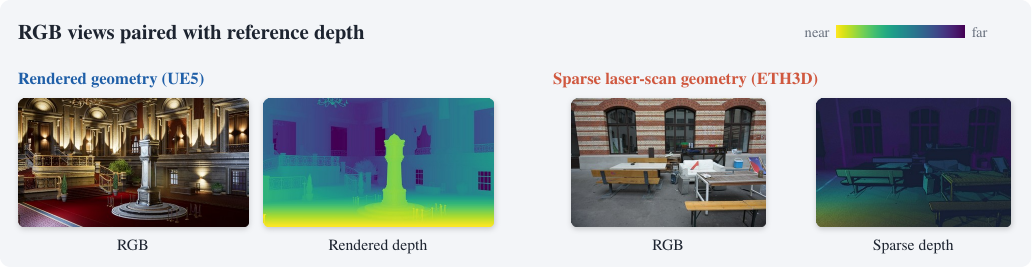}
\caption{\textbf{Start views and their ground-truth depth.} RGB start views paired with dense rendered depth (UE5) and sparse laser-scan depth (ETH3D). More start views are shown in Fig.~\ref{fig:bench}a.}
\label{fig:A.cases}
\end{figure}

\FloatBarrier
\subsection{Route construction and assignment}\label{app:B2}\label{app:A2}
As shown in Tab.~\ref{tab:1}, the four route families vary the translation axes, the lengths of action blocks, the turning directions, and the insertion positions. G1 tests back-and-forth cancellation (\emph{cancel}), and G2--G4 test reordering without a turn, after a turn, and across turns (\emph{swap}). The route library contains $96$ pairs per family, for $384$ pairs and $768$ distinct action sequences in total. Each of the $30$ scenes is assigned one pair from each family, so all start views and models in a scene execute the same eight routes and the benchmark contains $120$ distinct route pairs. G2 routes have eight steps, G3 routes have eleven, and G1 and G4 routes have fourteen. Every step is one of the six basic actions, the two routes of a pair have the same number of steps, and no action sequence is repeated.

\begin{table}[!htbp]
\centering
\caption{\textbf{Schematic construction rules for the four route-pair families.} Each symbol denotes one step, and boldface marks the steps that are replaced or reordered. Actual benchmark routes contain $8$--$14$ steps.}
\label{tab:1}
\begingroup
\footnotesize
\setlength{\tabcolsep}{4pt}
\renewcommand{\arraystretch}{1.08}
\begin{tabularx}{\linewidth}{
    @{}l
    >{\raggedright\arraybackslash}X
    >{\raggedright\arraybackslash}X@{}
}
\toprule
Route family & Route 1 & Route 2 \\
\midrule
G1: cancel
& W $\to$ \textbf{A $\to$ D} $\to$ W
& W $\to$ \textbf{W $\to$ S} $\to$ W \\

G2: swap
& \textbf{W $\to$ W $\to$ D}
& \textbf{D $\to$ W $\to$ W} \\

G3: swap after a turn
& YR $\to$ \textbf{W $\to$ D}
& YR $\to$ \textbf{D $\to$ W} \\

G4: swap across turns
& YR $\to$ \textbf{W $\to$ D} $\to$ YL $\to$ \textbf{S $\to$ D}
& YR $\to$ \textbf{D $\to$ W} $\to$ YL $\to$ \textbf{D $\to$ S} \\
\bottomrule
\end{tabularx}
\endgroup
\end{table}

\FloatBarrier
\subsection{Control settings}\label{app:B}\label{app:B1}\label{app:B3}
As shown in Tab.~\ref{tab:B.1}, each model keeps its native control interface and output format, and its single-action and route videos use the same adapter settings. Discrete-key models receive their actions in the encoding of their released inference code, and camera-pose models receive per-step translations and yaw turns in their native input units. Matrix-Game 3.5, EVOKE, WorldCrafter, and SolarWM run their released distilled samplers with $3$, $3$, $6$, and $4$ denoising steps, and SolarWM uses its public stage-2 EMA checkpoint. HY-World 1.5 and HY-WorldPlay come from the same model family~\citep{worldplay}, the first as a step-distilled checkpoint conditioned on discrete control tokens and the second as an RL-trained autoregressive checkpoint conditioned on continuous pose tokens.

\begin{table}[!htbp]
\centering
\caption{\textbf{Model control interfaces and output configurations.} $^\ast$Uses the release default command per step (for Matrix-Game 2.0 and 3.0, default keyboard translation with yaw sensitivity scaled to stay within the first-frame field of view over $16$ steps). For the remaining models, per-step magnitudes are set as described in App.~\ref{app:magsens}.}
\label{tab:B.1}
\begingroup\footnotesize\setlength{\tabcolsep}{3pt}\renewcommand{\arraystretch}{1.08}
\begin{tabularx}{\linewidth}{@{}p{0.235\linewidth}ccl>{\raggedright\arraybackslash}X@{}}
\toprule
Model & \shortstack{Frames/\\unit} & FPS & Resolution & Control and preprocessing\\
\midrule
Matrix-Game 3.0$^\ast$ & 4 & 17 & 1280 $\times$ 704 & Keyboard 1; mouse 0.04; center crop \\
Matrix-Game 2.0$^\ast$ & 4 & 16 & 640 $\times$ 352 & Keyboard 1; mouse 0.035; center crop \\
SANA-WM$^\ast$ & 4 & 16 & 1280 $\times$ 704 & Native action sequence; center crop \\
DreamX-World$^\ast$ & 4 & 16 & 1280 $\times$ 704 & Action tokens; translation 0.075 and yaw $1.5^\circ$ per frame interval; stretch resize \\
HY-World 1.5$^\ast$ & 4 & 24 & 832 $\times$ 480 & Native w/s/a/d/left/right control tokens; translation 0.08 and yaw $3^\circ$ per action unit; native crop \\
HY-World 2.0 & 2 & 16 & 832 $\times$ 480 & Camera pose; translation 0.025 and yaw $3^\circ$ per action unit; stretch resize \\
Lyra 2.0 & 10 & 16 & 832 $\times$ 480 & Camera pose; translation 0.10 and yaw $2^\circ$ per action unit; stretch resize \\
SolarWM & 8 & 24 & 1344 $\times$ 768 & Camera pose; translation 0.2 and yaw $2.5^\circ$ per action unit; stretch resize \\
WorldCrafter & 16 & 16 & 640 $\times$ 384 & Camera pose; translation $16/33$ and yaw $4^\circ$ per action unit; stretch resize \\
LingBot-World v2 & 4 & 16 & 832 $\times$ 464 / 768 $\times$ 512 & Camera pose; translation 0.025 and yaw $0.75^\circ$ per frame interval; aspect-preserving resize \\
HY-WorldPlay$^\ast$ & 4 & 24 & 832 $\times$ 480 & Pose tokens; translation 0.08 and yaw $3^\circ$ per action unit (one official tick); native crop \\
EVOKE$^\ast$ & 8 & 24 & 640 $\times$ 384 & Camera pose; translation $1/6$ and yaw $5^\circ$ per action unit; stretch resize \\
Matrix-Game 3.5 & 8 & 16 & 1280 $\times$ 704 & Camera pose; translation 0.1 and yaw $0.6^\circ$ per action unit; center crop \\
AlayaWorld & 8 & 24 & 960 $\times$ 544 & Camera pose; forward/backward 1.27, lateral 0.0846, yaw $2^\circ$ per action unit; center crop \\
H3-World & 17 & 24 & 832 $\times$ 480 & Text: fixed camera-motion prompts; model-specific crop \\
\bottomrule
\end{tabularx}
\endgroup
\end{table}

\FloatBarrier
\subsection{Command magnitude}\label{app:magsens}
Discrete-key models and the camera-pose models EVOKE and HY-WorldPlay use their release default per-step command, whereas the remaining camera-pose interfaces accept arbitrary step magnitudes, as listed in Tab.~\ref{tab:B.1}. Matrix-Game 2.0 and 3.0 keep their default keyboard translation with the mouse yaw sensitivity lowered from $0.1$ to $0.035$ and $0.04$, so that $16$ held turn steps rotate the camera by roughly $17^\circ$--$18^\circ$ instead of about $54^\circ$, which is close to the roughly $60^\circ$ beyond which the camera is rarely recovered from the start-view anchors (App.~\ref{app:envelope}). SolarWM runs at half its default translation and a third of its default yaw, as $16$ steps of its default turn would rotate the camera by $120^\circ$. For AlayaWorld, the default per-step translation moves the camera by about the tracking noise, so we scale up its translation within the official interface. At the default command, its translation stride stays near the noise floor, its $16$-step pace still accelerates, and its \mHold\ changes by only $2.0$ points.

We also scale the per-step translation and yaw of EVOKE, WorldCrafter, and Lyra 2.0 by $0.5\times$, $1\times$, and $2\times$ on $30$ start views, the first view of each scene with seed $0$, across all six actions and eight routes ($420$ videos per setting). The composition findings hold at every magnitude, as EVOKE brings at most $3.3\%$ of its routes within one stride, WorldCrafter misses the endpoints of its completed routes by $2.1$ to $3.7$ strides, and Lyra 2.0 brings $61.7\%$ to $85.0\%$ of its routes within one stride. For EVOKE, these runs scale a smaller base command ($0.03125$ units and $0.5^\circ$ per action unit, about $1/5$ of the release default translation and $1/10$ of the default yaw in Tab.~\ref{tab:main}). Only the held pace of EVOKE varies with the command magnitude, as its median pace ratio grows from $0.51$, $0.64$, and $0.81$ at $0.5\times$, $1\times$, and $2\times$ of this command to $1.11$ at the release default on the same views.

\FloatBarrier
\subsection{Directional tests for Genie 3 and GameCraft}\label{app:G}\label{app:G1}
\paragraph{Genie 3.}
We collect $672$ videos from the closed-source Genie~3~\citep{genie3} through its web interface, comprising $288$ single-action and $384$ route videos from $48$ start views in ten UE5 scenes. Because the web stream does not provide frame-accurate input timestamps, step boundaries and strides cannot be recovered, so we test the direction of the net camera motion between the first and the last frame on videos whose endpoint pose reaches an inlier ratio and coverage of at least $0.3$. All $115$ translation videos with a recoverable endpoint move in the commanded direction, and the cosine between their net horizontal displacement and the commanded axis has a median between $98.9\%$ and $100.0\%$ across the four translations. Sustained turns mostly leave the field of view of the start view, so only $2$ of $48$ YL and $3$ of $48$ YR videos keep a recoverable endpoint.

\paragraph{GameCraft.}
We test Hunyuan-GameCraft~\citep{gamecraft} on $180$ videos in five UE5 scenes, $90$ single-action videos of two native control intervals and $90$ route videos, of which $172$ have valid poses at all step boundaries. It moves in the commanded direction in $100.0\%$ of forward and $93.3\%$ of backward videos, but shows a strong directional bias for lateral translations ($26.7\%$ for A and $58.3\%$ for D) and turns ($20.0\%$ for YL and $80.0\%$ for YR), as both turn commands mostly rotate the camera in the same direction.

\FloatBarrier
\subsection{Pitch control}\label{app:G2}
We test upward and downward pitch commands (PU and PD) for ten models on $19$ start views with three seeds, for $114$ videos of $16$ steps per model. Because sustained pitch in $16{:}9$ videos quickly moves the surfaces of the start view out of frame, we evaluate all models over the first eight steps. Direction is the share of videos whose recovered pitch $\vartheta_8$ at step $8$ has the commanded sign, and the extra rotation $C_{\mathrm{rot}}(8)=100\sqrt{\psi_8^2+\varphi_8^2}/|\vartheta_8|$ measures the yaw $\psi_8$ and roll $\varphi_8$ that accompany the pitch. As shown in Tab.~\ref{tab:G.3}, eight of the ten models pitch in the commanded direction in $99.1\%$ to $100.0\%$ of videos with less than $7\%$ extra rotation, whereas AlayaWorld and the text-driven H3-World control vertical camera motion poorly.

\begin{table}[!htbp]
\centering
\caption{\textbf{Pitch control over the first eight steps} (standard pose-quality threshold $0.1$). All $114$ videos of every model are measurable except H3-World ($87/114$). Under a relaxed pose-coverage threshold ($0.0$) only H3-World changes ($90/114$ measurable, direction $60.5\%$).}
\label{tab:G.3}\label{tab:G.4}
\begingroup\footnotesize\setlength{\tabcolsep}{3pt}\renewcommand{\arraystretch}{1.08}
\begin{tabular*}{0.58\linewidth}{@{\extracolsep{\fill}}lrr@{}}
\toprule
Model & \shortstack{Direction\\(\%) $\uparrow$} & \shortstack{Extra rotation\\median (\%) $\downarrow$} \\
\midrule
Matrix-Game 3.0 & \textbf{100.0} & \underline{2.9} \\
Matrix-Game 2.0 & \textbf{100.0} & 3.9 \\
SANA-WM & \textbf{100.0} & 3.7 \\
DreamX-World & \textbf{100.0} & 6.9 \\
HY-World 1.5 & \textbf{100.0} & 5.1 \\
HY-World 2.0 & \underline{99.1} & \textbf{2.4} \\
Lyra 2.0 & \textbf{100.0} & 3.9 \\
LingBot-World v2 & \textbf{100.0} & 6.5 \\
AlayaWorld & 78.1 & 42.6 \\
H3-World & 57.9 & 7.2 \\
\bottomrule
\end{tabular*}
\endgroup
\end{table}

\let\section\cwclearsection
\FloatBarrier
\section{Pose Recovery and Geometric Scoring Details}\label{app:C}
\subsection{Coordinates, anchors, and solver}\label{app:C1}\label{app:C2}
An anchor pixel $\mathbf p=[u,v,1]^\top$ of the start view with valid camera-axis depth $z(\mathbf p)$ gives the 3D point $\mathbf P_{\mathrm{cam}}=z(\mathbf p)\mathbf K^{-1}\mathbf p$ in the initial camera frame. With $\mathbf T^{\mathrm{c2w}}_0$ the initial camera-to-world transform and $\hat{\mathbf T}^{\mathrm{w2c}}_t$ the recovered world-to-camera transform at frame $t$, the camera pose of frame $t$ relative to the initial camera is
\begin{equation}
\mathbf G_t=
\left(\mathbf T^{\mathrm{c2w}}_0\right)^{-1}
\left(\hat{\mathbf T}^{\mathrm{w2c}}_t\right)^{-1}.
\label{eq:C.1}
\end{equation}
Rigidity uses the full $SE(3)$ pose, and Pace and Composition use its planar part (App.~\ref{app:D1}).

Anchors are selected once per start view as SIFT keypoints whose neighbouring depths lie within $5\%$ of their own, or as Shi--Tomasi corners on ETH3D views whose laser depth is sparse. The image is divided into an $8\times6$ grid whose cells keep at most $12$ anchors each, and the anchors, sorted by a hash of their identifiers, are assigned alternately to the pose anchors $\mathcal P$ and the held-out anchors $\mathcal Q$. A start view thus has a median of $446$ anchors ($33$ to $576$) spread over $42$ of the $48$ cells, with $223$ pose and $223$ held-out anchors, and every model is evaluated with the same anchors. CoTracker3~\citep{cotracker3} tracks both sets through the generated video, and a track at frame $t$ is kept if its coordinates are finite and in frame, its visibility is at least $0.9$, and its confidence is at least $0.5$. The camera of each frame is initialized from $\mathcal P$ by EPnP with RANSAC~\citep{pnp,fischler1981random} at an angular threshold of $1^\circ$ ($\sqrt{f_xf_y}\tan(1^\circ)$ in pixels) and refined by Levenberg--Marquardt and nonlinear least squares on the viewing rays. Initialization requires at least $12$ valid pose anchors, and the refined solution must keep at least six inliers within $1^\circ$. With $v_{it}\in\{0,1\}$ indicating a valid track of pose anchor $i$ at frame $t$ and $\phi_{it}^{\mathrm{Pose}}$ its angular residual, the tracking coverage $c_t$ and inlier ratio $\rho_t$ are
\begin{equation}
c_t=\frac{\sum_{i\in\mathcal P}v_{it}}{|\mathcal P|},
\qquad
\rho_t=
\frac{\#\{i\in\mathcal P:v_{it}=1,\ \phi_{it}^{\mathrm{Pose}}\le1^\circ\}}
{|\mathcal P|}.
\label{eq:C.2}
\end{equation}
A frame has a valid pose if the solver returns a finite rigid transform with $c_t\ge0.1$ and $\rho_t\ge0.1$.

\FloatBarrier
\subsection{Rigidity offset and Scene stability}\label{app:C3}
For a held-out anchor $\mathbf X_i\in\mathcal Q$ with tracked position $\widetilde{\mathbf x}_{i,t}$ and recovered camera-to-world transform $\widehat{\mathbf T}_t=(\hat{\mathbf T}^{\mathrm{w2c}}_t)^{-1}$, its rigid projection, angular ray residual, and the mean offset of video $v$ are
\begin{equation}
\begin{aligned}
\widehat{\mathbf x}_{i,t}
&=\pi_{\mathbf K}\!\left(\widehat{\mathbf T}_t^{-1}\mathbf X_i^h\right),\\
e_{i,t}^{\mathrm{ray}}
&=\angle\!\left(\mathbf K^{-1}\widetilde{\mathbf x}_{i,t}^{h},
\mathbf K^{-1}\widehat{\mathbf x}_{i,t}^{h}\right),\\
\bar e_{\mathrm{ray}}(v)
&=\frac{1}{|\mathcal E_v|}\sum_{(i,t)\in\mathcal E_v}e_{i,t}^{\mathrm{ray}},
\end{aligned}
\label{eq:1}
\end{equation}
where $\pi_{\mathbf K}$ is perspective projection, superscript $h$ denotes homogeneous coordinates, and $\mathcal E_v$ collects the valid held-out tracks in frames $t>0$ of video $v$. A frame enters the Rigidity evaluation if it has a valid pose and at least $12$ valid held-out anchors in front of the camera. Its offset $e_t$ averages $e^{\mathrm{ray}}_{i,t}$ over these anchors, and its image motion is the median angle between the viewing ray of each held-out anchor in the start view and its rigidly projected ray in frame $t$. Tab.~\ref{tab:main} reports the median $e_t$ over route-video frames whose image motion lies between $2^\circ$ and $8^\circ$, and App.~\ref{app:ceiling} establishes its floor on rigid scenes.

\paragraph{Scene stability (\mScene).}
The supplementary video-level score \mScene\ maps $\bar e_{\mathrm{ray}}(v)$, in degrees, to $s(\bar e_{\mathrm{ray}}(v)/1^\circ)$ with $s$ of Eq.~\eqref{eq:6}, assigns zero to a video without valid observations, and averages over the eight route videos of each start view and then over the $140$ start views. Because it pools all frames regardless of how far the camera moves, it favors models that move the camera little (App.~\ref{app:uncertainty}).

\FloatBarrier
\subsection{Measurable range of the start-view anchors}\label{app:envelope}
The camera can be recovered from the start-view anchors as long as enough of the initial surfaces remain in view. As shown in Tab.~\ref{tab:envelope}, pose recovery succeeds on at least $99.6\%$ of steps while the cumulative translation stays within $25\%$ of the start view's median depth and the cumulative yaw within $30^\circ$. It remains at $93\%$--$96\%$ up to $150\%$ of the depth and $45^\circ$--$60^\circ$ of yaw, and drops to $76.9\%$ beyond $60^\circ$, where rotation moves most of the initial scene out of view. The per-step yaw commands rotate the camera by roughly $15^\circ$--$35^\circ$ over $16$ held steps, keeping $99\%$ of single-action turn videos within the high-recovery range through step $16$ (App.~\ref{app:magsens}).

\begin{table}[H]
\centering
\caption{\textbf{Measurable range of the start-view anchors.} Share of steps with successful pose recovery, given a valid pose at the preceding step, over steps 2--16 of all single-action videos of the fifteen models. Steps are grouped by their cumulative translation, relative to the median depth of the start view, and by their cumulative yaw.}
\label{tab:envelope}
\begingroup\footnotesize\setlength{\tabcolsep}{5pt}\renewcommand{\arraystretch}{1.08}
\begin{tabular*}{0.84\linewidth}{@{\extracolsep{\fill}}lrr@{\hskip 18pt}lrr@{}}
\toprule
Translation (\% of depth) & Recovered (\%) & Steps & Turn ($^\circ$) & Recovered (\%) & Steps \\
\midrule
0--10 & 99.9 & 153{,}432 & 0--5 & 100.0 & 48{,}155 \\
10--25 & 99.6 & 108{,}677 & 5--10 & 99.9 & 35{,}542 \\
25--50 & 98.4 & 68{,}710 & 10--20 & 99.7 & 55{,}468 \\
50--75 & 95.4 & 20{,}465 & 20--30 & 99.7 & 30{,}613 \\
75--100 & 94.5 & 6{,}662 & 30--45 & 95.8 & 12{,}291 \\
100--150 & 93.3 & 3{,}692 & 45--60 & 93.5 & 1{,}974 \\
$>150$ & 89.2 & 1{,}019 & $>60$ & 76.9 & 536 \\
\bottomrule
\end{tabular*}
\endgroup
\end{table}

\FloatBarrier
\section{Strides, Pace, and Composition Scoring Details}\label{app:D}
\subsection{Planar transforms and strides}\label{app:D1}
Let $c_0$ be the start frame of an action sequence and $c_k$ the frame at the end of step $k$. The relative pose $\mathbf G_{c_0}^{-1}\mathbf G_{c_k}$ has rotation $\mathbf R_{a,k}$ and translation $(t_{x,a,k},t_{y,a,k},t_{z,a,k})^\top$, and the projection $\Pi$ keeps its horizontal position and heading,
\begin{equation}
\begin{aligned}
\mathbf H_{a,k}
&=\Pi\!\left(\mathbf G_{c_0}^{-1}\mathbf G_{c_k}\right)
=\begin{bmatrix}
\cos\theta_{a,k}&\sin\theta_{a,k}&t_{x,a,k}\\
-\sin\theta_{a,k}&\cos\theta_{a,k}&t_{z,a,k}\\
0&0&1
\end{bmatrix},\\
\theta_{a,k}
&=\operatorname{atan2}\!\left((\mathbf R_{a,k})_{02},(\mathbf R_{a,k})_{00}\right),
\end{aligned}
\label{eq:D.1}
\end{equation}
where matrix indices are zero-based, transforms act on column vectors, $\mathbf H_{a,0}=\mathbf I$, and positive yaw rotates the camera toward $+x$. For $\mathbf H\in SE(2)$, $\mathbf t(\mathbf H)$ is its 2D translation and $\theta(\mathbf H)=\operatorname{atan2}(H_{01},H_{00})$ its yaw. For a twist $\boldsymbol\xi=(\rho_x,\rho_z,\theta)^\top\in\mathfrak{se}(2)$, the hat operator and the exponential and logarithmic maps of Eq.~\eqref{eq:2} are
\begin{equation}
\boldsymbol\xi^\wedge=
\begin{bmatrix}0&\theta&\rho_x\\-\theta&0&\rho_z\\0&0&0\end{bmatrix},
\qquad
\operatorname{Exp}(\boldsymbol\xi)=\exp(\boldsymbol\xi^\wedge),
\qquad
\bigl(\operatorname{Log}(\mathbf H)\bigr)^\wedge=\log\mathbf H.
\label{eq:D.2}
\end{equation}
Calibration requires valid poses at the start and at each of the first four step boundaries, with cumulative yaw below $170^\circ$. With $\bar{\boldsymbol\xi}_a=(\rho_{x,a},\rho_{z,a},\omega_a)^\top$ the mean twist of Eq.~\eqref{eq:2}, the signed axial stride component $u_a$ is positive in the commanded direction,
\begin{equation}
u_a=
\begin{cases}
\rho_{z,a},&a=\mathrm W,\\
-\rho_{z,a},&a=\mathrm S,\\
-\rho_{x,a},&a=\mathrm A,\\
\rho_{x,a},&a=\mathrm D,\\
-\omega_a,&a=\mathrm{YL},\\
\omega_a,&a=\mathrm{YR}.
\end{cases}
\label{eq:D.3}
\end{equation}
It enters the stall check of the composition score and the stride variation statistics (App.~\ref{app:F}).

\paragraph{Pace score (\mHold).}
Let $\mathcal A_t=\{\mathrm{W,S,A,D}\}$ denote the translations and $\mathcal A_r=\{\mathrm{YL,YR}\}$ the turns. At the checkpoints $k\in\{12,16\}$, the relative error of the predicted pose $\widehat{\mathbf H}_{a,k}=\mathbf U_a^k$ is
\begin{equation}
r_{a,k}=
\begin{cases}
\displaystyle
\frac{
\|\mathbf t(\mathbf H_{a,k})
-\mathbf t(\widehat{\mathbf H}_{a,k})\|_2
}{
k\|\mathbf t(\mathbf U_a)\|_2
},
&a\in\mathcal A_t,\\[10pt]
\displaystyle
\frac{
\left|\operatorname{wrap}\!\left(
\theta(\mathbf H_{a,k})
-\theta(\widehat{\mathbf H}_{a,k})
\right)\right|
}{
k|\theta(\mathbf U_a)|
},
&a\in\mathcal A_r,
\end{cases}
\qquad k\in\{12,16\},
\label{eq:3}
\end{equation}
where $\operatorname{wrap}(\cdot)$ returns the shortest signed angular difference. Each relative error is mapped to $[0,100]$ by
\begin{equation}
s(r)=\frac{100}{1+r},
\label{eq:6}
\end{equation}
and the \mHold\ of a video averages the scores at steps $12$ and $16$. A checkpoint without a recovered pose and every checkpoint of a video that fails the response check below score zero.

\paragraph{Response (\mResp).}\label{app:resp}
For a single-action video, let $u_k$ be the signed cumulative displacement along the commanded axis after $k$ steps and $k^\ast\le 16$ the last step with a recovered pose. The video responds if $u_{k^\ast}\ge 0.02\,d_v$ for a translation, with $d_v$ the median valid depth of the start view, or $u_{k^\ast}\ge 2^\circ$ for a turn, and a video without any recovered pose does not respond. As shown in Tab.~\ref{tab:resp}, halving or doubling both thresholds changes \mResp\ and \mHold\ by at most $1.5$ points for thirteen models and by up to $12.2$ points for AlayaWorld and Matrix-Game 3.5, the two models with the smallest single-step motion.

\begin{table}[!ht]
\centering
\caption{\textbf{\mResp\ by action and its sensitivity to the response thresholds.} Left: \mResp\ (\%) of each action. Right: overall \mResp\ and \mHold\ when both thresholds, $2\%$ of the start-view median depth and $2^\circ$ of yaw, are scaled by $0.5\times$ and $2\times$.}
\label{tab:resp}
\begingroup\footnotesize\setlength{\tabcolsep}{2.2pt}\renewcommand{\arraystretch}{1.08}
\begin{tabular*}{\linewidth}{@{\extracolsep{\fill}}lrrrrrrrrrr@{}}
\toprule
& \multicolumn{6}{c}{\mResp\ per action} & \multicolumn{2}{c}{\mResp} & \multicolumn{2}{c}{\mHold} \\
\cmidrule(lr){2-7}\cmidrule(lr){8-9}\cmidrule(lr){10-11}
Model & W & S & A & D & YL & YR & $\times0.5$ & $\times2$ & $\times0.5$ & $\times2$ \\
\midrule
Matrix-Game 3.0 & 100.0 & 100.0 & 98.7 & 98.9 & 100.0 & 100.0 & 99.8 & 99.5 & 77.1 & 77.0 \\
Matrix-Game 2.0 & 100.0 & 100.0 & 8.0 & 98.0 & 100.0 & 100.0 & 84.6 & 84.2 & 64.9 & 64.7 \\
SANA-WM & 100.0 & 100.0 & 97.8 & 98.7 & 100.0 & 100.0 & 99.5 & 98.9 & 76.3 & 76.0 \\
DreamX-World & 100.0 & 100.0 & 97.1 & 97.1 & 100.0 & 100.0 & 99.5 & 98.7 & 46.6 & 46.3 \\
HY-World 1.5 & 100.0 & 99.6 & 99.1 & 99.1 & 100.0 & 100.0 & 99.9 & 99.2 & 77.8 & 77.5 \\
HY-World 2.0 & 100.0 & 100.0 & 98.7 & 99.0 & 100.0 & 98.1 & 99.4 & 99.0 & 86.9 & 86.6 \\
Lyra 2.0 & 100.0 & 100.0 & 98.7 & 98.4 & 100.0 & 100.0 & 99.6 & 99.0 & 88.2 & 87.7 \\
SolarWM & 98.0 & 99.1 & 98.5 & 97.1 & 100.0 & 100.0 & 98.8 & 98.4 & 85.4 & 85.3 \\
WorldCrafter & 99.4 & 100.0 & 99.1 & 98.1 & 100.0 & 100.0 & 99.7 & 99.0 & 80.3 & 79.9 \\
LingBot-World v2 & 99.7 & 100.0 & 97.3 & 98.4 & 100.0 & 100.0 & 99.4 & 98.6 & 82.7 & 82.2 \\
HY-WorldPlay & 100.0 & 100.0 & 99.6 & 99.1 & 100.0 & 100.0 & 99.9 & 99.6 & 74.9 & 74.8 \\
EVOKE & 100.0 & 100.0 & 97.5 & 97.6 & 100.0 & 100.0 & 99.3 & 98.9 & 74.0 & 73.8 \\
Matrix-Game 3.5 & 78.7 & 60.9 & 92.2 & 84.8 & 100.0 & 97.0 & 87.3 & 75.7 & 66.7 & 57.5 \\
AlayaWorld & 78.9 & 88.4 & 81.0 & 75.8 & 87.5 & 41.7 & 81.1 & 63.3 & 20.8 & 18.4 \\
H3-World & 90.3 & 89.1 & 88.5 & 91.6 & 94.2 & 97.4 & 92.7 & 90.3 & 25.1 & 24.6 \\
\bottomrule
\end{tabular*}
\endgroup
\end{table}

\FloatBarrier
\subsection{Composition scoring}\label{app:D3}\label{app:D2}\label{app:scoring_cvpr}
The route prediction $\widehat{\mathbf H}_q=\mathbf U_{a_1}\cdots\mathbf U_{a_n}$ and the errors $\mathbf E_q=\widehat{\mathbf H}_q^{-1}\mathbf H_q$ and $\mathbf E_{qq'}$ (Eq.~\eqref{eq:4}) use the strides calibrated under the same model, start view, and seed. For either error $\mathbf E$, position and heading are measured in units of the model's own translation stride $b_t$ and turn stride $b_r$,
\begin{equation}
\begin{aligned}
b_t
&=\operatorname{median}_{a\in\mathcal A_t}
\|\mathbf t(\mathbf U_a)\|_2,
&
r_t&=\frac{\|\mathbf t(\mathbf E)\|_2}{b_t},\\
b_r
&=\operatorname{median}_{a\in\mathcal A_r}
|\theta(\mathbf U_a)|,
&
r_\psi&=\frac{|\operatorname{wrap}(\theta(\mathbf E))|}{b_r},
\end{aligned}
\label{eq:5}
\end{equation}
so that $s(r_t)=50$ when the position error equals one stride and $s(r_\psi)=50$ when the heading error equals one turn stride. The score of a route $q$ or a pair $p$ averages the two components, $s_q=\frac{1}{2}(s(r_t)+s(r_\psi))$ and $s_p=\frac{1}{2}(s(r_t)+s(r_\psi))$.

A route is \emph{completed} and enters the set $\mathcal V_q$ if (i)~all six strides of its start view and seed are calibrated with $b_t,b_r>0$, (ii)~its start frame and every step boundary have valid poses, (iii)~no step stalls, meaning that every step moves or turns the camera along the commanded axis by at least $0.1|u_a|$, and (iv)~the generated first frame matches the adapter-processed start view with a mean RGB $L_1$ difference of at most $32$ on the $0$--$255$ scale. A pair is completed and enters $\mathcal V_p$ when both of its routes are completed. Over the $N_q=1{,}120$ routes and $N_p=560$ pairs of a model, incomplete routes and pairs score zero,
\begin{equation}
\mathrm{\mRoute}=\frac{1}{N_q}\sum_{q\in\mathcal V_q}s_q,
\qquad
\mathrm{\mPair}=\frac{1}{N_p}\sum_{p\in\mathcal V_p}s_p,
\label{eq:7}
\end{equation}
\mComp\ is $\frac{1}{2}(\mathrm{\mRoute}+\mathrm{\mPair})$, and the completion rate is $100\,|\mathcal V_q|/N_q$. End err.\ in Tab.~\ref{tab:main} is the median of $r_t$ over $\mathcal V_q$, and Near is the share of all $N_q$ routes that are completed with both $r_t\le 1$ and $r_\psi\le 1$, so a completed route whose position lies within one stride of the prediction but whose heading deviates by more than one turn stride is not counted as Near. This heading condition matters mainly for Matrix-Game 3.0, whose Near would be $65.6\%$ instead of $42.5\%$ with the position condition alone, while it changes Near by at most $4.3$ points for every other model.

\paragraph{Aggregation.}\label{sec:aggregation}
Relative errors are mapped to scores before averaging. \mResp\ and \mHold\ average over the three seeds of a view, then over the views of a scene and across the $30$ scenes, weighting the six actions equally. \mRoute, \mPair, and \mComp\ average over all routes and pairs of a model (Eq.~\eqref{eq:7}). \mRep\ and \mView\ average relative standard deviations within each scene and then across scenes. Confidence intervals use $2{,}000$ bootstrap resamples of the $30$ scenes, and scores restricted to completed routes and split by domain are reported in App.~\ref{app:H2}.

\FloatBarrier
\section{Further Results on Rigidity}\label{app:res_f1}\label{app:cvpr}
\subsection{Rigidity offset at matched speed and under other statistics}\label{app:f1speed}
Tab.~\ref{tab:f1speed} splits the frames whose image motion is $2^\circ$--$8^\circ$ by the speed of that motion, the change of image motion from the previous frame. Below $0.4^\circ$ per frame, where $58\%$ to $100\%$ of each model's frames lie, the offsets of twelve models are $2.1$ to $6.7$ times Lyra 2.0's in every speed range, and rigid renderings give a floor of $0.11^\circ$ to $0.15^\circ$ in every range including the fastest. HY-World 2.0 and Lyra 2.0 share resolution, frame rate, and resize and differ only in frames per step, yet HY-World 2.0's offset is $3.2$ times Lyra's ($95\%$ interval $[3.0, 3.4]$ over videos). With equal weight on each $1^\circ$ bin of image motion or with per-frame medians, every model except SolarWM still stays at least $2.4$ times Lyra's offset, and SolarWM stays at $1.6$ to $1.7$ times it (App.~\ref{app:ceiling}). Across the fourteen ranked models, neither the offset nor its ratio to the rigid counterpart is related to the \mComp, with Spearman's $\rho$ of $-0.01$ and $-0.07$.

\begin{table*}[!htbp]
\centering
\caption{\textbf{Rigidity robustness.} Left: median held-out anchor offset (degrees) in frames whose image motion is $2^\circ$--$8^\circ$, by the speed of the image motion (degrees per frame; --: fewer than $300$ frames), and the share of frames above $0.4^\circ$ per frame. Right: offset relative to Lyra 2.0 for the main-text statistic (pooled), with equal weight on each $1^\circ$ bin of image motion (equal), and with per-frame medians (median). Rigid renders: the rigid floor of App.~\ref{app:ceiling}.}
\label{tab:f1speed}
\begingroup\footnotesize\setlength{\tabcolsep}{4pt}
\renewcommand{\arraystretch}{1.08}\begin{tabular*}{\textwidth}{@{\extracolsep{\fill}}lccccccccc@{}}
\toprule
 & \multicolumn{5}{c}{Offset by image speed ($^\circ$/frame)} & & \multicolumn{3}{c}{Ratio to Lyra 2.0} \\
\cmidrule(lr){2-6}\cmidrule(lr){8-10}
Model & $<0.05$ & $0.05$--$0.1$ & $0.1$--$0.2$ & $0.2$--$0.4$ & $\ge0.4$ & $\ge0.4$ (\%) & pooled & equal & median \\
\midrule
Matrix-Game 3.0 & 0.40 & 0.40 & 0.40 & 0.38 & 0.36 & 24 & 3.6 & 4.1 & 3.5 \\
Matrix-Game 2.0 & 0.40 & 0.41 & 0.41 & 0.39 & 0.41 & 13 & 3.7 & 4.2 & 3.7 \\
SANA-WM & 0.24 & 0.28 & 0.29 & 0.30 & 0.28 & 19 & 2.5 & 2.6 & 2.5 \\
DreamX-World & 0.37 & 0.37 & 0.38 & 0.40 & 0.58 & 5 & 3.6 & 4.4 & 3.8 \\
HY-World 1.5 & 0.35 & 0.38 & 0.33 & 0.34 & 0.22 & 10 & 3.1 & 3.2 & 3.6 \\
HY-World 2.0 & 0.37 & 0.37 & 0.38 & 0.38 & 0.31 & 42 & 3.2 & 3.6 & 3.3 \\
Lyra 2.0 & 0.10 & 0.13 & 0.12 & 0.09 & -- & 0 & 1.0 & 1.0 & 1.0 \\
SolarWM & 0.18 & 0.19 & 0.17 & 0.15 & 0.17 & 2 & 1.6 & 1.7 & 1.6 \\
WorldCrafter & 0.34 & 0.35 & 0.30 & 0.35 & 0.83 & 1 & 3.1 & 3.3 & 3.0 \\
LingBot-World v2 & 0.41 & 0.41 & 0.40 & 0.37 & 0.34 & 24 & 3.6 & 3.9 & 3.4 \\
HY-WorldPlay & 0.41 & 0.41 & 0.39 & 0.37 & 0.35 & 13 & 3.6 & 4.0 & 3.6 \\
EVOKE & 0.33 & 0.34 & 0.34 & 0.35 & 0.33 & 10 & 3.2 & 3.4 & 3.0 \\
Matrix-Game 3.5 & 0.25 & 0.34 & 0.42 & 0.59 & -- & 2 & 3.1 & 5.3 & 3.2 \\
AlayaWorld$^\dagger$ & 0.13 & 0.45 & 0.46 & 0.11 & 1.09 & 6 & 2.4 & 2.5 & 2.4 \\
H3-World & 0.35 & 0.33 & 0.28 & 0.27 & 0.44 & 5 & 3.0 & 3.1 & 2.5 \\
\midrule
Rigid renders & 0.12 & 0.15 & 0.13 & 0.12 & 0.11 & 17 & 1.2 & 1.2 & 1.1 \\
\bottomrule
\end{tabular*}
\endgroup
\end{table*}

\paragraph{Observation support of the video-level score.}\label{app:scene_support}
For all fifteen models, every route video keeps a nonempty set of held-out anchors with valid observations for \mScene, the video-level rigidity score of App.~\ref{app:C3}, and the observations lost over a video mostly come from initial anchors leaving the field of view as the camera moves.

\FloatBarrier
\section{Further Results on Pace}\label{app:res_f2}

\subsection{How held actions change pace}\label{app:pace}
Tab.~\ref{tab:pace} follows the motion of each held step along the commanded axis, $u_k-u_{k-1}$, relative to its mean over steps 1--4. The last column leaves out the first step, which alone lowers the estimate of Eq.~\eqref{eq:2} when it is slow. As shown in Tab.~\ref{tab:pace}, HY-World 2.0, Lyra 2.0, HY-World 1.5, Matrix-Game 2.0 and 3.0, and WorldCrafter keep their per-step motion within $5\%$ on both comparisons, and SolarWM, LingBot-World v2, and Matrix-Game 3.5 slow by about a tenth. DreamX-World and AlayaWorld accelerate, with steps 13--16 moving $1.96$ and $4.71$ times as far as steps 1--4. H3-World starts with a short first step ($0.59$) and then keeps a steady pace relative to steps 2--5 ($1.01$), so its lower \mHold\ comes from this slow start.

\begin{table}[!htbp]
\centering
\caption{\textbf{Pace of held actions.} Per-step motion along the commanded axis relative to its mean over steps 1--4 (median over the single-action videos measurable at all 16 steps), at step 1 and averaged over steps 13--16, and the motion of steps 13--16 relative to steps 2--5 ($1$ = steady pace).}
\label{tab:pace}
\begingroup\footnotesize\setlength{\tabcolsep}{3pt}\renewcommand{\arraystretch}{1.08}
\begin{tabular*}{0.7\linewidth}{@{\extracolsep{\fill}}lrrr@{}}
\toprule
Model & Step 1 & Steps 13--16 & 13--16 / 2--5\\
\midrule
Matrix-Game 3.0 & 1.02 & 0.98 & 0.99 \\
Matrix-Game 2.0 & 1.00 & 0.95 & 0.98 \\
SANA-WM & 0.84 & 1.00 & 0.93 \\
DreamX-World & 0.53 & 1.96 & 1.49 \\
HY-World 1.5 & 0.99 & 0.95 & 0.95 \\
HY-World 2.0 & 0.99 & 1.00 & 1.00 \\
Lyra 2.0 & 0.99 & 0.99 & 0.98 \\
SolarWM & 0.99 & 0.89 & 0.90 \\
WorldCrafter & 0.84 & 1.02 & 0.98 \\
LingBot-World v2 & 0.96 & 0.88 & 0.89 \\
HY-WorldPlay & 0.86 & 1.07 & 1.01 \\
EVOKE & 0.67 & 0.94 & 0.84 \\
Matrix-Game 3.5 & 1.02 & 0.87 & 0.90 \\
AlayaWorld & 0.35 & 4.71 & 3.05 \\
H3-World & 0.59 & 1.12 & 1.01 \\
\bottomrule
\end{tabular*}
\endgroup
\end{table}

\FloatBarrier
\subsection{Strides with equal weights}\label{app:stridealt}
The stride estimator of Eq.~\eqref{eq:2} averages the per-step twist estimates over the first one, two, three, and four steps, which gives steps 1--4 effective weights of $0.52$, $0.27$, $0.15$, and $0.06$. Tab.~\ref{tab:stridealt} recomputes the pace ratio after $16$ steps and the share of routes ending near the prediction with the unweighted four-step mean $\mathbf U^{(4)}_a=\operatorname{Exp}(\operatorname{Log}\mathbf H_{a,4}/4)$ (M4) and with the unweighted mean over steps 2--5, $\operatorname{Exp}(\operatorname{Log}(\mathbf H_{a,1}^{-1}\mathbf H_{a,5})/4)$ (M2--5). As shown in Tab.~\ref{tab:stridealt}, the same two models accelerate under equal step weights (DreamX-World at $1.62$ and AlayaWorld at $2.94$), while H3-World, whose first step is short, falls within the steady band ($1.11$). Calibrating on steps 2--5 lowers DreamX-World's ratio to $1.25$ because its acceleration continues through step 6, and AlayaWorld remains at $1.92$. With equal-weight strides, Near changes by at most $6.7$ percentage points, and HY-World 2.0 ($77.6\%$) and Lyra 2.0 ($76.2\%$) remain the only models that bring most routes within one stride.

\begin{table}[!htbp]
\centering
\caption{\textbf{Pace ratio and route outcomes under three stride estimators.} K4: the stride of Eq.~\eqref{eq:2}. M4: the unweighted mean of steps 1--4. M2--5: the unweighted mean of steps 2--5. The pace ratio is the median after $16$ steps, in red outside $0.8$--$1.25$. Near: routes ending within one stride (\%). $^\dagger$Stride not reliably measurable.}
\label{tab:stridealt}
\begingroup\footnotesize\setlength{\tabcolsep}{3.5pt}
\renewcommand{\arraystretch}{1.08}\begin{tabular*}{0.8\linewidth}{@{\extracolsep{\fill}}lccccc@{}}
\toprule
 & \multicolumn{3}{c}{Pace ratio} & \multicolumn{2}{c}{Near (\%)} \\
\cmidrule(lr){2-4}\cmidrule(lr){5-6}
Model & K4 & M4 & M2--5 & K4 & M4 \\
\midrule
Matrix-Game 3.0 & 0.96 & 0.98 & 0.99 & 42.5 & 42.0 \\
Matrix-Game 2.0 & 0.96 & 0.97 & 0.99 & 24.7 & 21.9 \\
SANA-WM & 1.09 & 1.04 & 0.97 & 7.2 & 8.7 \\
DreamX-World & \textcolor{alertred}{2.11} & \textcolor{alertred}{1.62} & \textcolor{alertred}{1.25} & 0.5 & 2.5 \\
HY-World 1.5 & 0.97 & 0.98 & 0.98 & 3.8 & 4.8 \\
HY-World 2.0 & 1.00 & 1.00 & 1.00 & 84.3 & 77.6 \\
Lyra 2.0 & 1.01 & 1.00 & 0.99 & 75.9 & 76.2 \\
SolarWM & 0.94 & 0.95 & 0.95 & 25.4 & 24.5 \\
WorldCrafter & 1.08 & 1.03 & 0.98 & 16.0 & 16.2 \\
LingBot-World v2 & 0.94 & 0.94 & 0.94 & 11.2 & 9.8 \\
HY-WorldPlay & 1.11 & 1.05 & 1.00 & 3.8 & 5.5 \\
EVOKE & 1.12 & 0.97 & 0.89 & 3.1 & 4.4 \\
Matrix-Game 3.5 & 0.95 & 0.94 & 0.97 & 8.0 & 11.4 \\
AlayaWorld$^\dagger$ & \textcolor{alertred}{4.17} & \textcolor{alertred}{2.94} & \textcolor{alertred}{1.92} & 0.4 & 0.7 \\
H3-World & \textcolor{alertred}{1.29} & 1.11 & 0.99 & 0.2 & 0.4 \\
\bottomrule
\end{tabular*}
\endgroup
\end{table}

\FloatBarrier
\subsection{Response profiles}\label{app:profiles}
As shown in Figs.~\ref{fig:D.response} and~\ref{fig:D.response_new}, the response to each of the six basic actions grows close to linearly over the execution steps for most models, with fitted exponents $\gamma$ within $0.85$--$1.22$, while DreamX-World ($1.30$--$1.52$), H3-World ($1.11$--$1.49$), and AlayaWorld ($1.54$--$2.20$) accelerate. Only two single actions depart from their model's range, A (left) of Matrix-Game 2.0, which decays ($\gamma=0.56$), and A of HY-World 1.5, which accelerates ($\gamma=2.53$). The power-law fits describe the shape of the response, and \mHold\ uses the strides of Eqs.~\eqref{eq:2} and~\eqref{eq:3}.
\begin{figure}[!ht]
\centering
\includegraphics[width=\linewidth,height=0.74\textheight,keepaspectratio]{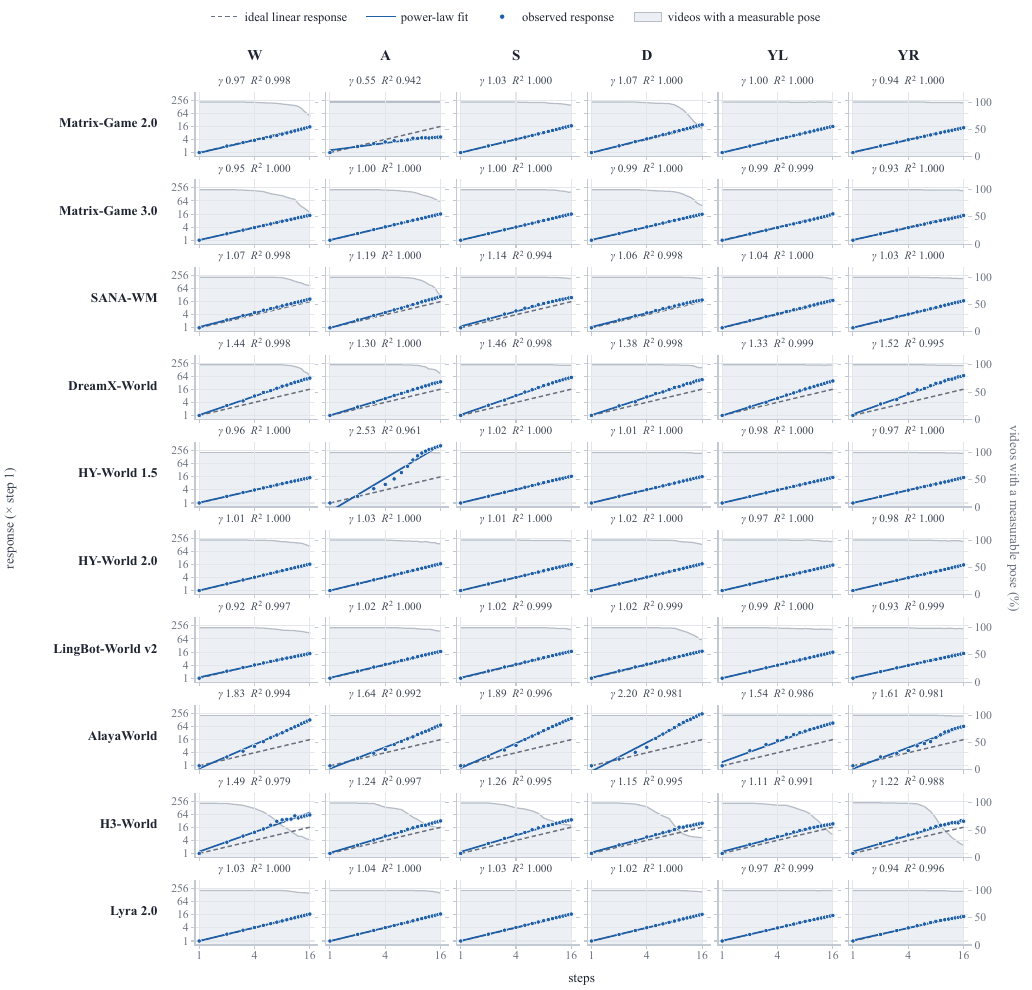}
\caption{Single-action response profiles of ten of the fifteen models across the six actions (Fig.~\ref{fig:D.response_new} shows the other five). Each video's response is divided by its first-step response, and the median is taken over the videos measurable through all $16$ steps. Annotations report the exponent $\gamma$ of a log--log least-squares fit of $m(k)\propto k^\gamma$ over steps $1$--$16$, where $k$ is the execution step, and its $R^2$. Blue curves show the fits, grey dashed lines ideal linear responses ($\gamma=1$), and the shaded grey area the share of videos with a measurable pose (right axis).}
\label{fig:D.response}
\end{figure}

\begin{figure}[!ht]
\centering
\includegraphics[width=\linewidth]{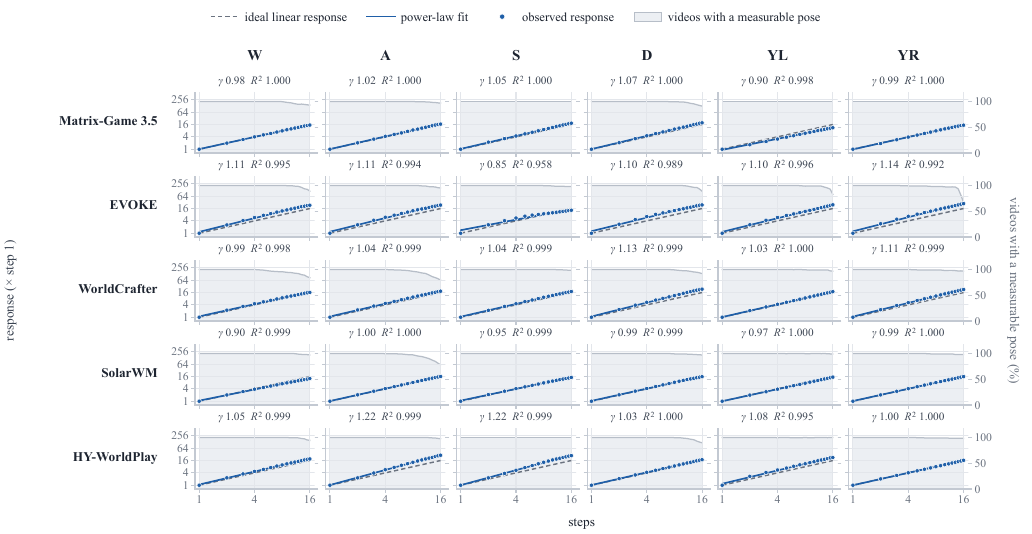}
\caption{Continuation of Fig.~\ref{fig:D.response} for Matrix-Game 3.5, EVOKE, WorldCrafter, SolarWM, and HY-WorldPlay, with the same normalization, fit, and annotations ($283$--$420$ videos per panel).}
\label{fig:D.response_new}
\end{figure}

\paragraph{Measurable videos and video length.}\label{app:length}
\mHold\ assigns zero to evaluation points whose pose cannot be recovered, and on the single-action videos measurable at both steps 12 and 16, neither the drop of \mHold\ from step 12 to step 16 nor the step-16 score is related to the number of frames per 16 steps (Spearman $\rho=0.31$, $p=0.25$, and $\rho=-0.17$, $p=0.54$, over $15$ models).

\FloatBarrier
\section{Further Results on Composition}\label{app:res_f3}
\subsection{Results by reference domain}\label{app:H2}
As shown in Tab.~\ref{tab:H.3}, HY-World 2.0, Lyra 2.0, and SolarWM take the first three places in \mComp\ on both the rendered (UE5, $104$ views) and the real-photo (ETH3D, $36$ views) start views, while the domain moves \mHold\ in both directions across models. On the completed pairs alone, the \mPair\ score exceeds the \mRoute\ score for every model, by $1.3$ (DreamX-World) to $17.7$ (WorldCrafter) points in position, and HY-World 2.0 and Lyra 2.0 hold the top two places in both.

\begin{table}[H]
\centering
\caption{\textbf{\mHold\ and \mComp\ on the rendered (UE5) and real-photo (ETH3D) start views.} Main configuration and strides, with single-action weights renormalized within each domain. Bold and underline: best and second best in each column.}
\label{tab:H.3}
\begingroup\footnotesize\setlength{\tabcolsep}{4pt}\renewcommand{\arraystretch}{1.08}
\begin{tabular*}{0.8\linewidth}{@{\extracolsep{\fill}}lrrrr@{}}
\toprule
& \multicolumn{2}{c}{UE5} & \multicolumn{2}{c}{ETH3D} \\
\cmidrule(lr){2-3}\cmidrule(lr){4-5}
Model & Pace & Comp. & Pace & Comp. \\
\midrule
Matrix-Game 3.0 & 80.4 & 62.9 & 63.7 & 48.9 \\
Matrix-Game 2.0 & 66.9 & 53.0 & 56.3 & 40.0 \\
SANA-WM & 78.0 & 28.8 & 69.3 & 26.9 \\
DreamX-World & 46.0 & 24.0 & 48.2 & 24.6 \\
HY-World 1.5 & 74.1 & 17.2 & \textbf{91.9} & 35.3 \\
HY-World 2.0 & \underline{85.6} & \textbf{80.6} & \underline{91.6} & \textbf{90.4} \\
Lyra 2.0 & \textbf{87.9} & \underline{77.0} & 88.8 & \underline{80.9} \\
SolarWM & 84.8 & 65.6 & 87.6 & 67.4 \\
WorldCrafter & 82.6 & 57.4 & 70.3 & 55.2 \\
LingBot-World v2 & 82.1 & 52.9 & 84.6 & 59.7 \\
HY-WorldPlay & 72.8 & 36.9 & 83.2 & 42.9 \\
EVOKE & 74.0 & 31.3 & 73.4 & 37.4 \\
Matrix-Game 3.5 & 63.1 & 18.4 & 75.7 & 33.2 \\
AlayaWorld & 15.8 & 11.2 & 35.7 & 10.3 \\
H3-World & 25.2 & 7.2 & 23.6 & 8.7 \\
\bottomrule
\end{tabular*}
\endgroup
\end{table}

\FloatBarrier
\subsection{Composition step by step}\label{app:f3steps}
\begin{figure*}[t]
    \centering
    \includegraphics[width=\textwidth]{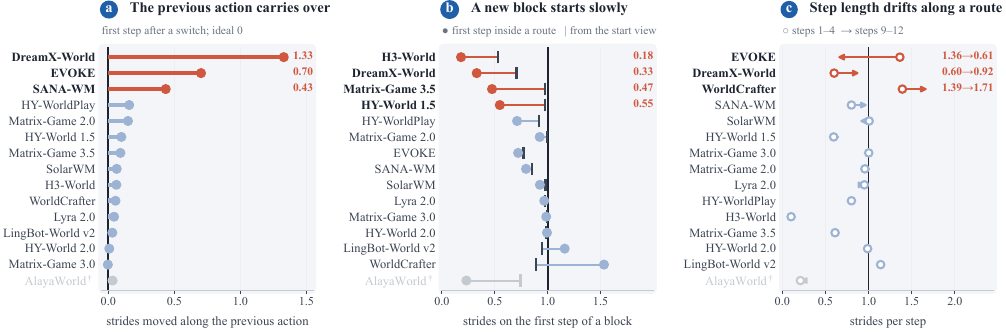}
    \caption{\textbf{Step-by-step decomposition of composed routes}, showing median displacement per translation step along the commanded action in units of that action's stride. (a)~Motion along the new and previous actions on the first step after an orthogonal switch. (b)~Starting steps of blocks of at least four identical actions versus the first step held from the start view. (c)~Early versus late steps along routes of at least twelve steps.}
    \label{fig:anatomy}\label{fig:whymiss}
\end{figure*}
Fig.~\ref{fig:anatomy} and Tab.~\ref{tab:f3steps} use every route whose camera is recovered at all step boundaries and whose strides are calibrated, including routes in which a step stalls ($742$ to $1{,}120$ routes per model). For step $j$ of a route, $\Delta\mathbf H_j=\mathbf H(b_{j-1})^{-1}\mathbf H(b_j)$ is its motion in the camera frame at the start of the step, and $\rho_j=\langle\mathbf t(\Delta\mathbf H_j),\mathbf t(\mathbf U_a)\rangle/\|\mathbf t(\mathbf U_a)\|^2$ is its motion along the commanded axis in strides (the yaw ratio for a turn), with $\mathbf U_a$ the seed-0 stride of the same view. A step is misdirected if it passes the motion check of the scoring pipeline but $\rho_j<0.1$. For the first step after a switch between orthogonal translations, we solve $\mathbf t(\Delta\mathbf H_j)=c_{\text{new}}\,\mathbf t(\mathbf U_{a})+c_{\text{prev}}\,\mathbf t(\mathbf U_{a'})$, where $a'$ is the previous action and $c_{\text{prev}}$ is the motion carried over (App.~\ref{app:D3}).

As shown in Tab.~\ref{tab:f3steps}, the misdirected switches of DreamX-World and EVOKE are mostly carry-over. In $86\%$ of DreamX-World's orthogonal switches the camera keeps moving at least half a stride along the previous action, and this carry-over lasts through the second step after the switch, while Lyra 2.0 carries over only in the first frame and HY-World 2.0 not at all. HY-World 1.5 and Matrix-Game 3.5 instead start each new block slowly, with a held translation beginning at $0.98$ strides but a block in a route at $0.55$ and $0.47$. A step model built from these average behaviors on one half of the scenes predicts EVOKE's swap gap on the other half ($2.9$ against an observed $2.5$ strides) and a third of DreamX-World's ($1.7$ against $4.9$). The remaining error is specific to each route and recurs when the route is generated again with other seeds (App.~\ref{app:seeds}).

\begin{table*}[!htbp]
\centering
\caption{\textbf{Composition step by step (translation steps; medians in strides of the step's action).} New and Carried: first step after a switch between orthogonal translations, motion along the new and the previous action ($95\%$ scene-bootstrap interval). Held: first step of the same translations held from the start view (single-action videos), along their own stride. Block: steps 1, 2, and 4 of the same blocks of at least four equal actions. Route: steps 1--4 and 9--12 of the same routes of at least twelve steps. Misdir.\ and Stall: share of first steps after a switch that are misdirected or that stall (\%), and in brackets the difference from later steps of a block at the same route position (points). $^\dagger$Stride not reliably measurable.}
\label{tab:f3steps}
\begingroup\footnotesize\setlength{\tabcolsep}{3.5pt}
\renewcommand{\arraystretch}{1.08}\begin{tabular*}{\textwidth}{@{\extracolsep{\fill}}lccccccccccc@{}}
\toprule
 & & \multicolumn{2}{c}{After a switch} & Held & \multicolumn{3}{c}{Block} & \multicolumn{2}{c}{Route} & \multicolumn{2}{c}{Switch steps (\%)} \\
\cmidrule(lr){3-4}\cmidrule(lr){5-5}\cmidrule(lr){6-8}\cmidrule(lr){9-10}\cmidrule(lr){11-12}
Model & Routes & New & Carried & 1st & 1st & 2nd & 4th & 1--4 & 9--12 & Misdir. & Stall \\
\midrule
Matrix-Game 3.0 & 1063 & 0.90 & -0.00 [-0.01, 0.01] & 0.99 & 0.99 & 1.03 & 0.92 & 1.00 & 0.94 & 1 (-0) & 0 (-0) \\
Matrix-Game 2.0 & 1098 & 0.90 & 0.15 [0.13, 0.17] & 1.00 & 0.92 & 1.01 & 0.99 & 0.96 & 1.02 & 13 (+5) & 1 (+0) \\
SANA-WM & 1108 & 0.62 & 0.43 [0.39, 0.49] & 0.85 & 0.80 & 1.21 & 1.09 & 0.80 & 1.02 & 17 (+14) & 7 (+5) \\
DreamX-World & 1120 & 0.36 & 1.33 [1.25, 1.43] & 0.70 & 0.33 & 0.72 & 2.03 & 0.60 & 0.92 & 46 (+35) & 1 (+0) \\
HY-World 1.5 & 1120 & 0.46 & 0.10 [0.07, 0.13] & 0.98 & 0.55 & 0.82 & 0.81 & 0.60 & 0.70 & 23 (+17) & 11 (+4) \\
HY-World 2.0 & 1113 & 1.00 & 0.01 [0.00, 0.01] & 0.98 & 0.99 & 1.01 & 1.01 & 0.99 & 1.00 & 1 (-0) & 0 (-0) \\
Lyra 2.0 & 1120 & 0.94 & 0.04 [0.04, 0.05] & 0.98 & 0.97 & 1.01 & 1.01 & 0.95 & 0.99 & 1 (+0) & 0 (+0) \\
SolarWM & 1117 & 0.89 & 0.06 [0.05, 0.08] & 0.98 & 0.93 & 0.98 & 0.90 & 1.01 & 0.87 & 1 (+0) & 0 (-0) \\
WorldCrafter & 1098 & 1.49 & 0.05 [0.03, 0.08] & 0.89 & 1.53 & 1.60 & 1.52 & 1.39 & 1.71 & 1 (-0) & 0 (+0) \\
LingBot-World v2 & 1057 & 0.99 & 0.03 [0.02, 0.04] & 0.95 & 1.16 & 1.22 & 1.14 & 1.14 & 1.14 & 2 (-0) & 0 (+0) \\
HY-WorldPlay & 1109 & 0.66 & 0.16 [0.11, 0.21] & 0.92 & 0.71 & 0.99 & 0.80 & 0.80 & 0.82 & 20 (+11) & 1 (+1) \\
EVOKE & 1110 & 0.45 & 0.70 [0.64, 0.77] & 0.77 & 0.72 & 1.22 & 1.04 & 1.36 & 0.61 & 27 (+20) & 4 (+4) \\
Matrix-Game 3.5 & 1120 & 0.38 & 0.09 [0.06, 0.12] & 0.98 & 0.47 & 0.89 & 0.68 & 0.61 & 0.62 & 17 (+9) & 18 (+9) \\
AlayaWorld$^\dagger$ & 1068 & 0.22 & 0.03 [0.00, 0.10] & 0.74 & 0.23 & 0.23 & 0.40 & 0.21 & 0.18 & 40 (+4) & 8 (+0) \\
H3-World & 742 & 0.11 & 0.06 [0.04, 0.11] & 0.53 & 0.18 & 0.26 & 0.26 & 0.10 & 0.09 & 43 (+8) & 9 (+1) \\
\bottomrule

\end{tabular*}
\endgroup
\end{table*}

\FloatBarrier
\subsection{Regeneration error}\label{app:f3diag}
Since each route is a generation of its own, we compare its end error with the \emph{regeneration error}, which we estimate from the single-action videos alone. For each completed route $q$, we chain the strides of the same view calibrated on seed $s\in\{1,2\}$ in place of seed $0$ and measure how far this end lies from the seed-0 prediction, $\|\mathbf t(\widehat{\mathbf H}_q^{-1}\widehat{\mathbf H}_q^{(s)})\|/b_t$, averaged over the two seeds. For swap pairs (G2--G4), the expected gap chains the strides of route $q$ from one seed and those of $q'$ from another, $\|\mathbf t(\widehat{\mathbf H}_q^{(s)-1}\widehat{\mathbf H}_{q'}^{(s')})\|/b_t$, median over the seed pairs $(0,1)$, $(0,2)$, and $(1,2)$. Chaining uses $\mathbf U_a$ of Eq.~\eqref{eq:2}, as the scoring pipeline does.

As shown in Tab.~\ref{tab:f3diag}, the observed end error exceeds the regeneration error for every model, by $1.4$ to $9.0$ times. Across the fourteen models whose strides are above the measurement noise, the ratio of observed to expected swap gap follows the carry-over of Tab.~\ref{tab:f3steps} (Spearman $\rho=0.59$, permutation $p=0.03$), as plotted in Fig.~\ref{fig:regen}.

\begin{table}[!htbp]
\centering
\caption{\textbf{End error against regeneration error.} Median end error of completed routes and median regeneration error, in strides, with their ratio, and the observed and expected swap gap between two orders of the same actions. $^\dagger$Stride not reliably measurable.}
\label{tab:f3diag}
\begingroup\footnotesize\setlength{\tabcolsep}{4pt}
\renewcommand{\arraystretch}{1.08}\begin{tabular*}{0.84\linewidth}{@{\extracolsep{\fill}}lccccc@{}}
\toprule
 & \multicolumn{3}{c}{End error} & \multicolumn{2}{c}{Swap gap} \\
\cmidrule(lr){2-4}\cmidrule(lr){5-6}
Model & observed & regen. & ratio & observed & expected \\
\midrule
Matrix-Game 3.0 & 0.67 & 0.34 & 1.9 & 0.41 & 0.33 \\
Matrix-Game 2.0 & 1.45 & 0.34 & 4.3 & 1.14 & 0.36 \\
SANA-WM & 2.47 & 0.94 & 2.6 & 1.95 & 1.04 \\
DreamX-World & 5.22 & 1.83 & 2.8 & 4.92 & 1.68 \\
HY-World 1.5 & 2.42 & 1.77 & 1.4 & 1.56 & 1.65 \\
HY-World 2.0 & 0.30 & 0.15 & 2.0 & 0.10 & 0.15 \\
Lyra 2.0 & 0.51 & 0.21 & 2.4 & 0.34 & 0.22 \\
SolarWM & 1.61 & 0.78 & 2.1 & 0.77 & 0.84 \\
WorldCrafter & 2.38 & 0.87 & 2.7 & 0.88 & 0.85 \\
LingBot-World v2 & 2.32 & 1.03 & 2.3 & 0.78 & 0.96 \\
HY-WorldPlay & 3.44 & 2.16 & 1.6 & 1.71 & 2.02 \\
EVOKE & 4.57 & 0.79 & 5.8 & 2.59 & 0.81 \\
Matrix-Game 3.5 & 2.13 & 0.85 & 2.5 & 1.10 & 0.85 \\
AlayaWorld$^\dagger$ & 50.57 & 5.64 & 9.0 & 64.76 & 5.30 \\
H3-World & 9.02 & 5.52 & 1.6 & 4.52 & 5.23 \\
\bottomrule
\end{tabular*}
\endgroup
\end{table}

\begin{figure}[!htbp]
\centering
\includegraphics{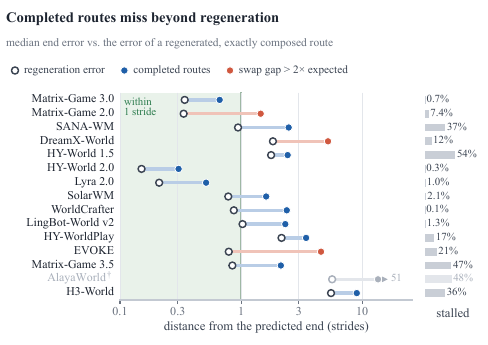}
\caption{\textbf{End error and regeneration error of completed routes.} Dots show the end error and circles the regeneration error, in strides, with rows as in Tab.~\ref{tab:main}. The axis is logarithmic, so the bar shows the ratio. Red: the median swap gap is more than twice the gap expected from regeneration (Tab.~\ref{tab:f3diag}). Right: share of routes with a stalled step. $^\dagger$Stride not reliably measurable.}
\label{fig:regen}
\end{figure}

\FloatBarrier
\subsection{Repeated route generations}\label{app:seeds}
To test whether route errors repeat, we generate the routes again with seeds $1$ and $2$ for all fifteen models. Each model regenerates all eight routes of the first view of every scene, which gives $240$ routes per seed, and each seed's routes are scored with the strides calibrated on the single actions of the same seed, so the seeds are evaluated independently with the unchanged pipeline. For DreamX-World, SANA-WM, LingBot-World v2, and AlayaWorld, seed $0$ is generated again together with seeds $1$ and $2$. As shown in Tab.~\ref{tab:seeds}, the \mComp\ of the median model varies by $2.2$ points across the three seeds and by at most $8.2$ points for any model, which is small against the differences between models. The order of the fourteen ranked models is nearly the same under every seed, with Kendall's $\tau$ of $0.96$ to $0.98$ between seeds. Only two pairs of neighbouring models in the lower part of the ranking change places, SANA-WM with DreamX-World and HY-World 1.5 with Matrix-Game 3.5, and their scores differ by at most $4.0$ points under any seed.

To test whether the errors of single routes are systematic, we place the ends of the three generations of a route in the frame of the start camera and measure their spread as the root mean square of the pairwise distances in strides of seed $0$. As shown in Tab.~\ref{tab:seeds_spread}, the end error of a route is a median $1.8$ to $3.7$ times this spread, and the errors of different seeds mostly point in similar directions, with vector correlations of $0.27$ to $0.94$ between seeds. For $42\%$ to $86\%$ of the routes the errors of both new seeds lie within $45^\circ$ of the direction of the seed-$0$ error, and $53\%$ to $83\%$ of all routes end with the same outcome under all three seeds. These ranges cover the thirteen ranked models with at least $40$ routes completed under all three seeds. Stalls also repeat, since every ranked model with frequent stalls except H3-World has more routes that stall under both seeds of a pair than expected if stalls were independent of the route, for example $22$ to $29$ against $10$ to $12$ for EVOKE. The observed spread lies within a factor of three of the regeneration error of App.~\ref{app:f3diag} for every model that meets the threshold.

\begin{table}[!htbp]
\centering
\caption{\textbf{Route scores under three seeds ($240$ routes per model and seed, first view of each scene).} Seed $0$ is the run of Tab.~\ref{tab:main} restricted to these routes. $^{\S}$Seed $0$ is generated again together with seeds $1$ and $2$. Each seed is scored with its own strides. Grey: not ranked, as in Tab.~\ref{tab:main}. Metrics as in Tab.~\ref{tab:main}.}
\label{tab:seeds}
\begingroup\scriptsize\setlength{\tabcolsep}{4pt}\renewcommand{\arraystretch}{1.0}
\begin{tabular*}{\linewidth}{@{\extracolsep{\fill}}llccccc@{}}
\toprule
Model & Seed & Stall\,(\%) & Near\,(\%) & End err.\,(str.) & Swap gap\,(str.) & \mComp \\
\midrule
Matrix-Game 3.0 & 0 & 0.8 & 50.4 & 0.60 & 0.41 & 60.3 \\
 & 1 & 0.0 & 59.6 & 0.61 & 0.38 & 63.8 \\
 & 2 & 0.4 & 50.8 & 0.69 & 0.37 & 62.8 \\
\midrule
Matrix-Game 2.0 & 0 & 9.6 & 25.8 & 1.40 & 1.14 & 49.7 \\
 & 1 & 8.8 & 23.8 & 1.27 & 1.05 & 49.6 \\
 & 2 & 7.9 & 21.2 & 1.46 & 1.16 & 49.5 \\
\midrule
SANA-WM$^{\S}$ & 0 & 37.5 & 4.6 & 2.77 & 2.07 & 26.1 \\
 & 1 & 37.5 & 5.8 & 2.96 & 1.99 & 27.0 \\
 & 2 & 38.3 & 7.1 & 2.77 & 1.82 & 27.1 \\
\midrule
DreamX-World$^{\S}$ & 0 & 10.0 & 0.8 & 7.02 & 6.52 & 22.1 \\
 & 1 & 14.2 & 2.5 & 3.99 & 4.24 & 30.3 \\
 & 2 & 12.9 & 1.7 & 4.21 & 3.91 & 28.6 \\
\midrule
HY-World 1.5 & 0 & 58.8 & 5.0 & 2.06 & 1.72 & 18.9 \\
 & 1 & 60.4 & 7.5 & 2.38 & 2.15 & 19.6 \\
 & 2 & 59.2 & 4.2 & 2.93 & 2.63 & 19.6 \\
\midrule
HY-World 2.0 & 0 & 0.4 & 86.2 & 0.28 & 0.08 & 85.0 \\
 & 1 & 0.8 & 85.8 & 0.28 & 0.08 & 84.6 \\
 & 2 & 0.4 & 86.7 & 0.28 & 0.09 & 85.9 \\
\midrule
Lyra 2.0 & 0 & 0.4 & 75.0 & 0.52 & 0.34 & 78.9 \\
 & 1 & 1.2 & 79.6 & 0.43 & 0.31 & 79.9 \\
 & 2 & 0.4 & 83.8 & 0.43 & 0.31 & 79.8 \\
\midrule
SolarWM & 0 & 2.5 & 20.4 & 1.66 & 0.79 & 65.6 \\
 & 1 & 0.8 & 24.6 & 1.55 & 0.66 & 67.5 \\
 & 2 & 0.0 & 17.1 & 2.02 & 0.84 & 65.6 \\
\midrule
WorldCrafter & 0 & 0.0 & 16.2 & 2.12 & 0.84 & 58.6 \\
 & 1 & 0.4 & 20.8 & 1.94 & 0.71 & 59.5 \\
 & 2 & 0.0 & 20.0 & 1.88 & 0.83 & 59.6 \\
\midrule
LingBot-World v2$^{\S}$ & 0 & 2.5 & 13.3 & 2.21 & 0.63 & 50.9 \\
 & 1 & 0.4 & 9.2 & 2.86 & 1.13 & 54.8 \\
 & 2 & 0.8 & 11.2 & 2.73 & 0.78 & 58.0 \\
\midrule
HY-WorldPlay & 0 & 17.1 & 2.5 & 3.47 & 1.65 & 38.9 \\
 & 1 & 16.7 & 5.8 & 3.05 & 2.23 & 40.0 \\
 & 2 & 20.0 & 6.2 & 2.59 & 2.16 & 37.5 \\
\midrule
EVOKE & 0 & 20.4 & 3.3 & 4.14 & 2.12 & 33.6 \\
 & 1 & 20.4 & 4.6 & 4.69 & 2.31 & 35.7 \\
 & 2 & 24.2 & 1.2 & 4.68 & 2.84 & 31.6 \\
\midrule
Matrix-Game 3.5 & 0 & 48.8 & 12.5 & 1.75 & 1.17 & 22.0 \\
 & 1 & 54.2 & 10.4 & 1.65 & 1.04 & 18.9 \\
 & 2 & 51.7 & 6.7 & 1.90 & 0.87 & 19.7 \\
\midrule
\textcolor{gray}{AlayaWorld$^{\S}$} & \textcolor{gray}{0} & \textcolor{gray}{52.1} & \textcolor{gray}{0.0} & \textcolor{gray}{111.98} & \textcolor{gray}{79.48} & \textcolor{gray}{10.5} \\
 & \textcolor{gray}{1} & \textcolor{gray}{48.3} & \textcolor{gray}{0.0} & \textcolor{gray}{69.14} & \textcolor{gray}{74.95} & \textcolor{gray}{12.2} \\
 & \textcolor{gray}{2} & \textcolor{gray}{51.2} & \textcolor{gray}{0.4} & \textcolor{gray}{93.50} & \textcolor{gray}{93.25} & \textcolor{gray}{10.5} \\
\midrule
H3-World & 0 & 35.8 & 0.4 & 9.60 & 3.38 & 9.0 \\
 & 1 & 22.9 & 0.4 & 10.06 & 2.60 & 6.0 \\
 & 2 & 31.7 & 0.0 & 12.83 & 8.61 & 10.1 \\
\bottomrule
\end{tabular*}
\endgroup
\end{table}

\begin{table}[!htbp]
\centering
\caption{\textbf{Spread of route ends across three seeds and agreement of their errors.} $n$: routes completed under all three seeds. Spread: root mean square of the pairwise distances between the three ends (strides of seed $0$, median and 90th percentile). Err.: median end error of these routes. Ratio: median of error over spread per route. Regen.: regeneration error of App.~\ref{app:f3diag} on the same routes. Corr.: vector correlation of the end errors between seed pairs (range). ${\le}45^\circ$: share of routes whose errors under seeds $1$ and $2$ both lie within $45^\circ$ of the seed-$0$ error (\%). Agree: share of all $240$ routes with the same outcome (stalled, near, far, or unmeasurable) under all three seeds (\%). $^{\S}$As in Tab.~\ref{tab:seeds}. Grey: not ranked.}
\label{tab:seeds_spread}
\begingroup\scriptsize\setlength{\tabcolsep}{2.5pt}\renewcommand{\arraystretch}{1.08}
\begin{tabular*}{\linewidth}{@{\extracolsep{\fill}}lcccccccccc@{}}
\toprule
 & & \multicolumn{2}{c}{Spread} & & & & & & & \\
\cmidrule(lr){3-4}
Model & $n$ & Med. & P90 & Err. & Ratio & Regen. & Corr. & ${\le}45^\circ$ & Agree \\
\midrule
Matrix-Game 3.0 & 221 & 0.30 & 0.71 & 0.59 & 2.10 & 0.34 & 0.79--0.82 & 59 & 66 \\
Matrix-Game 2.0 & 190 & 0.51 & 1.58 & 1.27 & 2.44 & 0.26 & 0.82--0.85 & 78 & 59 \\
SANA-WM$^{\S}$ & 109 & 1.36 & 3.36 & 2.77 & 1.75 & 0.88 & 0.75--0.79 & 62 & 61 \\
DreamX-World$^{\S}$ & 181 & 2.26 & 4.63 & 4.50 & 2.13 & 1.79 & 0.93--0.94 & 78 & 74 \\
HY-World 1.5 & 49 & 0.90 & 2.16 & 2.00 & 1.91 & 1.50 & 0.27--0.59 & 49 & 53 \\
HY-World 2.0 & 236 & 0.11 & 0.72 & 0.27 & 2.02 & 0.12 & 0.55--0.64 & 57 & 83 \\
Lyra 2.0 & 237 & 0.24 & 0.86 & 0.48 & 2.07 & 0.20 & 0.68--0.74 & 62 & 78 \\
SolarWM & 233 & 0.47 & 1.10 & 1.67 & 3.34 & 0.72 & 0.51--0.60 & 63 & 70 \\
WorldCrafter & 236 & 0.56 & 1.50 & 1.93 & 3.70 & 0.96 & 0.93--0.94 & 86 & 78 \\
LingBot-World v2$^{\S}$ & 204 & 0.87 & 2.31 & 2.57 & 3.02 & 1.21 & 0.73--0.77 & 70 & 67 \\
HY-WorldPlay & 161 & 1.65 & 3.23 & 3.10 & 1.80 & 2.37 & 0.52--0.64 & 42 & 62 \\
EVOKE & 144 & 1.83 & 4.60 & 4.50 & 2.39 & 0.79 & 0.67--0.72 & 78 & 62 \\
Matrix-Game 3.5 & 71 & 0.69 & 3.97 & 1.52 & 2.17 & 0.58 & 0.72--0.81 & 72 & 53 \\
\textcolor{gray}{AlayaWorld$^{\S}$} & \textcolor{gray}{75} & \textcolor{gray}{33.57} & \textcolor{gray}{405.43} & \textcolor{gray}{128.73} & \textcolor{gray}{1.73} & \textcolor{gray}{6.08} & \textcolor{gray}{0.76--0.83} & \textcolor{gray}{65} & \textcolor{gray}{70} \\
H3-World & 13 & 14.38 & 482.82 & 10.19 & 0.62 & 4.79 & -0.05--0.10 & 15 & 13 \\
\bottomrule
\end{tabular*}
\endgroup
\end{table}

\FloatBarrier
\subsection{What a fitted scale absorbs}\label{app:scale}
Trajectory metrics that fit a scale per video or segment absorb any error that changes only the length of a route. Fitted to each completed route, $s=\arg\min_{s\ge0}\|s\,\mathbf t(\widehat{\mathbf H}_q)-\mathbf t(\mathbf H_q)\|$ is a projection coefficient, and it equals the \emph{distance ratio} $\|\mathbf t(\mathbf H_q)\|/\|\mathbf t(\widehat{\mathbf H}_q)\|$ only when the route ends in its predicted direction. As shown in Tab.~\ref{tab:scale}, DreamX-World ends its routes at a median $2.31$ times the predicted distance and its fitted scale ($1.89$) absorbs most of this, while HY-World 1.5, Matrix-Game 3.5, and HY-WorldPlay travel only $0.79$, $0.82$, and $0.68$ times as far. The fit also lowers the end error of HY-World 2.0 and Lyra 2.0, whose median scale is $1.00$ and $1.01$, since it removes any error along the predicted direction, so the fitted scale and the distance ratio reveal systematic length errors and the reduction of the end error does not.

\begin{table}[!htbp]
\centering
\caption{\textbf{What a fitted scale absorbs.} Median end error of completed routes, in strides, before and after fitting one scale per route, with the median fitted scale and the median distance ratio, the observed over the predicted distance from the start. $^\dagger$Stride not reliably measurable.}
\label{tab:scale}
\begingroup\footnotesize\setlength{\tabcolsep}{5pt}
\renewcommand{\arraystretch}{1.08}\begin{tabular*}{0.74\linewidth}{@{\extracolsep{\fill}}lcccc@{}}
\toprule
Model & End err. & Scale-aligned & Fitted scale & Distance ratio \\
\midrule
Matrix-Game 3.0 & 0.67 & 0.28 & 0.92 & 0.95 \\
Matrix-Game 2.0 & 1.45 & 0.79 & 0.96 & 1.02 \\
SANA-WM & 2.47 & 1.29 & 1.08 & 1.22 \\
DreamX-World & 5.22 & 2.36 & 1.89 & 2.31 \\
HY-World 1.5 & 2.42 & 1.16 & 0.68 & 0.79 \\
HY-World 2.0 & 0.30 & 0.14 & 1.00 & 1.01 \\
Lyra 2.0 & 0.51 & 0.23 & 1.01 & 1.01 \\
SolarWM & 1.61 & 0.76 & 0.81 & 0.88 \\
WorldCrafter & 2.38 & 1.26 & 1.21 & 1.35 \\
LingBot-World v2 & 2.32 & 1.12 & 0.82 & 1.03 \\
HY-WorldPlay & 3.44 & 1.39 & 0.50 & 0.68 \\
EVOKE & 4.57 & 1.94 & 0.64 & 1.04 \\
Matrix-Game 3.5 & 2.13 & 0.82 & 0.74 & 0.82 \\
AlayaWorld$^\dagger$ & 50.57 & 26.27 & 0.66 & 7.57 \\
H3-World & 9.02 & 5.15 & 0.41 & 1.74 \\
\bottomrule
\end{tabular*}
\endgroup
\end{table}

\subsection{Observed held motion as the predictor}\label{app:profile}
We replace $\mathbf U_a^n$ in the route prediction by the observed held motion $\mathbf H_a(n)$, the planar pose after $n$ held steps in the single-action video of the same model, start view, and seed ($H_0$), or by its Lie-algebra mean over the three seeds of that view ($H_{0:2}$), which also removes most of the regeneration noise. Since route blocks have at most four steps, these predictions contain any slow start from rest. As shown in Tab.~\ref{tab:profile}, with $H_0$ the median end error changes by $-6\%$ to $+6\%$, and with $H_{0:2}$ it falls by at most $26\%$ (HY-WorldPlay) and by less than $6\%$ for eleven of the fifteen models, while \mComp\ changes by at most $1.8$ points wherever all three seeds are available. The errors of composed routes therefore do not stem from held-action dynamics that a constant stride misses.

\begin{table}[!htbp]
\centering
\caption{\textbf{Route predictions from strides and from observed held motion (seed $0$ routes).} End err.: median end error of completed routes (strides). $H_0$: held motion of the same seed. $H_{0:2}$: its mean over three seeds. $^\ast$Three-seed held motion is missing for $2$ (HY-World 2.0) and $43$ (H3-World) completed routes, so the end error is over the available routes and the \mComp\ is given as the range over all possible values of the missing ones. $^\dagger$Stride not reliably measurable.}
\label{tab:profile}
\begingroup\footnotesize\setlength{\tabcolsep}{3.5pt}
\renewcommand{\arraystretch}{1.08}\begin{tabular*}{0.8\linewidth}{@{\extracolsep{\fill}}lccccc@{}}
\toprule
 & \multicolumn{3}{c}{End err.} & \multicolumn{2}{c}{\mComp} \\
\cmidrule(lr){2-4}\cmidrule(lr){5-6}
Model & Stride & $H_0$ & $H_{0:2}$ & Stride & $H_{0:2}$ \\
\midrule
Matrix-Game 3.0 & 0.67 & 0.67 & 0.65 & 59.3 & 60.4 \\
Matrix-Game 2.0 & 1.45 & 1.46 & 1.43 & 49.6 & 49.7 \\
SANA-WM & 2.47 & 2.42 & 2.39 & 28.3 & 28.6 \\
DreamX-World & 5.22 & 4.92 & 4.48 & 24.2 & 24.9 \\
HY-World 1.5 & 2.42 & 2.38 & 2.03 & 21.8 & 22.4 \\
HY-World 2.0 & 0.30 & 0.30 & 0.30$^\ast$ & 83.1 & 83.0--83.2$^\ast$ \\
Lyra 2.0 & 0.51 & 0.53 & 0.52 & 78.0 & 78.0 \\
SolarWM & 1.61 & 1.59 & 1.61 & 66.1 & 66.1 \\
WorldCrafter & 2.38 & 2.41 & 2.28 & 56.8 & 56.9 \\
LingBot-World v2 & 2.32 & 2.34 & 2.24 & 54.6 & 55.1 \\
HY-WorldPlay & 3.44 & 3.39 & 2.54 & 38.4 & 40.2 \\
EVOKE & 4.57 & 4.60 & 4.69 & 32.9 & 32.8 \\
Matrix-Game 3.5 & 2.13 & 2.12 & 2.03 & 22.2 & 22.2 \\
AlayaWorld$^\dagger$ & 50.57 & 53.66 & 51.79 & 11.0 & 11.0 \\
H3-World & 9.02 & 9.03 & 7.99$^\ast$ & 7.6 & 7.1--10.0$^\ast$ \\
\bottomrule
\end{tabular*}
\endgroup
\end{table}

\subsection{Existing control metrics on the same videos}\label{app:existing}
We compute two kinds of existing control metrics on the seed-$0$ videos of Tab.~\ref{tab:main}. Direction accuracy is \mResp, the share of single-action videos that move in the commanded direction. A trajectory error with a fitted reference aligns the recovered camera positions of a video to its predicted path, with one scale or with a similarity transform estimated by the Umeyama method, and divides the root mean square position error over the step boundaries by the length of the predicted path. As shown in Tab.~\ref{tab:existing}, direction accuracy exceeds $98\%$ for eleven of the fifteen models and does not rank composition, with a Spearman correlation of $0.02$ with \mComp\ over the fourteen ranked models. Even when given our stride-predicted path as the reference, which existing benchmarks do not have for key-driven models, fitted trajectory errors rate many failed routes as accurate. After fitting one scale, $20\%$ of the stalled routes of the fourteen ranked models and $24\%$ of their completed routes that end more than two strides from their predictions have an error below $10\%$ of the path, and after fitting a similarity transform $53\%$ and $63\%$. With the calibrated strides kept as a fixed reference instead of fitting a scale per video, the trajectory error at the end of a route equals End err.\ of Tab.~\ref{tab:main} up to the normalization by path length rather than stride, and its per-step version compares every step with its stride, as the stall check and the pace ratio do, and Tab.~\ref{tab:humanmetrics} compares these variants with human judgments.

\begin{table}[!htbp]
\centering
\caption{\textbf{Existing control metrics on the seed-$0$ videos.} Resp.: direction accuracy (\%). ATE: median position error after fitting one scale or a similarity transform (Sim(3)), in \% of the predicted path, for routes and for $16$-step held actions. Rated accurate: routes with an error below $10\%$ after fitting one scale among the stalled routes and among the completed routes ending more than two strides from their predictions. $^\dagger$Stride not reliably measurable.}
\label{tab:existing}
\begingroup\footnotesize\setlength{\tabcolsep}{3.5pt}
\renewcommand{\arraystretch}{1.08}\begin{tabular*}{\linewidth}{@{\extracolsep{\fill}}lccccccc@{}}
\toprule
 & & \multicolumn{2}{c}{Route ATE} & \multicolumn{2}{c}{Held ATE} & \multicolumn{2}{c}{Rated accurate} \\
\cmidrule(lr){3-4}\cmidrule(lr){5-6}\cmidrule(lr){7-8}
Model & Resp. & Scale & Sim(3) & Scale & Sim(3) & Stalled & End err.\ $>2$ \\
\midrule
Matrix-Game 3.0 & 99.6 & 4.1 & 2.7 & 10.1 & 4.8 & 6/8 & 62/103 \\
Matrix-Game 2.0 & 84.3 & 9.5 & 5.6 & 19.7 & 10.8 & 38/83 & 78/309 \\
SANA-WM & 99.3 & 11.0 & 7.4 & 11.5 & 6.0 & 160/409 & 131/430 \\
DreamX-World & 99.5 & 13.5 & 9.4 & 14.9 & 7.0 & 21/138 & 135/849 \\
HY-World 1.5 & 99.8 & 14.4 & 8.6 & 7.7 & 3.4 & 125/605 & 49/322 \\
HY-World 2.0 & 98.9 & 2.1 & 1.3 & 6.4 & 3.1 & 2/3 & 23/57 \\
Lyra 2.0 & 99.4 & 2.6 & 1.6 & 4.7 & 2.0 & 0/11 & 50/78 \\
SolarWM & 99.0 & 7.8 & 4.4 & 5.2 & 3.0 & 2/24 & 147/390 \\
WorldCrafter & 99.6 & 8.2 & 4.7 & 9.2 & 5.2 & 0/1 & 315/638 \\
LingBot-World v2 & 99.2 & 10.0 & 5.9 & 8.0 & 4.5 & 7/15 & 201/609 \\
HY-WorldPlay & 99.7 & 17.4 & 9.5 & 22.7 & 5.2 & 25/192 & 46/723 \\
EVOKE & 99.0 & 16.0 & 11.0 & 10.6 & 7.1 & 19/239 & 93/726 \\
Matrix-Game 3.5 & 85.8 & 13.2 & 8.3 & 19.8 & 8.0 & 119/528 & 45/311 \\
AlayaWorld$^\dagger$ & 74.3 & 29.3 & 13.3 & 29.6 & 9.5 & 16/532 & 14/518 \\
H3-World & 94.0 & 26.6 & 11.9 & 21.2 & 8.6 & 8/404 & 5/327 \\
\bottomrule
\end{tabular*}
\endgroup
\end{table}

\FloatBarrier
\section{Validation on Known Motion and Rigid Renderings}\label{app:E}
\subsection{Pose recovery on known motion}
We first recover the cameras of $30$ segments from ten UE5 sequences with exact camera trajectories recorded by the engine, placing anchors on the first frame with the sampler of the benchmark (App.~\ref{app:C1}), which gives a median of $403$ anchors per segment. All $30$ segments are solved, with a median 2D tracking error of $0.117\%$ of the image diagonal (90th percentile, P90, $0.176\%$). After the ray-space refinement, the recovered cameras deviate from the true ones by a median of $1.48\%$ in translation and $0.31\%$ in rotation (P90 $3.98\%$ and $0.68\%$), against $2.28\%$ and $0.46\%$ after EPnP with RANSAC alone.

We then apply synthetic degradations to all frames after the first, and as shown in Tab.~\ref{tab:E.2}, all $390$ degraded segments are solved and pass the pose checks. Blur, compression, and flicker change the median pose errors by at most $0.12$ percentage points, and moderate wobble of $0.5\%$ of the diagonal, alone or combined with the other degradations, keeps the median translation and rotation errors within $3.98\%$ and $2.09\%$. Only the strongest wobble of $1\%$ raises them to $8.26\%$ and $4.40\%$.

\begin{table}[H]
\centering
\caption{\textbf{Pose recovery on the $30$ ground-truth segments under synthetic degradations (median / P90, in percent).} Tracking relative to the image diagonal, translation and rotation relative to the ground-truth motion. Blur: Gaussian $\sigma$ in pixels. H.264: x264 CRF. Flicker: brightness factor drawn uniformly from $[1-a,1+a]$. Wobble: non-rigid 2D displacement with correlation length of a quarter of the image width and RMS amplitude in \% of the image diagonal. Combined: $0.5\%$ wobble, $\sigma=2$ blur, $a=0.1$ flicker, and CRF $36$.}
\label{tab:E.2}
\begingroup\footnotesize\setlength{\tabcolsep}{4pt}\renewcommand{\arraystretch}{1.08}
\begin{tabular*}{0.74\linewidth}{@{\extracolsep{\fill}}lrrr@{}}
\toprule
Condition & Tracking & Translation & Rotation \\
\midrule
Unperturbed & 0.117 / 0.176 & 1.480 / 3.983 & 0.306 / 0.675 \\
\midrule
Blur $\sigma=1$\,px & 0.113 / 0.181 & 1.580 / 3.188 & 0.335 / 0.647 \\
Blur $\sigma=4$\,px & 0.117 / 0.179 & 1.500 / 3.796 & 0.356 / 0.653 \\
H.264 CRF 30 & 0.119 / 0.187 & 1.461 / 3.204 & 0.390 / 0.692 \\
H.264 CRF 42 & 0.139 / 0.240 & 1.597 / 5.637 & 0.413 / 0.703 \\
Flicker $a=0.05$ & 0.116 / 0.176 & 1.475 / 3.971 & 0.305 / 0.676 \\
Flicker $a=0.2$ & 0.117 / 0.178 & 1.561 / 3.466 & 0.307 / 0.681 \\
Wobble $0.25\%$ & 0.255 / 0.373 & 2.251 / 4.005 & 0.974 / 4.468 \\
Wobble $0.5\%$ & 0.476 / 0.624 & 3.184 / 7.096 & 2.094 / 8.994 \\
Wobble $1\%$ & 0.927 / 1.225 & 8.256 / 16.509 & 4.397 / 18.499 \\
Combined & 0.482 / 0.647 & 3.977 / 7.254 & 2.046 / 9.147 \\
\bottomrule
\end{tabular*}
\endgroup
\end{table}

\FloatBarrier
\subsection{Anchor layout and full-pipeline scoring}\label{app:H1}
On the same segments, keeping $25\%$ of the anchors spread uniformly over the image gives $2.17\times$ and $2.93\times$ lower median and P90 translation errors than keeping them in a single image region, which motivates spreading the anchors over a grid of image cells. We also pass the true and the recovered cameras of five groups of rendered UE5 control videos, each with six eight-step single actions and two composed routes, through stride calibration and scoring. \mHold\ differs by $1.2$ points ($98.8$ vs.\ $100.0$), \mPair\ by $0.7$ points in position and $1.8$ points in heading, and single-route scores by at most $3.3$ points, so camera recovery leaves the scores nearly unchanged.

\FloatBarrier
\subsection{Rigid floor and rigid counterparts}\label{app:ceiling}
We render $417$ rigid videos of three procedural layouts along the six held actions and eight routes of the benchmark at $832\times480$, $24$\,fps, and H.264 (CRF $23$), with translation strides from $1\%$ to $16\%$ of the start view's median depth and turns of $1^\circ$ to $6^\circ$ per step. Scored by the unchanged pipeline, their median Rigidity offset over frames with $2^\circ$--$8^\circ$ of image motion is $0.13^\circ$ ($12{,}660$ frames from $389$ videos) and stays between $0.10^\circ$ and $0.15^\circ$ across all six motion settings. With per-frame medians in place of per-frame means, the floor is $0.08^\circ$, which Lyra 2.0 matches, while SolarWM lies $1.6\times$ above Lyra and the other thirteen models lie $2.4\times$ to $3.8\times$ above it, as shown in Tab.~\ref{tab:f1speed}.

Since resolution, frame rate, step duration, compression, and camera path all affect tracking noise, we also build the rigid counterpart of Sec.~\ref{sec:res_f1} for each of the $20$ route videos of every model that have the most frames in $2^\circ$--$8^\circ$ of image motion. The start view and its ground-truth depth are triangulated into a mesh split at depth discontinuities and rendered along the camera path recovered from the generated video, at the model's resolution, frame rate, length, codec, and bitrate. As shown in Tab.~\ref{tab:twins}, every model exceeds its rigid counterpart on the same frames, with $95\%$ bootstrap intervals of the ratio entirely above $1.0$. The ratio is $2.1\times$ to $7.3\times$ for twelve models and $1.6\times$ for Lyra 2.0, SolarWM, and H3-World. The offsets of the counterparts themselves range from $0.054^\circ$ to $0.167^\circ$, because the selected videos differ in camera path, image motion, and content even between models of the same format, so the ratio and the absolute offset of Tab.~\ref{tab:main} order the ranked models differently (Spearman's $\rho=0.67$). HY-World 1.5, for example, has an absolute offset in the middle of the models but the largest ratio, as its counterparts have the smallest offset, and Tab.~\ref{tab:main} therefore reports both.

Since the held-out anchors lie at the ground-truth depth of the start view, a rigid scene generated with a wrong depth would also exceed its counterpart. On the same $20$ videos of every model and on their counterparts, we therefore fit the depth of each held-out anchor freely along its viewing ray in the start view, over $121$ log-spaced values from $0.2$ to $5$ times its ground-truth depth, to minimize the mean offset of the anchor over all later frames, and score each video with these depths on its frames with $2^\circ$--$8^\circ$ of image motion. Freeing the depth lowers the offsets of the generated videos to a median of $45\%$ to $77\%$ of their values at ground-truth depth across models and those of the counterparts to $42\%$ to $74\%$, so both shrink by similar fractions. For fourteen models the generated videos still exceed their counterparts by $1.5\times$ (SolarWM) to $9.5\times$ (HY-World 1.5), with $95\%$ intervals entirely above $1.0$ (Free depth in Tab.~\ref{tab:twins}), so their excess offset is not explained by a rigid scene of wrong depth and reflects deformation as the camera moves. We also fit the camera of every frame and a focal-length scale of every video between $0.7$ and $1.4$ jointly with these depths, to the pose and held-out anchors together, on generated videos and counterparts alike. The ratio then shrinks but stays above one for the same fourteen models, from $1.2\times$ (SolarWM) to $4.3\times$ (HY-World 1.5), again with intervals entirely above $1.0$. For Lyra 2.0 the ratio falls to $1.2\times$ under free depth and to $0.9\times$ under the joint fit, with intervals that include $1.0$, so most of its small excess comes from a depth error that stays fixed as the camera moves, as expected when every view is conditioned on a 3D cache built from estimated depth.

We further inject smooth low-frequency 3D displacement fields, six spatial Fourier modes growing linearly over time, into the counterpart meshes of Lyra 2.0, Matrix-Game 3.0, and SolarWM at amplitudes that move the held-out anchors by $0.25^\circ$, $0.5^\circ$, $1^\circ$, and $2^\circ$. In all three formats the Rigidity offset grows linearly with the injected deformation, with slopes of $0.59$, $0.69$, and $0.62$. The slopes stay below one because the camera fitted to the pose anchors absorbs part of any smooth surface deformation. Refitting a per-video focal-length scale on the pose anchors for eight models gives scales between $0.72$ and $1.16$ and changes the Rigidity offset by only $-5\%$ to $+2\%$ (Focal in Tab.~\ref{tab:twins}), so the offsets do not come from a mismatched focal length.

\begin{table*}[!htbp]
\centering
\caption{\textbf{Rigid counterparts at each model's native format, on matched frames with $2^\circ$--$8^\circ$ of image motion.} Frames: paired frames. Rigid, Gen.: median held-out anchor offset ($^\circ$) of rigid counterparts and generated videos under per-frame means, with their ratio and its $95\%$ video-bootstrap interval. Median: the ratio under per-frame medians. Focal: relative offset after refitting the focal length. Free depth: ratio of the median per-video offsets after fitting the depth of each held-out anchor along its start-view ray, with its $95\%$ interval over matched video pairs. Joint fit: the same ratio when the camera of every frame and a focal-length scale of every video are fitted together with these depths. --: not evaluated. $^\dagger$Stride not reliably measurable.}
\label{tab:twins}
\begingroup\footnotesize\setlength{\tabcolsep}{4pt}
\renewcommand{\arraystretch}{1.08}\begin{tabular*}{\textwidth}{@{\extracolsep{\fill}}lccccccccccc@{}}
\toprule
 & & \multicolumn{4}{c}{Per-frame mean} & Median & & \multicolumn{2}{c}{Free depth} & \multicolumn{2}{c}{Joint fit} \\
\cmidrule(lr){3-6}\cmidrule(lr){9-10}\cmidrule(lr){11-12}
Model & Frames & Rigid & Gen. & Ratio & $95\%$ int. & Ratio & Focal & Ratio & $95\%$ int. & Ratio & $95\%$ int. \\
\midrule
Matrix-Game 3.0 & 961 & 0.113 & 0.437 & 3.9 & [3.15, 4.72] & 3.9 & -- & 4.2 & [2.80, 5.98] & 2.6 & [1.73, 3.44] \\
Matrix-Game 2.0 & 977 & 0.113 & 0.374 & 3.3 & [2.65, 4.11] & 4.4 & -- & 4.6 & [2.91, 5.50] & 3.1 & [2.34, 3.58] \\
SANA-WM & 1062 & 0.097 & 0.208 & 2.1 & [1.85, 2.73] & 2.8 & -- & 2.4 & [1.69, 2.91] & 1.6 & [1.43, 1.96] \\
DreamX-World & 832 & 0.112 & 0.382 & 3.4 & [2.87, 4.18] & 4.0 & -- & 3.9 & [2.82, 5.30] & 2.5 & [1.97, 3.18] \\
HY-World 1.5 & 1017 & 0.054 & 0.391 & 7.3 & [5.59, 9.04] & 9.2 & -- & 9.5 & [6.07, 11.77] & 4.3 & [3.32, 5.02] \\
HY-World 2.0 & 527 & 0.088 & 0.352 & 4.0 & [2.86, 5.55] & 5.0 & -- & 4.1 & [2.27, 8.80] & 2.5 & [1.54, 3.70] \\
Lyra 2.0 & 2601 & 0.054 & 0.084 & 1.6 & [1.27, 1.89] & 1.5 & 1.00 & 1.2 & [0.98, 1.47] & 0.9 & [0.80, 1.01] \\
SolarWM & 2092 & 0.095 & 0.152 & 1.6 & [1.39, 1.91] & 1.7 & -- & 1.5 & [1.28, 1.94] & 1.2 & [1.10, 1.54] \\
WorldCrafter & 4174 & 0.136 & 0.318 & 2.3 & [1.94, 2.73] & 2.6 & 1.01 & 2.4 & [1.70, 3.30] & 1.7 & [1.28, 2.19] \\
LingBot-World v2 & 1024 & 0.133 & 0.325 & 2.4 & [1.97, 3.37] & 3.1 & 0.96 & 2.7 & [2.16, 3.96] & 2.2 & [1.53, 2.84] \\
HY-WorldPlay & 1022 & 0.124 & 0.485 & 3.9 & [2.73, 5.09] & 4.7 & 1.02 & 3.4 & [2.43, 6.49] & 2.3 & [1.90, 4.38] \\
EVOKE & 1878 & 0.155 & 0.461 & 3.0 & [2.25, 3.32] & 3.1 & 0.95 & 3.2 & [2.12, 4.11] & 1.9 & [1.56, 2.34] \\
Matrix-Game 3.5 & 1411 & 0.114 & 0.321 & 2.8 & [2.03, 3.77] & 2.8 & 0.98 & 1.8 & [1.44, 3.23] & 1.4 & [1.20, 2.13] \\
AlayaWorld$^\dagger$ & 1521 & 0.071 & 0.213 & 3.0 & [1.41, 4.75] & 3.1 & 0.96 & 2.9 & [1.21, 7.95] & 2.6 & [1.15, 3.39] \\
H3-World & 3448 & 0.167 & 0.266 & 1.6 & [1.05, 2.14] & 2.3 & 1.01 & 2.0 & [1.76, 2.92] & 1.5 & [1.21, 1.76] \\
\bottomrule
\end{tabular*}
\endgroup
\end{table*}

\FloatBarrier
\subsection{Controlled changes with known effects}\label{app:perturb}
We next change one property of an ideal rendering by a known amount and score the videos with the unchanged pipeline, together with direction accuracy, the trajectory errors of Tab.~\ref{tab:existing}, and the reprojection error of a feed-forward reconstruction. The renderings use the rigid counterparts of Lyra 2.0, Matrix-Game 3.0, and HY-World 1.5, which differ in resolution, frame rate, and step duration, and are encoded with each model's own encoder. Action effects are changed on the seed-$0$ strides of $28$ start views, and the rendered scene is changed on the $60$ counterpart records above, and Fig.~\ref{fig:perturb_summary} summarizes both experiments.

The camera of an ideal model advances by its calibrated stride at every step (M0), and four changes alter one property each. The stride grows linearly to $1.25\times$, $1.5\times$, or $2\times$ by step $16$ (M1), the first step after each switch adds $0.25$, $0.5$, or $1$ times the previous action (M2), the camera stops for $1$, $2$, or $4$ steps in the middle of a route (M3), or the first step after each switch shrinks to $0.5$ or $0.25$ of a stride (M4). As shown in Fig.~\ref{fig:perturb_m} and Tab.~\ref{tab:perturb_m}, the stride-based metrics follow every change while the existing control metrics barely move. A pace that doubles by step $16$ raises the pace ratio to $1.44$, close to the $1.43$ expected once the faster first four steps enter the stride, and lowers the \mHold\ from $87.8$ to $56.8$. Carrying half of the previous action over each switch lowers Near from $95.5\%$ to $65.2\%$, carrying all of it lowers Near to $11.6\%$, and starting each new action with a quarter of a stride lowers it to $40.2\%$. Over these changes, the direction accuracy of routes stays at $92\%$ or above, and the trajectory error after fitting one scale stays below the $10\%$ threshold under which a route counts as accurate, which only stalls of two or more steps exceed.

In Fig.~\ref{fig:perturb_summary}a, the first column flags a held video whose pace ratio lies outside $0.8$--$1.25$ (Pace) and a route that is not completed within one stride (Near), the stall check flags a route with a step that moves less than $10\%$ of its stride, the direction check flags a video with a step that does not move in the commanded direction, and the trajectory errors flag a video whose position error exceeds $10\%$ of the path after fitting one scale or a similarity transform. Held rows count the $125$ to $154$ videos measurable at step $16$, and route rows count $112$ routes. Panel~(b) covers the $60$ counterpart records, with blur and compression at $\sigma=2$ and CRF $36$ (mild) or $\sigma=4$ and CRF $42$ (strong), and its whiskers are $95\%$ start-view bootstrap intervals. Counted route by route, the Near decision flags $29\%$ to $89\%$ of the routes with a single carried-over step, slow start, or stalled step, whereas a fitted scale flags at most $19\%$ of them and the direction check flags $8\%$ to $49\%$. The stall check flags $79\%$ of the routes with a single stalled step, and the Pace test flags $99\%$ of the held videos whose pace doubles, which neither direction checks nor fitted errors flag in more than $3\%$ of the cases. With the threshold of each test instead set so that it flags $5\%$ of the ideal held videos, the Pace test flags $94\%$ of the $140$ measurable held videos whose pace grows to $1.5\times$, whereas the trajectory error after fitting one scale flags $43\%$.

\begin{figure*}[!htbp]
\centering
\includegraphics[width=\textwidth]{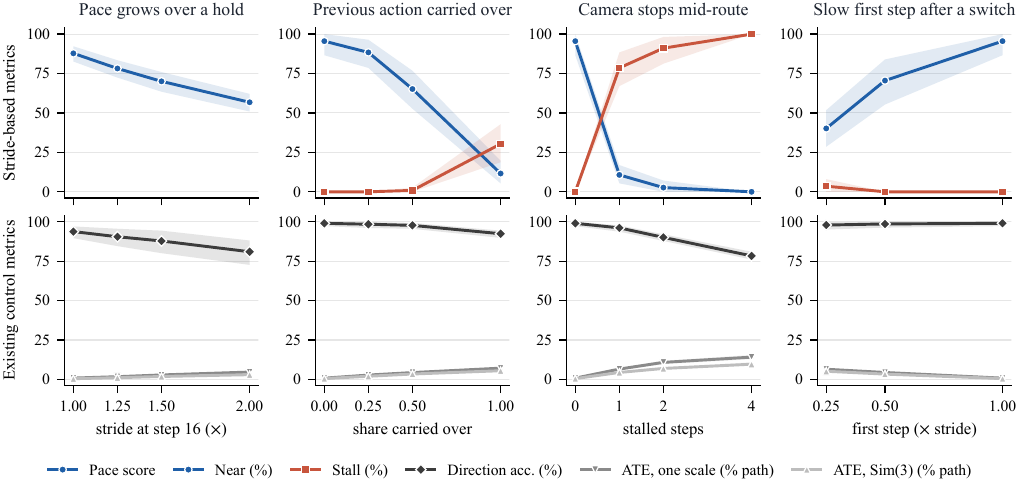}
\caption{\textbf{Response to known changes of action effects, pooled over three formats.} Top: stride-based metrics. Bottom: direction accuracy and median trajectory errors after fitting one scale or a similarity transform, in \% of the commanded path. Panels from left to right: M1 ($168$ held videos per level), M2, M3, and M4 ($112$ routes per level). Bands: $95\%$ start-view bootstrap intervals over $2{,}000$ resamples.}
\label{fig:perturb_m}
\end{figure*}

\begin{table*}[!htbp]
\centering
\caption{\textbf{Routes under known changes of action effects (M0 to M4), pooled over three formats ($112$ routes per level).} Metrics as in Tab.~\ref{tab:main}. Dir.: share of steps in the commanded direction (\%). ATE: as in Fig.~\ref{fig:perturb_m}. --: no completed route.}
\label{tab:perturb_m}
\begingroup\footnotesize\setlength{\tabcolsep}{4pt}\renewcommand{\arraystretch}{1.08}
\begin{tabular*}{\textwidth}{@{\extracolsep{\fill}}lccccccc@{}}
\toprule
 & \multicolumn{4}{c}{Stride-based metrics} & \multicolumn{3}{c}{Existing control metrics} \\
\cmidrule(lr){2-5}\cmidrule(lr){6-8}
Change & Near\,(\%) & Stall\,(\%) & End err.\,(str.) & Swap gap\,(str.) & Dir.\,(\%) & ATE scale\,(\%) & ATE Sim(3)\,(\%) \\
\midrule
Ideal (M0) & 95.5 & 0.0 & 0.12 & 0.04 & 99.0 & 0.8 & 0.5 \\
\midrule
M1, stride $1.25\times$ by step $16$ & 95.5 & 0.0 & 0.12 & 0.04 & 99.0 & 0.9 & 0.6 \\
M1, stride $1.5\times$ & 95.5 & 0.0 & 0.13 & 0.04 & 98.9 & 1.0 & 0.7 \\
M1, stride $2\times$ & 97.3 & 0.0 & 0.19 & 0.04 & 98.7 & 1.4 & 1.0 \\
\midrule
M2, carry-over $0.25$ & 88.4 & 0.0 & 0.39 & 0.33 & 98.4 & 2.8 & 2.0 \\
M2, carry-over $0.5$ & 65.2 & 0.9 & 0.75 & 0.33 & 97.7 & 4.4 & 3.4 \\
M2, carry-over $1$ & 11.6 & 30.4 & 1.64 & 0.81 & 92.4 & 7.1 & 5.5 \\
\midrule
M3, stall of $1$ step & 10.7 & 78.6 & 0.96 & 1.40 & 96.1 & 6.6 & 4.4 \\
M3, stall of $2$ steps & 2.7 & 91.1 & 1.55 & 1.55 & 90.1 & 10.8 & 7.0 \\
M3, stall of $4$ steps & 0.0 & 100.0 & -- & -- & 78.3 & 14.1 & 9.7 \\
\midrule
M4, first step $0.5\times$ & 70.5 & 0.0 & 0.73 & 0.05 & 98.6 & 4.4 & 3.4 \\
M4, first step $0.25\times$ & 40.2 & 3.6 & 1.06 & 0.05 & 97.9 & 6.5 & 5.2 \\
\bottomrule
\end{tabular*}
\endgroup
\end{table*}

Geometric changes of the rendered scene deform each counterpart mesh with the smooth field above or with eight local bumps, both growing linearly over the video until the held-out anchors move by $0.25^\circ$ to $2^\circ$ at the final frame. Appearance changes keep the geometry rigid, and among them texture drift blends the surface inside random rectangles covering $10\%$, $25\%$, or $50\%$ of the image into a shifted copy of the start view, flicker and blur with compression follow Tab.~\ref{tab:E.2}, and occluders lasting two to four frames cover up to half of the frames. The recovered camera is compared with the known render path in strides without any fitted scale. As shown in Tab.~\ref{tab:perturb_ga}, the Rigidity offset grows with both kinds of deformation, from $0.062^\circ$ on the rigid counterparts to $0.325^\circ$ at $1^\circ$ of smooth deformation, with a correlation of $0.84$ between the change of a video and the amplitude. Here the amplitude is the displacement at the final frame, whereas the slopes above relate the offset of each frame to the displacement in that frame, so the two scales are not directly comparable. Texture drift, flicker, and occluders keep it within $0.02^\circ$ of the rigid counterparts, and only the strongest blur with compression raises it by $0.07^\circ$. The recovered camera stays within $0.11$ strides of the true path under appearance changes except strong blur. Across the fourteen ranked models, the offset shows no relation to the end error of completed routes (Spearman $\rho=-0.05$), so deformation does not account for which models end their routes far from the prediction. Within a single video, however, the recovered camera can absorb part of a deformation, as $1^\circ$ of smooth deformation moves it from the true path by a median of $0.64$ strides, so we also relate every route to its own offset. Among the seed-$0$ routes with a measurable offset ($11{,}794$ routes, $10{,}027$ of them completed), routes with a larger offset end slightly farther from their prediction within each model, with a median Spearman's $\rho$ of $0.14$ across the fourteen models. Of the completed routes in the third of each model with the largest offset, $23\%$ are Near, against $30\%$ in the third with the smallest offset, whereas stalls occur in $14\%$ and $17\%$ of these routes and show no such trend. Restricting every model to the third of its routes with the smallest offset leaves the order of the models unchanged, with Spearman's $\rho$ of $0.99$ for Near and $0.98$ for Stall against all routes.

To measure how deformation changes route decisions, we compute Near and Stall on the $60$ counterpart routes once from the true render camera and once from the recovered camera, both with the strides and the prediction of the source route. On the rigid counterparts, the two decisions differ for $2$ of the $60$ routes. The fifteen models exceed their rigid counterparts by a median of $0.21^\circ$, which lies between the excess produced by $0.5^\circ$ and by $1^\circ$ of injected deformation ($0.09^\circ$ to $0.26^\circ$). At these two levels, the Near decision changes for $5$ to $12$ of the $60$ routes, and in $4$ to $8$ of these cases a route that ends within one stride is judged to end beyond it. From $1^\circ$ on, the recovered camera also shows stalls that the true camera does not, in $8$ to $12$ of the $60$ routes at $1^\circ$, although in the generated videos the routes with the largest offset do not stall more often than those with the smallest. The stride-based decisions thus support comparisons between models, while the decision for a single route of a strongly deforming model carries an uncertainty of this size.

We also score every video with the reprojection error of VGGT~\citep{vggt}, which lifts a grid of $256$ first-frame points with its own depth, projects them into later frames with its own cameras, and compares them with its own tracks. As shown in Fig.~\ref{fig:perturb_summary}b, this error rises at least as much with appearance changes as with deformation. Texture drift over half of the image raises it by $0.40$, more than $0.5^\circ$ of smooth deformation ($0.24$), and the strongest blur with compression raises it by $2.14$, more than $2^\circ$ of smooth deformation ($1.52$). On the generated videos of the fifteen models, it separates the videos from their rigid counterparts for only three, namely Matrix-Game 3.0, HY-World 1.5, and HY-WorldPlay, whereas the Rigidity offset separates every model, as shown in Tab.~\ref{tab:twins}.

\begin{table}[!htbp]
\centering
\caption{\textbf{Known changes of the rendered world on the $60$ counterpart records, pooled over three formats (medians over records).} Offset: Rigidity offset ($^\circ$). Camera: error of the recovered camera against the true path (strides, median and P90 over frames). Anchors: surviving held-out anchors (\%). VGGT: reprojection error (\% of the image diagonal). $^*$$59$ records, as one deformed mesh crosses the camera plane and cannot be rendered. $^\ddagger$VGGT defined on $52$ records.}
\label{tab:perturb_ga}
\begingroup\footnotesize\setlength{\tabcolsep}{3.5pt}\renewcommand{\arraystretch}{1.06}
\begin{tabular*}{\linewidth}{@{\extracolsep{\fill}}lccccc@{}}
\toprule
 & & \multicolumn{2}{c}{Camera} & & \\
\cmidrule(lr){3-4}
Change & Offset & Median & P90 & Anchors & VGGT \\
\midrule
Rigid counterpart & 0.062 & 0.08 & 0.14 & 85.6 & 0.34 \\
\midrule
\multicolumn{6}{@{}l}{\textit{Smooth deformation}}\\[-1pt]
$0.25^\circ$ & 0.109 & 0.22 & 0.36 & 85.8 & 0.41 \\
$0.5^\circ$ & 0.179 & 0.36 & 0.72 & 85.0 & 0.64 \\
$1^\circ$ & 0.325 & 0.64 & 1.49 & 83.6 & 1.44 \\
$2^\circ$$^*$ & 0.621 & 1.61 & 3.99 & 79.9 & 1.79 \\
\multicolumn{6}{@{}l}{\textit{Local bumps}}\\[-1pt]
$0.25^\circ$ & 0.096 & 0.20 & 0.38 & 86.0 & 0.39 \\
$0.5^\circ$ & 0.150 & 0.35 & 0.67 & 85.9 & 0.53 \\
$1^\circ$ & 0.252 & 0.58 & 1.38 & 85.1 & 0.83 \\
$2^\circ$ & 0.449 & 0.97 & 1.91 & 81.9 & 1.44 \\
\midrule
\multicolumn{6}{@{}l}{\textit{Texture drift (share of the image)}}\\[-1pt]
$10\%$ & 0.069 & 0.08 & 0.16 & 85.5 & 0.36 \\
$25\%$ & 0.071 & 0.09 & 0.18 & 85.6 & 0.52 \\
$50\%$ & 0.083 & 0.11 & 0.22 & 85.8 & 1.03 \\
\multicolumn{6}{@{}l}{\textit{Flicker}}\\[-1pt]
$a=0.05$ & 0.064 & 0.08 & 0.16 & 85.7 & 0.35 \\
$a=0.1$ & 0.064 & 0.09 & 0.18 & 85.8 & 0.37 \\
$a=0.2$ & 0.067 & 0.09 & 0.16 & 85.6 & 0.34 \\
\multicolumn{6}{@{}l}{\textit{Occluders (share of the frames)}}\\[-1pt]
$10\%$ & 0.063 & 0.08 & 0.15 & 84.8 & 0.31 \\
$25\%$ & 0.063 & 0.08 & 0.14 & 83.8 & 0.33 \\
$50\%$ & 0.063 & 0.08 & 0.16 & 82.4 & 0.32 \\
\multicolumn{6}{@{}l}{\textit{Blur and compression}}\\[-1pt]
$\sigma=1$, CRF $30$ & 0.069 & 0.09 & 0.19 & 85.8 & 0.64 \\
$\sigma=2$, CRF $36$ & 0.086 & 0.14 & 0.28 & 86.4 & 1.43 \\
$\sigma=4$, CRF $42$ & 0.149 & 0.25 & 0.54 & 86.5 & 2.51$^\ddagger$ \\
\bottomrule
\end{tabular*}
\endgroup
\end{table}

\FloatBarrier

\FloatBarrier
\section{Sensitivity and Robustness}\label{app:H}

\subsection{Cameras from other anchors}\label{app:splitanchor}
To test whether scene deformation affects the recovered camera, we recover the cameras of all $1{,}960$ seed-$0$ videos of every model again from either disjoint half of the pose anchors (P1, P2) and from the held-out anchors (Q) in place of the pose anchors (P), with the rest of the pipeline unchanged. As shown in Tab.~\ref{tab:splitanchor}, no model's pace ratio enters or leaves the steady band, Near changes by at most $3.3$ percentage points and Stall by at most $2.2$, and the Spearman rank correlation of the fifteen models between any two anchor sets is at least $0.97$ for every metric.

\begin{table*}[!htbp]
\centering
\caption{\textbf{Cameras recovered from other anchors ($1{,}960$ seed-$0$ videos per model).} P: the pose anchors (as in the paper). P1, P2: either half of them. Q: the held-out anchors in place of the pose anchors. Pace ratio: median over the seed-$0$ single actions (Tab.~\ref{tab:main} pools three seeds). Near, Stall: share of routes (\%). End err.: median end error of completed routes (strides). $^\dagger$Stride not reliably measurable.}
\label{tab:splitanchor}
\begingroup\footnotesize\setlength{\tabcolsep}{2.2pt}\renewcommand{\arraystretch}{1.08}
\begin{tabular*}{\textwidth}{@{\extracolsep{\fill}}lcccccccccccccccc@{}}
\toprule
 & \multicolumn{4}{c}{Pace ratio} & \multicolumn{4}{c}{Near (\%)} & \multicolumn{4}{c}{Stall (\%)} & \multicolumn{4}{c}{End err.\,(str.)}\\
\cmidrule(lr){2-5}\cmidrule(lr){6-9}\cmidrule(lr){10-13}\cmidrule(lr){14-17}
Model & P & P1 & P2 & Q & P & P1 & P2 & Q & P & P1 & P2 & Q & P & P1 & P2 & Q\\
\midrule
\multicolumn{17}{@{}l}{\textit{Discrete keys}}\\[-1pt]
Matrix-Game 3.0 & 0.97 & 0.97 & 0.98 & 0.97 & 42.5 & 42.0 & 40.5 & 43.0 & 0.7 & 0.4 & 1.1 & 0.6 & 0.67 & 0.68 & 0.67 & 0.66 \\
Matrix-Game 2.0 & 0.96 & 0.96 & 0.96 & 0.96 & 24.7 & 23.1 & 24.0 & 23.4 & 7.4 & 7.0 & 7.6 & 6.8 & 1.45 & 1.47 & 1.45 & 1.46 \\
SANA-WM & 1.10 & 1.10 & 1.10 & 1.10 & 7.2 & 7.9 & 7.7 & 7.9 & 36.5 & 36.0 & 36.0 & 36.5 & 2.47 & 2.45 & 2.54 & 2.44 \\
DreamX-World & 2.47 & 2.47 & 2.46 & 2.45 & 0.5 & 0.5 & 0.7 & 0.5 & 12.3 & 11.3 & 11.6 & 11.0 & 5.22 & 5.10 & 5.24 & 5.18 \\
HY-World 1.5 & 0.95 & 0.96 & 0.95 & 0.96 & 3.8 & 3.8 & 3.7 & 4.5 & 54.0 & 54.7 & 54.7 & 54.2 & 2.42 & 2.45 & 2.43 & 2.43 \\
\midrule
\multicolumn{17}{@{}l}{\textit{Camera path}}\\[-1pt]
HY-World 2.0 & 1.00 & 1.00 & 1.00 & 1.00 & 84.3 & 82.7 & 83.0 & 86.2 & 0.3 & 0.2 & 0.2 & 0.4 & 0.30 & 0.32 & 0.31 & 0.29 \\
Lyra 2.0 & 1.01 & 1.01 & 1.01 & 1.02 & 75.9 & 72.8 & 72.6 & 74.6 & 1.0 & 0.9 & 0.9 & 1.0 & 0.51 & 0.54 & 0.52 & 0.53 \\
SolarWM & 0.94 & 0.95 & 0.94 & 0.94 & 25.4 & 24.2 & 24.2 & 25.5 & 2.1 & 2.1 & 1.9 & 1.9 & 1.61 & 1.62 & 1.62 & 1.56 \\
WorldCrafter & 1.08 & 1.08 & 1.09 & 1.09 & 16.0 & 16.3 & 14.7 & 15.8 & 0.1 & 0.1 & 0.3 & 0.3 & 2.38 & 2.37 & 2.37 & 2.41 \\
LingBot-World v2 & 0.94 & 0.94 & 0.93 & 0.93 & 11.2 & 11.2 & 11.7 & 11.2 & 1.3 & 1.1 & 0.6 & 1.2 & 2.32 & 2.29 & 2.26 & 2.29 \\
HY-WorldPlay & 1.06 & 1.07 & 1.06 & 1.06 & 3.8 & 3.4 & 3.7 & 3.6 & 17.1 & 17.9 & 18.4 & 16.3 & 3.44 & 3.44 & 3.48 & 3.35 \\
EVOKE & 1.11 & 1.11 & 1.11 & 1.11 & 3.1 & 2.3 & 3.0 & 2.9 & 21.3 & 22.3 & 21.7 & 22.1 & 4.57 & 4.66 & 4.64 & 4.70 \\
Matrix-Game 3.5 & 0.93 & 0.94 & 0.94 & 0.93 & 8.0 & 7.7 & 7.0 & 7.5 & 47.1 & 46.7 & 47.2 & 46.1 & 2.13 & 2.05 & 2.20 & 2.23 \\
AlayaWorld$^\dagger$ & 4.26 & 4.23 & 4.11 & 4.05 & 0.4 & 0.2 & 0.4 & 0.3 & 47.5 & 46.2 & 47.4 & 48.2 & 50.57 & 52.96 & 52.30 & 57.83 \\
\midrule
\multicolumn{17}{@{}l}{\textit{Text prompt}}\\[-1pt]
H3-World & 1.62 & 1.60 & 1.48 & 1.55 & 0.2 & 0.2 & 0.1 & 0.2 & 36.1 & 35.3 & 33.9 & 35.3 & 9.02 & 8.38 & 8.37 & 8.95 \\
\bottomrule
\end{tabular*}

\endgroup
\end{table*}

\subsection{Evaluation parameters}\label{app:H3}\label{app:H4}
Varying the parameters of the evaluation protocol leaves the findings unchanged. Calibrating strides on the first $2$, $6$, or $8$ steps instead of four changes \mComp\ by at most $1.8$ points for every model, while longer windows predict held actions slightly better because they observe more of the $16$-step hold. Setting the stall threshold to $0.05$ or $0.15$ instead of $0.10$ changes the number of completed pairs by at most $10$ of $560$ for the leading models and changes the scores of completed pairs by at most $2.1$ points for thirteen models, preserving the separation between models. Halving or doubling the stride units $b_t$ and $b_r$ reverses the order of at most two of the $105$ model pairs. Predicting routes from idealized axis-aligned strides instead of the full planar strides $\mathbf U_a\in SE(2)$ changes composition scores by at most $5.8$ points in position and $2.8$ points in heading. Removing the stall check adds back exactly the routes counted as stalled in Tab.~\ref{tab:main}, confirming that the check filters only stalls.

\subsection{Opposite actions do not cancel}\label{app:H5}
Loop-closure tests assume that opposite commands cancel, but calibrating strides on each action shows that they generally do not. On the cancel pairs (G1), the calibrated strides predict a nonzero offset between the two routes, and scoring against these predictions gives lower composition scores than assuming exact cancellation for thirteen models, by up to $22.6$ points in position for LingBot-World v2. This offset is consistent across seeds, as the predicted offset between the two routes exceeds twice its standard error across the three seeds for $56\%$ to $94\%$ of the cancel pairs in fourteen models, at least three times the rate expected for a zero offset, and only H3-World stays near chance. On the swap pairs (G2--G4), where both routes execute the same actions in different orders, the two predictions differ by at most $0.2$ points.

\subsection{Model-wide stride references}\label{app:reference}
Models share no common unit of motion, and their median translation strides span a factor of $80$ and their forward strides alone a factor of $139$, as shown in Fig.~\ref{fig:bench}b. Three camera-pose interfaces also tie the length of a commanded translation to the scene. HY-World 2.0 sets the median depth of the start view to one, Lyra 2.0 scales trajectories by its monocular depth estimate, and LingBot-World v2 normalizes each trajectory by its largest step. Calibrating a stride on each video's first steps gives every model its own unit. Replacing it with a single stride per model and action, taken as the median across all videos, lowers \mHold\ for fourteen models because each video is then penalized for variation across views and scenes as well, but the model order stays largely the same (Spearman $\rho=0.95$), with Lyra 2.0 first and AlayaWorld and H3-World last under both references.

\subsection{Uncertainty of the main results}\label{app:uncertainty}
Tab.~\ref{tab:bootstrap} gives $95\%$ intervals over $2{,}000$ resamples of the $30$ scenes, with the same resampled scenes shared across all models. HY-World 2.0, Lyra 2.0, and SolarWM take the first three places in \mComp\ in at least $95\%$ of the resamples. The intervals for \mResp\ and completed routes are widest for the models with many unresponsive or stalled videos. Across the fifteen models, \mComp\ follows the share of completed routes closely (Spearman $\rho=0.96$) and \mHold\ less closely ($\rho=0.82$), so held actions and composed routes rank models differently. The supplementary video-level rigidity score \mScene\ shows no clear relation to the control scores, but decreases with stride length ($\rho=-0.88$) because a camera that moves little keeps its anchors near their initial rays, so the main text evaluates Rigidity at matched image motion (Sec.~\ref{sec:rigidity}).

\begin{table}[!htbp]
\centering
\caption{\textbf{Scene-bootstrap $95\%$ intervals of the main results.} Intervals over $2{,}000$ resamples of the $30$ scenes, with the range of each model's rank in \mHold\ and \mComp\ from the 2.5th to the 97.5th percentile. Repeat and Views: stride variation of App.~\ref{app:F}.}
\label{tab:bootstrap}
\begingroup\scriptsize\setlength{\tabcolsep}{2pt}\renewcommand{\arraystretch}{1.08}
\begin{tabular*}{\linewidth}{@{\extracolsep{\fill}}lrrrrrrcc@{}}
\toprule
Model & \mResp\ (\%) & \mHold & Compl. (\%) & \mComp & Repeat (\%) & Views (\%) & \mHold\ rank & \mComp\ rank\\
\midrule
Matrix-Game 3.0 & [99.1, 100.0] & [73.0, 80.7] & [87.8, 98.6] & [54.7, 62.7] & [4.1, 5.7] & [19.6, 32.0] & 6--10 & 4--6 \\
Matrix-Game 2.0 & [83.2, 85.8] & [62.0, 67.4] & [87.8, 93.1] & [46.3, 53.0] & [13.2, 18.1] & [36.7, 47.1] & 11--12 & 6--7 \\
SANA-WM & [98.8, 99.9] & [73.8, 78.3] & [57.4, 67.4] & [25.2, 31.5] & [12.2, 16.9] & [32.0, 41.9] & 6--9 & 10--11 \\
DreamX-World & [98.5, 99.5] & [44.4, 48.3] & [84.4, 91.0] & [22.3, 26.2] & [22.3, 25.4] & [31.7, 46.4] & 13 & 11--13 \\
HY-World 1.5 & [99.1, 100.0] & [74.7, 80.8] & [38.4, 54.0] & [17.1, 27.0] & [40.6, 51.9] & [41.2, 54.4] & 5--8 & 11--13 \\
HY-World 2.0 & [98.7, 99.7] & [83.1, 89.7] & [98.1, 99.8] & [80.4, 85.4] & [1.7, 3.5] & [31.0, 46.3] & 1--3 & 1 \\
Lyra 2.0 & [98.7, 100.0] & [86.4, 89.5] & [97.8, 99.9] & [75.8, 80.1] & [2.6, 4.9] & [23.0, 36.6] & 1--2 & 2 \\
SolarWM & [97.0, 99.8] & [83.0, 87.2] & [96.1, 98.9] & [63.8, 67.9] & [15.2, 24.3] & [24.2, 37.6] & 2--3 & 3 \\
WorldCrafter & [98.9, 99.9] & [76.4, 82.9] & [95.8, 99.4] & [54.2, 59.1] & [11.5, 14.1] & [28.2, 40.8] & 4--6 & 4--6 \\
LingBot-World v2 & [98.5, 99.8] & [80.4, 84.6] & [89.5, 96.2] & [51.4, 57.5] & [16.5, 22.4] & [28.0, 38.1] & 4--5 & 4--7 \\
HY-WorldPlay & [99.4, 100.0] & [72.6, 77.0] & [78.5, 85.3] & [35.7, 41.1] & [31.3, 38.2] & [38.1, 51.2] & 7--10 & 8 \\
EVOKE & [98.5, 99.7] & [72.0, 75.7] & [74.0, 81.3] & [30.3, 35.7] & [8.4, 12.3] & [31.9, 46.2] & 9--10 & 9 \\
Matrix-Game 3.5 & [80.1, 90.9] & [61.1, 70.1] & [45.5, 59.7] & [18.3, 26.2] & [19.8, 30.7] & [51.0, 76.2] & 11--12 & 11--13 \\
AlayaWorld & [68.3, 82.0] & [16.5, 23.4] & [41.2, 54.7] & [9.3, 12.8] & [46.6, 59.7] & [89.7, 106.9] & 15 & 14 \\
H3-World & [90.3, 93.4] & [22.4, 27.2] & [25.1, 35.2] & [6.1, 9.0] & [72.7, 81.9] & [79.7, 89.0] & 14 & 15 \\
\bottomrule
\end{tabular*}
\endgroup
\end{table}

\subsection{Step duration and control interface}\label{app:duration}
Because each interface defines its own command interval, one step lasts between $0.125$ and $1.0$ second across the fifteen models, as listed in Tab.~\ref{tab:duration}. Across the fourteen ranked models, step duration shows no clear relation to any control metric, with Spearman correlations within $\pm0.19$ for the pace ratio, \mHold, Stall, Near, End err., and Swap gap. All three findings also hold within each interface group, as every model exceeds the Rigidity offset of its rigid counterpart, all ranked camera-pose models and four of the five key-driven models keep a steady pace, and most models in both groups bring fewer than a quarter of their routes within one stride of the prediction.

\begin{table}[!htbp]
\centering
\caption{\textbf{Step duration and control interface.} Rigidity offset in degrees at matched image motion. $^\dagger$Stride not reliably measurable.}
\label{tab:duration}
\begingroup\footnotesize\setlength{\tabcolsep}{3.5pt}
\renewcommand{\arraystretch}{1.08}\begin{tabular*}{\linewidth}{@{\extracolsep{\fill}}llcccccc@{}}
\toprule
Model & Interface & Frames/step & FPS & Seconds/step & Pace ratio & Rigidity offset & Rigid-counterpart ratio \\
\midrule
Matrix-Game 3.0 & Keyboard & 4 & 17 & 0.235 & 0.96 & 0.39 & 3.85 \\
Matrix-Game 2.0 & Keyboard & 4 & 16 & 0.250 & 0.96 & 0.40 & 3.30 \\
SANA-WM & Keyboard & 4 & 16 & 0.250 & 1.09 & 0.27 & 2.15 \\
DreamX-World & Keyboard & 4 & 16 & 0.250 & 2.11 & 0.38 & 3.40 \\
HY-World 1.5 & Keyboard & 4 & 24 & 0.167 & 0.97 & 0.33 & 7.30 \\
HY-World 2.0 & Camera pose & 2 & 16 & 0.125 & 1.00 & 0.35 & 4.01 \\
Lyra 2.0 & Camera pose & 10 & 16 & 0.625 & 1.01 & 0.11 & 1.58 \\
SolarWM & Camera pose & 8 & 24 & 0.333 & 0.94 & 0.17 & 1.60 \\
WorldCrafter & Camera pose & 16 & 16 & 1.000 & 1.08 & 0.33 & 2.34 \\
LingBot-World v2 & Camera pose & 4 & 16 & 0.250 & 0.94 & 0.38 & 2.45 \\
HY-WorldPlay & Camera pose & 4 & 24 & 0.167 & 1.11 & 0.38 & 3.92 \\
EVOKE & Camera pose & 8 & 24 & 0.333 & 1.12 & 0.34 & 2.97 \\
Matrix-Game 3.5 & Camera pose & 8 & 16 & 0.500 & 0.95 & 0.34 & 2.82 \\
AlayaWorld$^\dagger$ & Camera pose & 8 & 24 & 0.333 & 4.17 & 0.26 & 3.01 \\
H3-World & Text & 17 & 24 & 0.708 & 1.29 & 0.32 & 1.59 \\
\bottomrule
\end{tabular*}
\endgroup
\end{table}

\subsection{Routes with matched camera motion}\label{app:matched}
Each model runs with its own command magnitude, so the same route moves the camera by different amounts across models. We therefore describe every seed-$0$ route by the summed translation of its calibrated strides divided by the depth of the start view, and by the summed turn of its calibrated strides, and keep only the routes inside the middle half of the pooled distribution of both quantities, from $0.12$ to $0.33$ of the start-view depth and up to $5.6^\circ$ of turn. Every ranked model keeps $169$ to $630$ of its $1{,}120$ routes in this band. As shown in Tab.~\ref{tab:matched}, the order of the fourteen ranked models on these routes is nearly the same as on all routes, with Spearman's $\rho$ of $0.98$ for Near, $0.95$ for Stall, and $1.00$ for End err. Weighting the nine cells of a three-by-three grid of the two quantities equally instead gives $\rho$ of $0.99$ for Near and $0.97$ for Stall. HY-World 2.0 and Lyra 2.0 still bring most matched routes within one stride ($86.8\%$ and $89.9\%$), and the four models that stall most often still stall on $33\%$ to $53\%$ of them. Matching the amount of motion does not match the duration of a step, which remains a property of each interface (App.~\ref{app:duration}).

\begin{table}[!htbp]
\centering
\caption{\textbf{Route outcomes on all seed-$0$ routes and on routes with matched camera motion.} Matched: routes whose depth-relative translation and summed turn both lie in the middle half of the pooled distribution. Near and Stall in \%, End err.\ in strides, $n$: matched routes of the model.}
\label{tab:matched}
\begingroup\footnotesize\setlength{\tabcolsep}{4pt}\renewcommand{\arraystretch}{1.08}
\begin{tabular*}{\linewidth}{@{\extracolsep{\fill}}lccccccr@{}}
\toprule
 & \multicolumn{2}{c}{Near (\%)} & \multicolumn{2}{c}{Stall (\%)} & \multicolumn{2}{c}{End err.\ (str.)} & \\
\cmidrule(lr){2-3}\cmidrule(lr){4-5}\cmidrule(lr){6-7}
Model & All & Matched & All & Matched & All & Matched & $n$ \\
\midrule
Matrix-Game 3.0 & 42.5 & 40.7 & 0.7 & 0.5 & 0.67 & 0.78 & 393 \\
Matrix-Game 2.0 & 24.7 & 29.5 & 7.4 & 7.2 & 1.45 & 1.40 & 488 \\
SANA-WM & 7.2 & 5.7 & 36.5 & 33.3 & 2.47 & 2.51 & 441 \\
DreamX-World & 0.5 & 0.6 & 12.3 & 10.8 & 5.22 & 4.22 & 351 \\
HY-World 1.5 & 3.8 & 3.2 & 54.0 & 52.5 & 2.42 & 2.44 & 558 \\
HY-World 2.0 & 84.3 & 86.8 & 0.3 & 0.3 & 0.30 & 0.28 & 630 \\
Lyra 2.0 & 75.9 & 89.9 & 1.0 & 0.0 & 0.51 & 0.37 & 287 \\
SolarWM & 25.4 & 30.1 & 2.1 & 0.9 & 1.61 & 1.40 & 458 \\
WorldCrafter & 16.0 & 17.8 & 0.1 & 0.2 & 2.38 & 2.13 & 477 \\
LingBot-World v2 & 11.2 & 9.6 & 1.3 & 1.7 & 2.32 & 2.27 & 363 \\
HY-WorldPlay & 3.8 & 4.0 & 17.1 & 15.0 & 3.44 & 3.57 & 621 \\
EVOKE & 3.1 & 4.2 & 21.3 & 16.5 & 4.57 & 4.10 & 333 \\
Matrix-Game 3.5 & 8.0 & 12.0 & 47.1 & 32.9 & 2.13 & 1.68 & 258 \\
H3-World & 0.2 & 0.0 & 36.1 & 41.4 & 9.02 & 9.57 & 169 \\
\bottomrule
\end{tabular*}
\endgroup
\end{table}

\FloatBarrier
\subsection{Stride variation across seeds and views}\label{app:F}
Repeat and Views measure how much the axial component $u$ of an action's stride varies across the three seeds of a start view and across the start views of a scene, defined as the relative standard deviation
\begin{equation}
\operatorname{RSD}(u)=100\,\frac{\sqrt{\frac{1}{n-1}\sum_{i=1}^{n}(u_i-\bar u)^2}}{\frac{1}{n}\sum_{i=1}^{n}|u_i|},
\label{eq:F.1}
\end{equation}
averaged within each scene and then across scenes, separately for the four translations and the two turns. These two statistics characterize the models and do not enter the main-text scores. As shown in Tab.~\ref{tab:F.1}, HY-World 2.0 and Lyra 2.0 repeat their strides most closely across seeds, and for every model the translation stride varies more across views of a scene than across seeds from the same view.

\begin{table}[!htbp]
\centering
\caption{\textbf{Stride variation (RSD, \%) across repeated generations from one view (Repeat) and across the start views of a scene (Views).}}
\label{tab:F.1}\label{tab:F.2}
\begingroup\footnotesize\setlength{\tabcolsep}{3pt}\renewcommand{\arraystretch}{1.08}
\begin{tabular*}{0.8\linewidth}{@{\extracolsep{\fill}}lrrrr@{}}
\toprule
 & \multicolumn{2}{c}{Repeat} & \multicolumn{2}{c}{Views}\\
\cmidrule(lr){2-3}\cmidrule(lr){4-5}
Model & Transl.\,$\downarrow$ & Turn\,$\downarrow$ & Transl.\,$\downarrow$ & Turn\,$\downarrow$ \\
\midrule
Matrix-Game 3.0 & 4.8 & 14.4 & \textbf{25.3} & 13.6 \\
Matrix-Game 2.0 & 15.6 & 1.8 & 42.0 & 9.0 \\
SANA-WM & 14.4 & 3.1 & 36.8 & 5.3 \\
DreamX-World & 23.8 & 33.0 & 38.6 & 19.3 \\
HY-World 1.5 & 46.2 & 6.6 & 47.6 & 9.0 \\
HY-World 2.0 & \textbf{2.5} & \underline{0.7} & 38.2 & \underline{4.8} \\
Lyra 2.0 & \underline{3.6} & \textbf{0.4} & \underline{29.2} & \textbf{0.9} \\
SolarWM & 19.1 & 2.3 & 30.2 & 6.1 \\
WorldCrafter & 12.8 & 3.4 & 34.1 & 6.1 \\
LingBot-World v2 & 19.2 & 4.0 & 32.7 & 10.7 \\
HY-WorldPlay & 34.6 & 13.7 & 44.2 & 17.0 \\
EVOKE & 10.3 & 2.8 & 38.9 & 5.5 \\
Matrix-Game 3.5 & 25.2 & 5.1 & 63.8 & 11.3 \\
AlayaWorld & 53.0 & 58.7 & 98.1 & 106.9 \\
H3-World & 77.0 & 52.0 & 84.0 & 51.9 \\
\bottomrule
\end{tabular*}
\endgroup
\end{table}

\FloatBarrier
\raggedbottom
\section{Human Study}\label{app:human}
Ten paid research assistants who are not authors rated generated videos through a web interface, blind to the models and the metric scores. All of them gave informed consent before taking part, and each rater worked independently on all three parts of the study. The commanded actions were shown beneath every video with the active step highlighted, playback stopped at the final evaluated step, and an answer could be submitted only after the full clip had played once.

\paragraph{Questions.} For single videos (Part~A), raters answered \emph{``Overall, does the camera move stably according to the commands?''} A rating of \emph{Good} means that every step moves in the commanded direction at a roughly steady pace without sudden acceleration, slowdown, stalling, jumping, or drift off course, and raters were instructed to judge camera motion independently of visual quality or surface deformation. For swap pairs (Part~B), the two route videos were played side by side and their final frames were then shown enlarged next to each other. Raters judged whether the two routes end at the same place by comparing the position and size of fixed objects such as corners, doors, and pillars in the two final frames. They counted them as the same place when these objects nearly coincide, with an offset of at most about $3\%$ of the image width, and as different places when the camera has visibly moved forward, backward, or sideways or turned, or when unsure. For rigidity pairs (Part~C), two videos generated by different models from the same start view along the same route were played side by side, and raters chose the video in which the scene keeps its shape more stably, following a few fixed objects such as doors, window frames, and furniture in both videos and ignoring sharpness, color, lighting, image quality, and whether the camera follows the commands. Parts~B and~C were preceded by worked examples, and both are forced choices. Their instructions were refined after pilot sessions with the first raters, and all ten raters completed both parts under the final instructions.

\paragraph{Items and scores.} Part~A covers $160$ randomly sampled videos from fifteen models, $80$ with a single held action and $80$ with a composed route, plus $10$ repeated videos. The three raters of the earlier rounds rated only the $50$ videos that are new in this round, and their earlier ratings of the remaining $110$ videos are reused. Each single-action video is scored by its \mHold\ and each route video by its route score $s_q$ of App.~\ref{app:D3}, which is zero for incomplete routes. Part~B covers $140$ G2--G4 swap pairs in which both routes are completed, ten from each of fourteen models spread evenly over that model's range of pair scores $s_p$, plus $10$ repeated pairs, and each pair is scored by its pair score. Part~C covers $40$ pairs, each scored by the per-video Rigidity offset with the frame selection of Tab.~\ref{tab:main}. Both videos of a pair have at least eight evaluated frames and a similar amount of camera motion, with their numbers of evaluated frames and their median image motion within a factor of $1.5$. In $32$ strong pairs the larger offset is at least four times the smaller and at least $0.8^\circ$, and in $8$ moderate pairs it is two to three times the smaller and at least $0.5^\circ$. Six strong pairs are repeated with the sides swapped.

\paragraph{Analysis.} The analysis was fixed before any answer was collected. The label of an item is the choice of at least six of the ten raters, and items split five to five are excluded. In Parts~A and~B, item-level agreement is the ROC AUC of the score for the majority label, with $95\%$ bootstrap intervals from $10{,}000$ resamples of start views within models, and model-level agreement is Spearman's $\rho$ between the share of positive ratings and the mean score, resampled over models. In Part~C, agreement is the share of pairs whose majority chose the video with the smaller offset, tested against chance with a binomial test. Individual choices are further modeled by a logistic regression on the log offset ratio, with standard errors clustered by rater, and the difference in MUSIQ quality~\citep{ke2021musiq} between the two videos is added as a covariate to test whether raters followed image quality instead of scene shape.

\paragraph{Results.} As shown in Tab.~\ref{tab:human3} and Fig.~\ref{fig:human3}, raters agree closely with one another and with the scores. In Part~A, Fleiss' $\kappa$ is $0.87$, mean pairwise agreement is $95\%$, and $81$ of $82$ repeated judgments are identical. The per-video score separates the $113$ of $160$ videos judged good with an AUC of $0.88$ [$0.82$, $0.93$], which is $0.87$ on single-action and $0.85$ on route videos. The majority judged $17$ of the $18$ stalled routes as bad, and these routes score zero. Completion alone, used as a binary score, separates the route videos judged good with an AUC of $0.75$ [$0.67$, $0.82$], and the route score raises this by $0.10$ [$0.04$, $0.16$] under the same resampling, so the distance between the end of a route and its prediction carries information that raters perceive beyond completion. On the $62$ completed routes alone, where $46$ are judged good and $16$ bad, the AUC of the route score is $0.71$ [$0.57$, $0.84$], because a single video does not show the stride-predicted endpoint and raters in Part~A instead judge whether the camera moves steadily along the way. Across models, the share of positive judgments follows the mean score with $\rho=0.77$ [$0.35$, $0.96$].

In Part~B, Fleiss' $\kappa$ is $0.83$, pairwise agreement is $92\%$, and $95$ of $100$ repeated judgments are identical. The majority judged $62$ pairs to end at the same place and $77$ at different places, with one tie, and the pair score separates the two groups with an AUC of $0.83$ [$0.76$, $0.89$]. Both routes of every pair are completed and their final frames are shown side by side, so this AUC is measured on completed routes in a setting where raters compare the two ends directly.

In Part~C, the majority chose the video with the smaller Rigidity offset in all $31$ strong pairs that had a majority ($p<10^{-9}$, one pair tied) and in all $8$ moderate pairs ($p=0.008$). Individual raters agreed with the offset in $87.5\%$ to $100\%$ of the pairs (median $95\%$), Fleiss' $\kappa$ is $0.86$, and $56$ of $60$ repeated judgments chose the same video after the sides were swapped. Because the pairs were selected for at least a twofold difference in offset and nearly every pair is answered in line with the offset, Part~C tests whether raters perceive clear differences in offset, and the logistic slope on the offset ratio is close to zero within this sampled range. The chosen video had the higher MUSIQ score in $60\%$ of the choices, and with the quality difference as a covariate the model still predicts that raters choose the video with the smaller offset in $95\%$ of the choices when the two videos are of equal quality (intercept $3.02$ [$1.25$, $4.79$]).

\paragraph{Comparison with other scores.} The majority labels of Part~A also allow a comparison with other control scores. Direction accuracy is the share of steps that move along the commanded axis or turn the commanded way. The trajectory errors are the root mean square position error over the step boundaries divided by the length of the predicted path, as in App.~\ref{app:existing}, either after fitting one scale or a similarity transform, or with the scale fixed to the calibrated strides. The per-step error is the mean distance between the displacement of each step and that of its stride, in strides, again without any fit. All trajectory errors use the stride-predicted path as their reference, which existing benchmarks do not have for key-driven models. As shown in Tab.~\ref{tab:humanmetrics}, over the $149$ videos on which every score is defined, the StrideWorld scores reach an AUC of $0.88$, against $0.53$ to $0.71$ for the other scores, and the difference to each of them has a $95\%$ bootstrap interval above zero. Most of this gap comes from held actions ($0.86$ against $0.40$ to $0.69$), although with $70$ held videos the interval of the difference includes zero for the trajectory errors with a fitted and with a fixed scale. On route videos, all scores that use the stride-predicted path agree with the raters to a similar degree ($0.78$ to $0.84$), and direction accuracy reaches $0.72$. On completed routes, the scores that follow the whole route agree better ($0.77$ to $0.85$, including $0.82$ for the per-step error) than the route score, which measures the end error ($0.71$), as raters judge whether the camera moves steadily along the way. For this reason, StrideWorld reports per-step diagnostics such as stalls, carry-over, and slow starts alongside the \mComp.

The majority labels of Part~B allow the same comparison on swap pairs, where every other score is averaged over the two routes of a pair. Over the $139$ pairs with a majority label, the pair score reaches an AUC of $0.83$, against $0.66$ to $0.71$ for the other scores (last row of Tab.~\ref{tab:humanmetrics}), and the difference to each of them has a $95\%$ bootstrap interval above zero, with $0.15$ [$0.06$, $0.23$] for the trajectory error with a fitted scale and $0.16$ [$0.08$, $0.23$] with a similarity transform. Taking the worse of the two routes instead of their mean changes no value by more than $0.02$. Because raters in Part~B compare the two ends, we also score each pair by how far the offset between the ends of its two routes departs from the offset between their predicted ends, as a share of the path length. With the scale fixed to the calibrated strides, this end offset is the position part of the pair score up to its normalization, and it agrees with the raters equally well ($0.84$). Fitting one scale to each route lowers its AUC to $0.74$, a drop of $0.10$ [$0.04$, $0.15$], and fitting a similarity transform lowers it to $0.68$, so a reference fitted to each video removes part of the difference between the two ends that raters see.

\begin{table}[H]
\centering
\caption{\textbf{Agreement of different scores with the majority labels of Parts~A and~B (ROC AUC).} StrideWorld: \mHold{} for held actions, the route score for routes, and the pair score for swap pairs, where every other score is the mean over the two routes of a pair. Trajectory errors and the per-step error are negated so that higher is better, and all of them use the stride-predicted path as reference, with a fitted or a fixed scale. Videos and pairs on which every score is defined.}
\label{tab:humanmetrics}
\begingroup\footnotesize\setlength{\tabcolsep}{3.5pt}\renewcommand{\arraystretch}{1.08}
\begin{tabular*}{\linewidth}{@{\extracolsep{\fill}}lrcccccc@{}}
\toprule
 & & & & \multicolumn{3}{c}{Trajectory error} & Per-step \\
\cmidrule(lr){5-7}
Videos & $n$ & StrideWorld & Direction & Scale fit & Sim(3) fit & Fixed & error \\
\midrule
All & 149 & \textbf{0.88} & 0.53 & 0.65 & 0.70 & 0.62 & 0.71 \\
Held actions & 70 & \textbf{0.86} & 0.40 & 0.63 & 0.64 & 0.69 & 0.54 \\
Routes & 79 & \textbf{0.84} & 0.72 & 0.82 & \textbf{0.84} & 0.78 & \textbf{0.84} \\
Completed routes & 62 & 0.71 & 0.73 & 0.80 & \textbf{0.85} & 0.77 & 0.82 \\
\midrule
Swap pairs (Part~B) & 139 & \textbf{0.83} & 0.71 & 0.68 & 0.67 & 0.71 & 0.66 \\
\bottomrule
\end{tabular*}
\endgroup
\end{table}

\begin{figure}[H]
    \centering
    \includegraphics{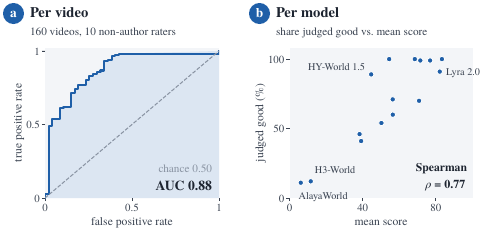}
    \caption{\textbf{Control scores align with human judgments (Part~A).} (a)~ROC curve of the per-video score for predicting the majority rating of ten raters across $160$ randomly sampled videos (AUC $0.88$). (b)~Model-level correlation between the share of positive judgments and the mean score of the rated videos ($\rho=0.77$).}
    \label{fig:human3}
\end{figure}

\begin{table}[H]
\centering
\caption{\textbf{Agreement between ten raters and the scores.} Part~A: single videos. Part~B: completed G2--G4 swap pairs. Part~C: video pairs whose Rigidity offsets differ at least twofold. Video and pair rows of Parts~A and~B give the ROC AUC, model rows Spearman's $\rho$ across models, and Part~C rows the share of majority choices that agree with the smaller offset, all with $95\%$ bootstrap intervals or binomial $p$-values.}
\label{tab:human3}
\begingroup\footnotesize\setlength{\tabcolsep}{4pt}\renewcommand{\arraystretch}{1.08}
\begin{tabular*}{0.92\linewidth}{@{\extracolsep{\fill}}llrl@{}}
\toprule
Part, level & Subset / score & $n$ & Agreement \\
\midrule
A, video & All sampled videos (primary) & 160 & AUC 0.88 [0.82, 0.93] \\
A, video & \quad single-action videos (\mHold{}) & 80 & AUC 0.87 [0.73, 0.98] \\
A, video & \quad route videos (route score) & 80 & AUC 0.85 [0.76, 0.93] \\
A, video & \quad completed routes only & 62 & AUC 0.71 [0.57, 0.84] \\
A, video & \quad route completion alone (binary) & 80 & AUC 0.75 [0.67, 0.82] \\
A, video & \quad gain of the route score over completion alone & 80 & $+$0.10 [0.04, 0.16] \\
A, model & Positive judgment share vs.\ mean score & 15 & $\rho$ 0.77 [0.35, 0.96] \\
\midrule
B, pair & Pair score (primary) & 139 & AUC 0.83 [0.76, 0.89] \\
B, pair & \quad pair translation / yaw score & 139 & AUC 0.83 / 0.76 \\
B, model & Same-place share vs.\ mean pair score & 14 & $\rho$ 0.76 [0.35, 0.96] \\
\midrule
C, pair & Strong pairs, majority agrees with offset (primary) & 31 & 31/31, $p<10^{-9}$ \\
C, pair & Moderate pairs, majority agrees with offset & 8 & 8/8, $p=0.008$ \\
C, rater & Individual agreement with offset, median (range) & 10 & 95\% (87.5--100\%) \\
\midrule
Reliability & Fleiss' $\kappa$ (Part A / B / C) & 160 / 140 / 40 & 0.87 / 0.83 / 0.86 \\
Reliability & Identical repeated judgments (Part A / B / C) & 82 / 100 / 60 & 81 / 95 / 56 \\
\bottomrule
\end{tabular*}
\endgroup
\end{table}

Two earlier rounds rated $130$ additional single-action and route videos across thirteen models using the same Part~A question and interface. Raters agreed with Fleiss' $\kappa=0.82$, the per-video score separated videos judged good with an AUC of $0.86$ [$0.79$, $0.92$] on the $125$ pose-measurable clips, and the share of positive judgments correlated with the mean score across models at $\rho=0.87$ [$0.55$, $0.97$]. None of the $13$ incomplete routes in those rounds was rated good by the majority of raters.

\FloatBarrier
\section{Fine-Tuning Details}\label{app:I}
This appendix gives the training data, recipe, and further results of the fine-tuning experiments of Sec.~\ref{sec:sft}. For HY-World 1.5, the model of Tab.~\ref{tab:sft} is SFT~(F), trained on clips rendered at the full step magnitude of the benchmark command, and an ablation SFT~(H) is trained on clips rendered at half that magnitude. Both are compared with the non-distilled base model (N0) and the released step-distilled checkpoint (Released).

\begin{table}[H]
\centering
\caption{\textbf{Supervised fine-tuning (SFT)} on procedural clips rendered at each model's own command magnitude. Base: non-distilled base model. Rel.: released distilled streaming model from Tab.~\ref{tab:main}. Each column covers $3{,}640$ videos. Stall, Completed, and Near are percentages of routes, End err.\ and Swap gap are medians in strides, and Offset is the Rigidity offset of Tab.~\ref{tab:main}.}
\label{tab:sft}
\begingroup
\footnotesize
\setlength{\tabcolsep}{2.5pt}
\renewcommand{\arraystretch}{1.06}
\begin{tabular*}{\linewidth}{@{\extracolsep{\fill}}lcccccc@{}}
\toprule
 & \multicolumn{3}{c}{HY-World 1.5} & \multicolumn{3}{c}{SANA-WM} \\
\cmidrule(lr){2-4}\cmidrule(lr){5-7}
 & Base & SFT & \textcolor{gray}{Rel.} & Base & SFT & \textcolor{gray}{Rel.} \\
\midrule
Stall (\%)$\downarrow$ & 72.0 & 11.2 & \textcolor{gray}{54.0} & 57.9 & 23.1 & \textcolor{gray}{36.5} \\
Completed (\%)$\uparrow$ & 27.5 & 88.8 & \textcolor{gray}{46.0} & 41.1 & 76.8 & \textcolor{gray}{62.4} \\
\midrule
End err.\ (str.)$\downarrow$ & 4.07 & 2.33 & \textcolor{gray}{2.42} & 3.66 & 1.46 & \textcolor{gray}{2.47} \\
Swap gap (str.)$\downarrow$ & 3.60 & 2.48 & \textcolor{gray}{1.56} & 1.65 & 0.71 & \textcolor{gray}{1.95} \\
Near (\%)$\uparrow$ & 0.7 & 11.4 & \textcolor{gray}{3.8} & 1.9 & 24.1 & \textcolor{gray}{7.2} \\
\midrule
\mComp$\uparrow$ & 10.1 & 48.7 & \textcolor{gray}{21.8} & 16.1 & 47.0 & \textcolor{gray}{28.3} \\
\mHold$\uparrow$ & 74.1 & 90.1 & \textcolor{gray}{77.7} & 73.3 & 81.8 & \textcolor{gray}{76.2} \\
\mResp\ (\%)$\uparrow$ & 95.9 & 99.6 & \textcolor{gray}{99.6} & 90.2 & 99.6 & \textcolor{gray}{99.4} \\
\midrule
Offset ($^\circ$)$\downarrow$ & 0.33 & 0.30 & \textcolor{gray}{0.33} & 0.28 & 0.16 & \textcolor{gray}{0.27} \\
MUSIQ & 69.5 & 70.6 & \textcolor{gray}{71.6} & 74.2 & 74.3 & \textcolor{gray}{69.6} \\
\bottomrule
\end{tabular*}
\endgroup
\end{table}

\subsection{Training data and recipe}\label{app:asset_audit}
Training clips are rendered in Blender in five procedurally generated indoor layouts, four for training and one held out for validation, each seen from two start views. The $112$ training clips comprise $48$ single-action clips ($4$ layouts $\times$ $2$ views $\times$ $6$ actions, $16$ steps each) and $64$ route clips ($8$ route programs $\times$ $4$ layouts $\times$ $2$ views, $8$--$16$ steps each), and the $20$ validation clips ($12$ single-action and $8$ route clips) come from the held-out layout. Every camera path moves by an exact displacement and yaw increment at each step and is rendered at $832\times480$ with four frames per step, the native interface of HY-World 1.5. SFT~(F) uses $0.08$ scene units and $3^\circ$ per step, matching the benchmark adapter, and SFT~(H) uses $0.04$ scene units and $1.5^\circ$ per step. Tab.~\ref{tab:I.routes} lists the eight route programs. All meshes and materials of the layouts are synthesized procedurally, and the training scenes and routes are disjoint from the scenes and routes of the benchmark.

Both runs train a rank-$32$ LoRA ($\alpha=32$) on the query, key, value, and output projections of all $54$ transformer blocks of the non-distilled HY-World 1.5 action model, with a learning rate of $10^{-5}$, $16$ linear warm-up updates, gradient norm clipping at $1.0$, and a global batch size of $16$ clips on eight H200 GPUs for $512$ updates.

\begin{table}[H]
\centering
\caption{\textbf{The eight route programs of the training clips (one step per symbol, superscripts denote repetitions).}}
\label{tab:I.routes}
\begingroup\footnotesize\setlength{\tabcolsep}{5pt}\renewcommand{\arraystretch}{1.08}
\begin{tabular*}{0.7\linewidth}{@{\extracolsep{\fill}}lllc@{}}
\toprule
Construction & Route a & Route b & Steps \\
\midrule
Turn--translate order & A$^8$\,YR$^8$ & YR$^8$\,A$^8$ & 16 \\
Segment order & A$^3$D$^3$S$^3$W$^3$ & S$^3$W$^3$A$^3$D$^3$ & 12 \\
Turn and counter-turn & YR$^2$A$^4$YL$^2$ & YL$^2$A$^4$YR$^2$ & 8 \\
Two-turn order & YR$^3$A$^3$YL$^3$S$^3$ & YR$^3$S$^3$YL$^3$A$^3$ & 12 \\
\bottomrule
\end{tabular*}
\endgroup
\end{table}

\subsection{Checkpoint selection}\label{app:sft_b2}
Checkpoints are selected on the $20$ validation clips alone, by the highest share of completed route pairs and then the highest ungated \mHold, scored with the unchanged benchmark pipeline. As shown in Fig.~\ref{fig:I.ckpt}, validation performance of the full-step run peaks and plateaus at update $256$, and the rule also selects update $256$ of the half-step run, so update $256$ of each run is evaluated on all $3{,}640$ benchmark requests.

\begin{figure}[H]
\centering
\includegraphics[width=0.85\linewidth]{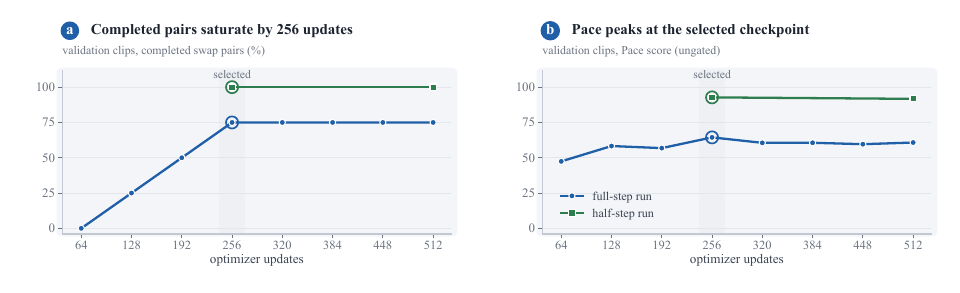}
\caption{\textbf{Validation completed pairs and ungated \mHold\ across training updates ($20$ held-out clips, seed 0).} The half-step run is scored at updates $256$ and $512$. Circles: the selected update-$256$ checkpoints.}
\label{fig:I.ckpt}
\end{figure}

\subsection{Generalization across real and rendered domains}\label{app:sft_split}
All training clips are rendered indoor layouts, while the benchmark comprises $36$ real-world ETH3D start views ($936$ videos, $288$ routes) and $104$ UE5 start views ($2{,}704$ videos, $832$ routes). As shown in Tab.~\ref{tab:I.3}, both SFT~(H) and SFT~(F) improve route completion, \mComp, \mHold, and repeat variation over N0 on the real photos as well as on the renders, without lowering MUSIQ or Aesthetic scores. On the ETH3D views, SFT~(F) raises completed routes from $43.1\%$ to $83.3\%$ and \mComp\ from $20.0$ to $47.1$, so the gains learned from synthetic indoor clips transfer to real images.

\begin{table}[H]
\centering
\caption{\textbf{Fine-tuning results on real-world ETH3D views ($36$ views, $936$ videos) and rendered UE5 views ($104$ views, $2{,}704$ videos).} Repeat variation is $\downarrow$, all other consistency metrics are $\uparrow$.}
\label{tab:I.3}
\begingroup\footnotesize\setlength{\tabcolsep}{2.6pt}\renewcommand{\arraystretch}{1.08}
\begin{tabular*}{\linewidth}{@{\extracolsep{\fill}}lcccccccc@{}}
\toprule
& \multicolumn{4}{c}{ETH3D (real photos)} & \multicolumn{4}{c}{UE5 (renders)} \\
\cmidrule(lr){2-5}\cmidrule(l){6-9}
Metric & Released & N0 & SFT (H) & SFT (F) & Released & N0 & SFT (H) & SFT (F) \\
\midrule
\mResp\ (\%) & 100.0 & 99.5 & 100.0 & 100.0 & 99.6 & 95.0 & 98.7 & 99.5 \\
\mHold & 91.9 & 88.1 & 88.9 & 90.2 & 74.1 & 70.6 & 89.4 & 90.1 \\
Completed (\%) & 61.1 & 43.1 & 87.2 & 83.3 & 40.7 & 22.1 & 87.7 & 90.6 \\
\mRoute\ / \mPair & 36.2 / 34.4 & 21.9 / 18.0 & 48.7 / 48.6 & 47.8 / 46.3 & 20.7 / 13.6 & 8.2 / 5.2 & 47.4 / 46.1 & 49.9 / 48.7 \\
\mComp & 35.3 & 20.0 & 48.6 & 47.1 & 17.2 & 6.7 & 46.8 & 49.3 \\
\mScene & 86.4 & 84.4 & 90.7 & 88.8 & 83.8 & 80.5 & 88.2 & 83.1 \\
Repeat, translation (\%) & 20.5 & 33.4 & 14.8 & 15.4 & 52.6 & 67.2 & 19.5 & 16.6 \\
MUSIQ & -- & 74.8 & 75.8 & 75.3 & -- & 67.7 & 69.5 & 68.9 \\
Aesthetic & -- & 47.6 & 48.3 & 48.0 & -- & 60.5 & 62.0 & 61.7 \\
\bottomrule
\end{tabular*}
\endgroup
\end{table}

\FloatBarrier
\subsection{Command-scaled stall control and perturbed-path ablation}\label{app:sftmatch}
Fine-tuning also changes the step size of the model, and doubling the training step magnitude from SFT~(H) to SFT~(F) scales the translation stride of the fine-tuned model by $1.7\times$--$1.8\times$ and its yaw stride by $1.5\times$, which brings its step size closer to that of the released model while keeping the consistency gains. Since the stall threshold is $10\%$ of a model's own stride, we test on the $30$-view subset of App.~\ref{app:magsens} ($420$ videos per setting) whether the stall reduction depends on stride length.

As shown in Tab.~\ref{tab:sftmatch}, scaling the translation command of N0 by $0.88\times$ and its yaw command by $1.04\times$ barely changes its forward stride ($0.97\times$) and leaves its stall rate unchanged ($68.8\%$ vs.\ $67.9\%$). We therefore compare directly on the start views where the calibrated stride of N0 itself falls in the stride range of SFT~(F). On these views N0 still stalls on $67.5\%$ of its routes, against $13.8\%$ for SFT~(F) on all $30$ views, so shorter strides do not explain the stall reduction. Fine-tuning also removes stalls at action switches, which fall from $15\%$ to $0.1\%$ of the switch steps.

We further repeat the training of SFT~(F) on the same $112$ clips rendered along perturbed camera paths, with labels, recipe, and checkpoint unchanged. Each step's displacement or turn is multiplied by an independent log-normal factor ($\sigma=0.3$, clipped to $[0.4, 2.0]$), and each action switch is shifted uniformly by up to $\pm 0.25$ steps. Training on perturbed paths lowers stalled routes from $67.9\%$ to $24.2\%$ and raises \mComp\ from $11.4$ to $39.7$, and training on exact paths lowers stalls further to $13.8\%$ and raises \mComp\ to $48.3$ and \mHold\ to $92.5$, against $87.0$ with perturbed paths. The perturbed clips, in which every command step still moves the camera, thus retain about $80\%$ of the stall reduction ($43.7$ of $54.1$ points) and about three quarters of the gain in \mComp, and exact step lengths add the rest.

\begin{table}[!htbp]
\centering
\caption{\textbf{Command-scaled control and perturbed-path ablation across $30$ start views ($420$ videos each).} N0: unmodified base model. N0, $0.88\times$: command-scaled base model. SFT: exact-path fine-tuning. Perturbed: perturbed-path fine-tuning.}
\label{tab:sftmatch}
\begingroup\footnotesize\setlength{\tabcolsep}{4pt}
\renewcommand{\arraystretch}{1.08}\begin{tabular*}{0.74\linewidth}{@{\extracolsep{\fill}}lcccc@{}}
\toprule
 & N0 & N0, $0.88\times$ & SFT & Perturbed \\
\midrule
Forward stride (rel.\ N0) & 1.00 & 0.97 & 0.52 & 0.53 \\
Turn stride ($^\circ$) & 1.25 & 1.27 & 1.75 & 1.57 \\
\mResp\ (\%) & 96.1 & 95.6 & 100.0 & 100.0 \\
\mHold & 73.6 & 71.6 & 92.5 & 87.0 \\
Pace ratio & 1.01 & 1.01 & 1.03 & 1.00 \\
\mComp & 11.4 & 9.6 & 48.3 & 39.7 \\
Completed (\%) & 30.8 & 30.0 & 86.2 & 75.8 \\
Near (\%) & 2.1 & 0.4 & 15.4 & 10.8 \\
Stall (\%) & 67.9 & 68.8 & 13.8 & 24.2 \\
End err.\ (strides) & 4.68 & 5.91 & 2.14 & 2.23 \\
Swap gap (strides) & 2.77 & 4.06 & 2.34 & 1.98 \\
Switch steps misdirected / stalled (\%) & 38 / 15 & 39 / 14 & 22 / 0 & 23 / 1 \\
\bottomrule
\end{tabular*}
\endgroup
\end{table}

\subsection{Second architecture: SANA-WM}\label{app:sft_sana}
We fine-tune the non-distilled bidirectional SANA-WM teacher model ($60$ denoising steps, classifier-free guidance scale $5.0$, paired with the LTX-2 refiner), whose distilled streaming variant is the released model in Tab.~\ref{tab:main}. A rank-$32$ LoRA ($\alpha=32$) is attached to the query, key, value, and output projections of both the backbone and the camera-conditioning branch ($160$ adapter modules, $22.9$M trainable parameters). The $112$ training and $20$ validation clips use the same layouts and route programs as for HY-World 1.5, rendered at the native command magnitude of SANA-WM of $0.1$ translation units and $2.4^\circ$ yaw per step at four frames per step. Training uses a learning rate of $10^{-5}$, $16$ warm-up updates, and an effective batch size of $16$ clips ($4$ GPUs with gradient accumulation of $4$) for $2{,}048$ updates. From update $512$ on, all validation route pairs are completed, and the other validation scores change little after about $1{,}000$ updates, so Tab.~\ref{tab:sft} reports the final checkpoint.

Across $2{,}000$ paired bootstrap resamples of the $140$ start views, the $95\%$ intervals of the improvements over the base model exclude zero for every consistency metric in Tab.~\ref{tab:sft}. Stalled routes change by $-34.7$ [$-38.8$, $-30.5$] percentage points, completed routes by $+35.7$ [$+31.6$, $+39.9$], and \mComp\ by $+30.9$ [$+28.3$, $+33.5$], while MUSIQ remains unchanged ($74.2$ vs.\ $74.3$).

The Rigidity offsets in Tab.~\ref{tab:sft} are computed from the saved camera fits of the benchmark runs with the statistic of Tab.~\ref{tab:main}, and the same extraction reproduces the offset of the released HY-World 1.5 videos in Tab.~\ref{tab:main}. Against the released step-distilled HY-World 1.5 rather than its base model, the fine-tuned model stalls far less and doubles the \mComp. Relative to the base models, fine-tuning lowers the offset by $8\%$ for HY-World 1.5 and by $43\%$ for SANA-WM, with paired $95\%$ bootstrap intervals over the $140$ start views of $1\%$ to $15\%$ and $36\%$ to $51\%$.

\FloatBarrier
\section{Reproducibility and Supplementary Video}\label{app:repro}
StrideWorld is built on synthetic scenes rendered in Unreal Engine~5 and real scenes from ETH3D. We will release all $140$ start views with ground-truth RGB images, depth maps, and camera parameters, together with the model adapters and input command sequences for all \nmodels{} models, the evaluation and scoring code, and the recovered camera trajectories and scores of all \nvideos{} generated videos. Start-view preprocessing and control settings are given in App.~\ref{app:A}, and pose recovery and scoring in App.~\ref{app:C} and~\ref{app:D}.

The supplementary video (\texttt{StrideWorld\_supp\_video.mp4}, $64$\,s) shows the measurements of the three tests of Fig.~\ref{fig:teaser} on generated videos, each for two models on the same start view and commands, with step-synchronized command overlays. For Rigidity it overlays the rigid projections of the held-out anchors and their tracked positions, for Pace it compares the distance covered by a held action with the prediction from its first four steps, and for Composition it draws the generated camera path of a route next to the path chained from single-action strides. A final pair shows HY-World 1.5 on the same route before and after fine-tuning (Sec.~\ref{sec:sft}).

\FloatBarrier
\section{\texorpdfstring{Evidence for Tab.~\ref{tab:related}}{Evidence for Table 1}}\label{app:related}
Tab.~\ref{tab:related} summarizes the benchmarks closest to StrideWorld, and the evidence behind every cell is listed below with its source sections. The 3D-test column records diagnostics of scene consistency rather than a common protocol on held-out points, and human validation denotes a reported comparison whose scope differs across studies. Benchmarks made public after September~15, 2026 are concurrent work. Of these, Orbia~\citep{orbia} appears in Tab.~\ref{tab:related} and is marked $^\dagger$. StreamEval~\citep{streameval}, also concurrent, scores whether each online control produces the requested scene operation in streaming generation and is not listed in the table. WorldRoamBench~\citep{worldodysseybench} was first released as WorldOdysseyBench in June 2026.
\begin{itemize}[leftmargin=*,itemsep=1pt,topsep=2pt]
\item \textbf{WorldScore}~\citep{worldscore}: Camera and text specifications drive sequential next-scene generation (Sec.~3.1). The camera error fits a translation scale, and 3D consistency is a dense reprojection error after joint refinement of depth and pose rather than a test on points held out from the fit (Sec.~3.3, Supp.~C.1--C.2). No held-action pace test is reported, and comparisons between metrics and human judgments are given in Supp.~D.
\item \textbf{WorldMark}~\citep{worldmark}: Key, camera, and text adapters are evaluated for direction, latency, and the speed stability of held commands from optical flow (Secs.~3.1.1, 3.4), and the action suite includes open sequences and round trips (Sec.~3.3). Global Memory uses cross-frame geometry from VGGT-$\Omega$ (Sec.~3.1.2), which makes it a 3D test rather than only an appearance proxy. Forty raters assess selected metrics, excluding Global Memory (Sec.~4.3).
\item \textbf{iWorld-Bench}~\citep{iworld}: Three native interfaces are tested on action tasks, camera following, and cyclic memory tasks (Sec.~3.2). ViPE trajectories support comparisons of direction and with reference trajectories, and the symmetry of images and displacements serves as an indirect proxy for scene consistency (Sec.~3.3). The human study covers visual quality and camera control with 12 raters and 16 tasks per model (Sec.~4.2.1, App.~E.1) and does not validate rigidity specifically.
\item \textbf{WBench}~\citep{wbench}: Key, camera, and text models are evaluated by comparing MegaSaM trajectories with references derived from the actions, together with the consistency of repeated actions (Sec.~4, App.~C.4.1), which is a proxy for pace rather than a direct trace of the speed of a held key. Navigation includes open paths and loops (App.~A.3), and reprojection with DA3 supplies the geometry tests (App.~C.5.6). Four hundred raters compare ten aspects at the granularity of model rankings (Sec.~5.4).
\item \textbf{WorldExam}~\citep{worldexam}: Camera, action, and text interfaces are evaluated, with camera inputs using their own trajectory as the reference and action and text segments using prescribed directions with magnitude normalization. Open control sequences and revisits through inverse actions are both evaluated (Secs.~3.2, 4.1), and geometry uses cycle reprojection with VGGT-$\Omega$. Human validation concerns the reactivity checklists and not all geometry metrics (Sec.~5.6).
\item \textbf{WorldRoamBench}~\citep{worldodysseybench}: A shared keyboard vocabulary is adapted to native key, camera, and text models (App.~B.1), and direction labels are complemented by adaptive reference trajectories with arc-length alignment (Sec.~3.2, App.~E). Open action sequences and revisits are included (Sec.~4.2). Geometry compares reconstructed point clouds of observations and revisits, and crops of engine references additionally support tests of perceptual consistency (Secs.~3.4--3.5, App.~G.2.3). Human verification concerns the physics and subject-memory judgments of a VLM (App.~L.2).
\item \textbf{MIND}~\citep{mind}: Key actions are tested at changed action scales (Secs.~3.2--3.3), with action accuracy measured from ViPE trajectories by Sim(3) alignment and relative pose errors, and generalization across the action space measured by the MSE between generated and reference frames (Sec.~3.5). Image MSE between forward and reverse passes tests revisits as an appearance proxy rather than held-out 3D rigidity. No direct test of speed decay under a held command is reported, and a human evaluation is mentioned without a protocol or an agreement statistic (Sec.~4.3).
\item \textbf{World Models as Group Actions}~\citep{wang2026world}: Continuous ego-motion increments are probed with identity, inverse, and local composition relations (Secs.~2.1, 3.3). DROID-SLAM recovers the motion states, and robustness is also assessed by comparing full trajectories across stochastic rollouts (App.~C.4, C.7), so ``Poses'' denotes the recovery of motion states rather than a test of scene rigidity. Neither a pace test for held commands nor a human validation is reported.
\item \textbf{WorldCycle (CycleBench)}~\citep{worldcycle}: Key and camera models are tested on reversible, closed, repeated, and cascaded cycles (Sec.~4, App.~A). RoMa evaluates endpoint closure, mirrored forward and reverse frames, and phase-aligned states of repeated cycles, so the evaluation is not limited to endpoints, and the drift over repeated cycles serves as an indirect pace proxy in image space. Neither 3D rigidity nor a human validation is reported.
\item \textbf{Orbia}$^\dagger$~\citep{orbia}: Native-interface adapters are evaluated with directional control scores and templates for exploration and revisits (Sec.~3, App.~C.1). Geometry uses registered RGB-D, cross-view tests with recovered cameras, and Gaussian rendering of 16 held-out views from 64 context frames (Sec.~4), which holds out views rather than splitting anchors into pose and test sets as StrideWorld does. Twenty raters compare 90 cases across six models and six evaluation tiers (Sec.~5.7), and no pace test for held actions is reported.
\end{itemize}

\FloatBarrier
\section{Limitations}\label{app:limitations}
\paragraph{Scope.} Strides test consistency with a model's own first steps and cannot be measured reliably when a model moves too little in its first steps (marked $^\dagger$). Anchors cover only surfaces visible in the start view, so large turns that move these surfaces out of view leave the measurable range, and we report scores on recoverable subsets alongside the overall scores. For the same reason, StrideWorld does not evaluate long-horizon exploration or revisits, whose references would have to be reconstructed from the generated frames themselves. All scenes are static and the main test scores planar motion, with pitch evaluated separately (App.~\ref{app:G2}, \ref{app:envelope}, and~\ref{app:length}). Extending anchor-based and self-referential evaluation to dynamic scenes and full 6-DoF trajectories is a natural next step.

\paragraph{Measurement.} The Rigidity offset measures departure from the true geometry of the start view, so it covers both deformation over time and a scene that stays rigid at a wrong depth, and the refitting experiments of App.~\ref{app:ceiling} separate the two. Because the camera is recovered from the same generated frames and tracks that the Rigidity test uses, a deforming scene shifts the recovered camera as well, by a median of $0.64$ strides at $1^\circ$ of smooth deformation. At deformation comparable to that of the generated videos, this changes the Near decision of $8\%$ to $20\%$ of the counterpart routes while leaving the model ranking unchanged (App.~\ref{app:perturb}). The \mComp\ compares where routes end, and the per-step diagnostics complement it along the route.

\paragraph{Control settings.} Every model is evaluated with its native interface and step duration, and with command magnitudes that are the release defaults for some models and set within the measurable range for others (App.~\ref{app:magsens}), so the same route spans different amounts of camera motion and generation time across models. Halving or doubling the command magnitude of three models leaves their route findings unchanged, and restricting all models to routes with a matched amount of camera motion keeps their ranking unchanged (App.~\ref{app:matched}). After fine-tuning, the median completed route of both models still ends more than one stride from its prediction.

\fi

\end{document}